\documentclass[lettersize,journal]{IEEEtran}
\usepackage{amsmath,amsfonts}
\usepackage{algorithm}
\usepackage{algpseudocode}
\usepackage{cite}
\usepackage{array}
\usepackage[caption=false,font=normalsize,labelfont=sf,textfont=sf]{subfig}
\usepackage{textcomp}
\usepackage{stfloats}
\usepackage{url}
\usepackage{verbatim}
\usepackage{graphicx}

\usepackage{float}
\usepackage{booktabs,multirow}
\usepackage{xcolor}

\def\BibTeX{{\rm B\kern-.05em{\sc i\kern-.025em b}\kern-.08em
    T\kern-.1667em\lower.7ex\hbox{E}\kern-.125emX}}
\usepackage{balance}

\newtheorem{theorem}{Theorem}
\newtheorem{proposition}{Proposition}

\newcommand\liang[2]{{\footnotesize\color{gray}#1}{\color{magenta}#2}}

\begin{document}
\title{
Semi-Supervised Clustering via Dynamic Graph Structure Learning
}
\author{Huaming Ling, Chenglong Bao,
\IEEEmembership{Member, IEEE}, Xin Liang, and Zuoqiang Shi
\thanks{H. Ling, C. Bao, X. Liang and Z. Shi are with the Yau Mathematical Sciences Center, Tsinghua University and Yanqi Lake Beijing Institute of Mathematical Sciences and Applications,
Beijing 100084, China (e-mail: linghm18@mails.tsinghua.edu.cn;  clbao@mail.tsinghua.edu.cn; 
liangxinslm@tsinghua.edu.cn;
zqshi@mail.tsinghua.edu.cn).
}
\thanks{C. Bao and Z. Shi are the co-corresponding authors.}
}


\maketitle

\begin{abstract}
Most existing semi-supervised graph-based clustering methods exploit the supervisory information by either refining the affinity matrix or directly constraining the low-dimensional representations of data points. The affinity matrix represents the graph structure and is vital to the performance of semi-supervised graph-based clustering. However, existing methods adopt a static affinity matrix to learn the low-dimensional representations of data points and do not optimize the affinity matrix during the learning process. In this paper, we propose a novel dynamic graph structure learning method for semi-supervised clustering. In this method, we simultaneously optimize the affinity matrix and the low-dimensional representations of data points by leveraging the given pairwise constraints. Moreover, we propose an alternating minimization approach with proven convergence to solve the proposed nonconvex model. During the iteration process, our method cyclically updates the low-dimensional representations of data points and refines the affinity matrix, leading to a dynamic affinity matrix (graph structure). Specifically, for the update of the affinity matrix, we enforce the data points with remarkably different low-dimensional representations to have an affinity value of 0. Furthermore, we construct the initial affinity matrix by integrating the local distance and global self-representation among data points. Experimental results on eight benchmark datasets under different settings show the advantages of the proposed approach.
\end{abstract}


\section{Introduction}
\IEEEPARstart{C}{lustering} is an important problem in data mining and machine learning. Clustering aims to partition a set of data points into several groups so that data points in the same group (cluster) are more similar to each other than those in other groups (clusters). Besides the data points, clustering methods do not require any supervisory information. In practice, although it is difficult to obtain the exact label information, partial/week supervisory information is available. One common supervisory information is given in the form of pairwise constraints, including must-link (ML) constraints (the pair of data points must belong to the same cluster) and cannot-link (CL) constraints (the pair of data points must belong to different clusters). Thus, semi-supervised clustering is to cluster data points by exploring the supervisory information and has been extensively studied during the past two decades, such as \cite{wagstaff2001constrained,anand2013semi,zeng2011semi,xiong2013active,kulis2009semi,basu2004probabilistic,bai2020semi,zass2005unifying}. Moreover, semi-supervised clustering has been widely applied in many tasks including medical diagnosis \cite{thangavel2010semi}, natural language processing \cite{huang2009active}, bioinformatics \cite{yu2017adaptive}, image processing \cite{ahn2016face,saha2016brain}, social networks \cite{yang2014unified} and information networks \cite{li2017semi}.

Among clustering methods, one important branch is graph-based clustering like spectral clustering (SC) \cite{ng2002spectral}. The performance of graph-based methods highly depends on the quality of the graph structure (affinity matrix) that represents the similarity among data points. Most of the existing semi-supervised graph-based clustering methods explore the supervisory information in the following two ways \cite{jia2018semi}: 1) refining the affinity matrix with the supervisory information; 2) constraining the low-dimensional representations with the supervisory information. For the first way, \cite{kamvar2003spectral} sets the corresponding element of the affinity matrix to 1 (resp. 0) if the two data points belong to ML (resp. CL). In \cite{wang2009integrated}, a positive (resp. negative) term is added to the corresponding element of the affinity matrix if the two data points belong to ML (resp. CL). Furthermore, \cite{lu2008constrained} refines the affinity matrix through an affinity propagation method.  \cite{nie2021semi} introduces a novel cannot-link graph regularization to enforce cannot-link constrained data points in different clusters. For the second way,  \cite{li2009constrained} proposes a constrained spectral clustering model to adapt the spectral representation towards an ideal representation as consistent with the pairwise constraints as possible. \cite{wang2010flexible} and \cite{wang2014constrained} propose a constrained spectral clustering model which uses a user-specific parameter to constrain how well the pairwise constraints are satisfied.  \cite{yang2014unified} introduces a semi-supervised clustering model to enforce the low-dimensional representations of ML constrained data points to be similar. 

\begin{figure*}[!t]
\centering
\includegraphics[width=6in]{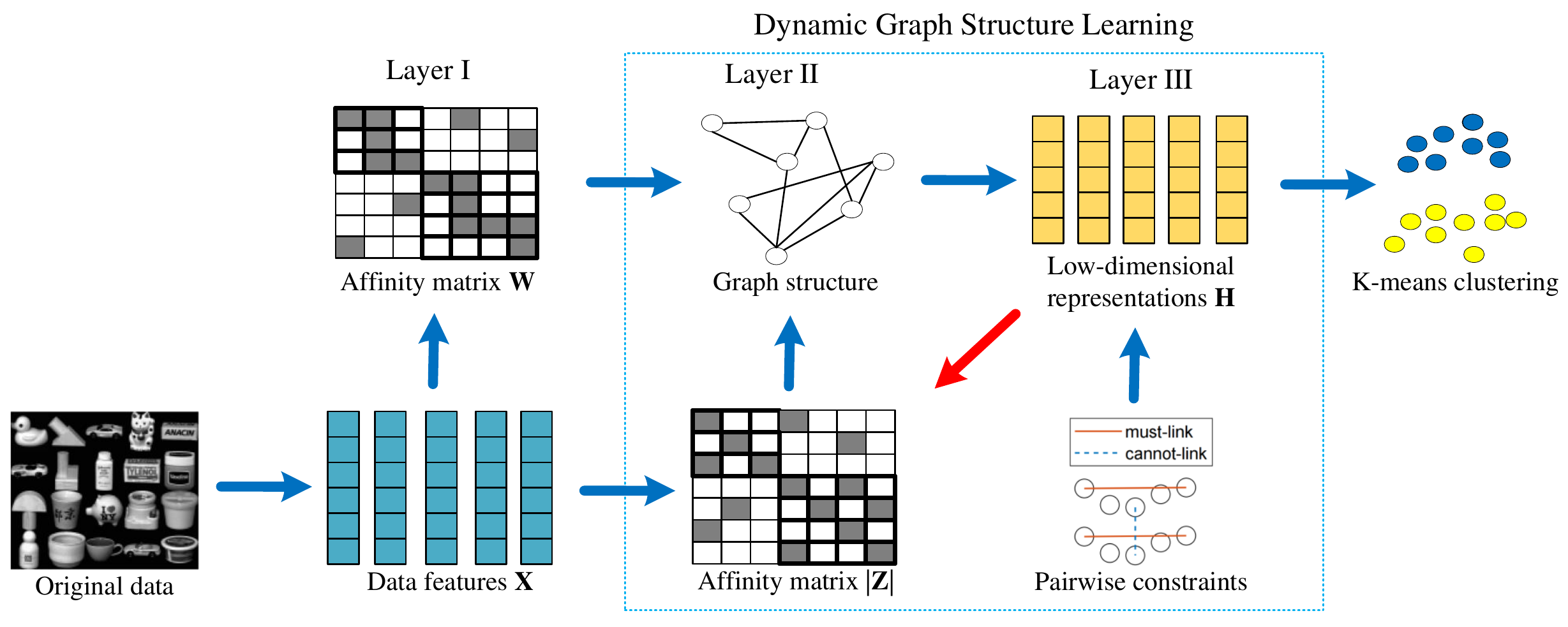}%
\caption{Illustration of the Dynamic Graph Structure Learning (DGSL) method for semi-supervised clustering. We simultaneously learn the affinity matrix $|\mathbf{Z}|$ and the low-dimensional representations $\mathbf{H}$ for all data points with the guidance of the given pairwise constraints. We construct the graph structure by integrating the local distance ($\mathbf{W}$) and global self-representation ($|\mathbf{Z}|$) among data points. During the learning process, DGSL cyclically updates the low-dimensional representations $\mathbf{H}$  with the graph structure as well as the pairwise constraints and refines the affinity matrix $|\mathbf{Z}|$ with $\mathbf{H}$, leading to a dynamic graph structure.}
\label{fig:DGSL diagram}
\end{figure*}

The performance of semi-supervised graph-based clustering highly depends on the affinity matrix. However, these methods do not change the affinity matrix during the learning of low-dimensional representations of data points, leading to a static affinity matrix. In this paper, we introduce a dynamic graph structure learning method for semi-supervised clustering. Given the initial affinity matrix, we propose a unified optimization framework to simultaneously optimize the affinity matrix itself and the low-dimensional representations of data points with the guidance of the given pairwise constraints, as shown in Fig. \ref{fig:DGSL diagram}.  In the unified optimization framework, we cyclically update the low-dimensional representations of data points and refine the affinity matrix during the iteration process. Specifically, we enforce the data points that belong to ML (resp. CL) to have similar (resp. different) low-dimensional representations. For the update of the affinity matrix, we enforce the data points with remarkably distinct low-dimensional representations to have an affinity value of 0. Thus the affinity matrix (graph structure) is dynamically updated during the cyclically updating process.
 
As a comparison, approaches in \cite{fang2015robust,li2015learning,wang2018unified} integrate the construction of the affinity matrix and the propagation of the partial labels into a unified optimization framework. However, these approaches use partial labels as supervisory information and learn the label matrix for all data points. Our method uses pairwise constraints as supervisory information and learns the low-dimensional representations for all data points. Moreover, the affinity matrix in \cite{fang2015robust,li2015learning,wang2018unified} is constructed based on the global self-representation model. We obtain the affinity matrix by integrating the local distance and global self-representation among data points.
Existing approaches that define affinity can be roughly divided into two categories: 1) the first category 
connects each data point to its $m$ nearest neighbors which are defined according to the Euclidean distance, e.g., \cite{belkin2004semi,zhu2003semi,saul2003think,wang2007label};  and 2) the second category represents each data point as a linear combination of all data points and induces the affinity by the representation coefficients, e.g., \cite{yan2009semi,chen2013robust,lu2018subspace}. Approaches in the first category only connect each data point to its $m$ nearest neighbors but neglect those "non-neighbor" data points with the same class label, while approaches in the second category connect each data point to other data points in the same linear subspace. We construct the affinity matrix by integrating these two categories, and our main contributions are as follows.
\begin{itemize}
\item We propose a novel dynamic graph structure learning method for semi-supervised clustering. Specifically, we introduce a unified optimization framework to simultaneously optimize the graph structure (affinity matrix) and the low-dimensional representations of data points by leveraging pairwise constraints. Moreover, we construct the affinity matrix by integrating the local Euclidean distance and global self-representation among data points.

\item We propose an alternating minimization solver to solve the proposed nonconvex model with proven convergence. Specifically, the graph structure and the low-dimensional representations of data points are cyclically updated in the alternating minimization solver. Moreover, at each iteration, we refine the graph structure based on the low-dimensional representations of data points, leading to a dynamic graph structure. 

\item We evaluate our approach in eight benchmark datasets and compare it with several state-of-the-art semi-supervised graph-based clustering methods. Extensive experiments have been conducted under different settings of pairwise constraints to show the effectiveness of our approach. Moreover, we extend our approach to hypergraph datasets and achieve competitive performance compared with state-of-the-art hypergraph learning methods.
\end{itemize}

\section{Related Work}
\textbf{Notations.} We denote matrices by boldface uppercase letters, e.g., $\mathbf{W}$, vectors by boldface lowercase letters, e.g., $\mathbf{h}$, and scalars by lowercase letters, e.g., $w$. We denote $w_{ij}$ or $\mathbf{W}_{ij}$ as the $(i,j)$-th element of $\mathbf{W}$ and denote $\mathbf{w}_i$ as the $i$-th column of $\mathbf{W}$. We use Tr($\mathbf{W}$) to denote the trace of $\mathbf{W}$. $|\mathbf{W}|$ is the matrix consisting of the absolute value of each element of $\mathbf{W}$. We denote $\text{diag}(\mathbf{W})$ as a diagonal matrix with its $i$-th diagonal element being the $i$-th diagonal element of $\mathbf{W}$ and denote $\mathbf{W}^\top$ as the transpose of $\mathbf{W}$. We set $\mathbf{I}$ as the identity matrix and $\mathbf{0}$ as a matrix of all zeros. We give the notations of some norms, e.g., $\ell_1$-norm $\|\mathbf{W}\|_1=\sum_{ij}|w_{ij}|$, $\ell_\infty$-norm $\|\mathbf{w}\|_\infty=\max_{i}|w_i|$ and Frobenius norm (or $\ell_2$-norm of a vector) $\|\mathbf{W}\|=\sqrt{\sum_{ij}w_{ij}^2}$.

\subsection{Subspace Clustering}
Many high-dimensional datasets approximately lie on the union of multiple low-dimensional linear subspaces. For example, we can model the motion trajectories in a video, face images, and hand-written digits as the union of subspaces, with each subspace corresponding to a class. Such a subspace structure has motivated the problem of subspace clustering, which aims to group the data points into clusters, with each cluster corresponding to a subspace. Subspace clustering has been applied in many areas, such as motion segmentation \cite{elhamifar2013sparse}, face clustering \cite{zhang2019self}, and image processing \cite{yang2016ell}. Subspace clustering represents each data point as a linear combination of other data points. Such a representation is not unique, and sparse subspace clustering (SSC) \cite{elhamifar2013sparse} pursues a sparse representation by solving the following problem
\begin{equation}
\min_{\mathbf{Z}}~\|\mathbf{Z}\|_{1}+\gamma\|\mathbf{X-XZ}\|_1,
~\text {s.t.}~\text{diag}(\mathbf{Z})=\mathbf{0},
\end{equation}
where $\mathbf{Z}=(z_{ij}) \in \mathbb{R}^{n\times n}$ is the self-representation matrix for $n$ data points  $\mathbf{X}=[\mathbf{x}_1,\mathbf{x}_2,\ldots,\mathbf{x}_n]$, and $|z_{ij}|$ reflects the affinity between data point $\mathbf{x}_i$ and data point $\mathbf{x}_j$. $\|\mathbf{X-XZ}\|_1$ measures the representation error and outliers.  
Then the affinity matrix is obtained by $\frac{1}{2}\left(|\mathbf{Z}|+\left|\mathbf{Z}\right|^{\top}\right)$, which is further used for SC \cite{ng2002spectral} to obtain the final clustering results.

\subsection{Spectral Clustering}
Spectral Clustering (SC) \cite{ng2002spectral} is one of the most important clustering methods. Given an affinity matrix $\mathbf{W}\in \mathbb{R}^{n\times n}$ of $n$ data points $\mathbf{X}=[\mathbf{x}_1,\mathbf{x}_2,\ldots,\mathbf{x}_n]\in \mathbb{R}^{d\times n}$, SC first obtains the low-dimensional representations $\mathbf{H}=[\mathbf{h_1},\mathbf{h_2},\ldots,\mathbf{h}_n]\in \mathbb{R}^{k\times n}$ for all data points by solving 
\begin{equation}
    \min_{\mathbf{H}} ~\text{Tr}\left(\mathbf{HLH}^\top\right), ~\text{s.t.} ~\mathbf{HH}^\top = \mathbf{I},
    \label{eq:SC}
\end{equation}
where $\mathbf{L}= \mathbf{I}-\mathbf{D}^{-1/2}\mathbf{WD}^{-1/2}$ is the normalized Laplacian matrix and $\mathbf{D}$ is the diagonal matrix with each diagonal element $d_{ii} = \sum_{j=1}^n w_{ij}$. The optimal solution of (\ref{eq:SC}) can be obtained by computing the $k$ eigenvectors of $\mathbf{L}$ corresponding to the $k$ smallest eigenvalues. Then SC computes $\hat{\mathbf{H}}\in \mathbb{R}^{k\times n}$ by normalizing each column of $\mathbf{H}$ into unit Euclidean length and performs K-means on the columns of $\hat{\mathbf{H}}$ to obtain the final clustering results. 

\subsection{Linear Discriminant Analysis}
The Linear Discriminant Analysis (LDA) \cite{zhang2010fast, jia2009trace} is a popular approach in supervised learning for feature extraction and dimensionality reduction. Given a set of $n$ training data points $\{\mathbf{x}_i\}_{i=1}^n\in\mathbb{R}^d$ and the corresponding labels $\{y_i\}_{i=1}^n$, with $y_i\in \{1,2,\ldots,c\}$, LDA finds the linear transformation matrix $\mathbf{G}\in\mathbb{R}^{m\times d}$ to project high-dimensional vectors, $\mathbf{x}_1,\mathbf{x}_2,\ldots,\mathbf{x}_n\in\mathbb{R}^d$, into low-dimensional vectors, $\mathbf{G}\mathbf{x}_1,\mathbf{G}\mathbf{x}_2,\ldots,\mathbf{G}\mathbf{x}_n\in \mathbb{R}^m$,  by simultaneously maximizing the inter-class scatter and minimizing the intra-class scatter in the low-dimensional space
\begin{equation}
\label{eq:lda}
    \mathbf{G}^{*}=\underset{\mathbf{GG}^\top = \mathbf{I}}{\arg \max } ~ \frac{\sum_{i=1}^{c} \frac{n_{i}}{n}\left\|\mathbf{G} \mathbf{m}_{i}-\mathbf{G} \mathbf{m}\right\|^{2}}{\sum_{i=1}^{n} \frac{n_{y_{i}}}{n}\left\|\mathbf{G} \mathbf{x}_{i}-\mathbf{G} \mathbf{m}_{y_{i}}\right\|^{2}},
\end{equation}
where $\mathbf{m}_i$ is the centroid of the data points belonging to the  $i$-th class and $\mathbf{m}$ is the centroid of all data points, and $n_i$ is the number of data points belonging to the $i$-th class. The problem (\ref{eq:lda}) can be rewritten as a trace ratio problem
\begin{equation}
\label{eq:LDA}
\mathbf{G}^{*}=\underset{\mathbf{GG}^\top = \mathbf{I}}{\arg \max} ~ \frac{\text{Tr}\left(\mathbf{GS}_b\mathbf{G}^\top\right)}{\text{Tr}\left(\mathbf{GS}_w\mathbf{G}^\top\right)},
\end{equation}
where $\mathbf{S}_w=\sum_{i=1}^{n} \frac{n_{y_{i}}}{n}\left(\mathbf{x}_{i}-\mathbf{m}_{y_{i}}\right)\left(\mathbf{x}_{i}-\mathbf{m}_{y_{i}}\right)^\top$ is the intra-class scatter matrix and $\mathbf{S}_b=\sum_{i=1}^{c} \frac{n_{i}}{n}\left(\mathbf{m}_{i}-\mathbf{m}\right)\left(\mathbf{m}_{i}-\mathbf{m}\right)^\top$ is the inter-class scatter matrix.

\section{Semi-Supervised Clustering via Dynamic Graph Structure Learning}
\subsection{Problem Formulation}
  Given a set of data points $\mathcal{X}=\{\mathbf{x}_i\}_{i=1}^n$ and the sets of pairwise constraints, $\mathcal{M}$ and $\mathcal{C}$, with
  \begin{equation}
  \begin{aligned}
      \mathcal{M} &= \{(\mathbf{x}_i,\mathbf{x}_j)|\text{$\mathbf{x}_i$ and $\mathbf{x}_j$ belong to the same cluster}\},\\
      \mathcal{C} &= \{(\mathbf{x}_i,\mathbf{x}_j)|\text{$\mathbf{x}_i$ and $\mathbf{x}_j$ belong to different clusters}\},
      \end{aligned}
  \end{equation}
  we aim to learn the low-dimensional representations $\mathbf{H}$ for all data points such that the intra-class distance is small and the inter-class distance is relatively large for $\mathbf{H}$. Then we conduct K-means on $\mathbf{H}$ to obtain the final clustering results. We formulate the original feature matrix as  $\mathbf{X}=[\mathbf{x}_1,\mathbf{x}_2,\ldots,\mathbf{x}_n]\in\mathbb{R}^{d\times n}$ and  the low-dimensional representations as $\mathbf{H}=[\mathbf{h}_1,\mathbf{h}_2,\ldots,\mathbf{h}_n]\in\mathbb{R}^{k\times n}$, with $\mathbf{h}_i$ corresponding to the low-dimensional representation of $\mathbf{x}_i$. In this paper, we set the dimension $k$ of low-dimensional representation $\mathbf{h}_i$ as the total number of classes for dataset $\mathcal{X}$. For notation convenience, we encode the pairwise constraints into two matrices, $\mathbf{M}=(m_{ij})$ and $\mathbf{C}=(c_{ij})$, with 
  \begin{equation}
 \label{eq:construct M and C}
     m_{ij} = \begin{cases} 1,   &\text{if} ~(\mathbf{x}_i,\mathbf{x}_j)\in \mathcal{M}\\
      0, &\text{otherwise}
     \end{cases},~ c_{ij} = \begin{cases} \frac{1}{n_c},   &\text{if} ~(\mathbf{x}_i,\mathbf{x}_j)\in \mathcal{C}\\
      0, &\text{otherwise}
     \end{cases}
 \end{equation}
 where $n_c$ is the total number of cannot-link constraints in $\mathcal{C}$.
 We use the affinity matrix $\mathbf{S}=(s_{ij})$ (which will be defined in (\ref{eq:construct S})) and the matrices, $\mathbf{M}$ and $\mathbf{C}$, to  learn the low-dimensional representations  $\mathbf{H}$ by solving the constrained optimization problem
\begin{equation}
\label{eq:trace raito2}
 \underset{\mathbf{HH}^\top = \mathbf{I}}{\arg \min}~ \frac{  \sum \limits_{i,j=1}^n s_{ij} \|\mathbf{h}_{i}-\mathbf{h}_{j}\|^2 + \lambda_M\sum \limits_{i,j=1}^n m_{ij}\|\mathbf{h}_i- \mathbf{h}_j\|^2 }{ \sum \limits_{i,j=1}^n c_{ij}\|\mathbf{h}_i - \mathbf{h}_{j}\|^2},
\end{equation}
where $\lambda_M>0$ is a tradeoff parameter and $s_{ij}\geq0$ is the affinity value between $\mathbf{x}_i$ and $\mathbf{x}_j$. We enforce the pair of data points with must-link (resp. cannot-link) constraint to have similar (resp. different) low-dimensional representations by minimizing $\sum _{i,j=1}^n m_{ij}\|\mathbf{h}_i- \mathbf{h}_j\|^2$ and maximizing $\sum _{i,j=1}^n c_{ij}\|\mathbf{h}_i- \mathbf{h}_j\|^2$ simultaneously in (\ref{eq:trace raito2}). Moreover, to exploit the information of all data points, we also minimize the  term $\sum_{i,j=1}^n s_{ij} \|\mathbf{h}_{i}-\mathbf{h}_{j}\|^2$ in (\ref{eq:trace raito2}). The
 role of minimizing this term is twofold. When the low-dimensional representations $\mathbf{H}$ is fixed, minimizing this term enforces $s_{ij}$ equal to 0 whenever $\|\mathbf{h}_i-\mathbf{h}_j\|^2$ is relatively large. In other words, minimizing this term enforces data points with remarkably different low-dimensional representations to have an affinity value of 0. When the affinity matrix $\mathbf{S}$ is fixed, minimizing this term
enforces $\mathbf{h}_i=\mathbf{h}_j$ whenever $s_{ij}$ is relatively large.
  
  The construction of the affinity matrix $\mathbf{S}$ is essential to the learning of the low-dimensional representations $\mathbf{H}$. To fully exploit the pairwise relations among data points, we construct the affinity matrix $\mathbf{S}$ by integrating the local distance and global self-representation among data points. Specifically, the affinity matrix $\mathbf{W}=(w_{ij})\in\mathbb{R}^{n\times n}$ based on the Euclidean distance is constructed by
 \begin{equation}
 \label{eq:construct W}
     w_{ij} = \begin{cases} \exp\left(-\frac{\|\mathbf{x}_i-\mathbf{x}_j\|^2}{\sigma^2_i}\right),   &\text{if} ~\mathbf{x}_j\in \mathcal{N}_i\\
      0, &\text{otherwise}
     \end{cases}
 \end{equation}
where $\sigma_i$ is defined as the distance from $\mathbf{x}_i$ to its $l$-th nearest neighbor and $\mathcal{N}_i$ is the set of the $m$ nearest neighbors of $\mathbf{x}_i$. The  affinity matrix  $|\mathbf{Z}|\in \mathbb{R}^{n\times n}$ based on the global self-representation model is constructed by solving the problem
\begin{equation}
\label{eq:ssc}
\begin{aligned}
    &\min_{\mathbf{Z}} ~  \frac{1}{2} \|\mathbf{X}-\mathbf{XZ}\|^2 + \lambda_Z \|\mathbf{Z}\|_1, ~\text{s.t.} ~\text{diag}(\mathbf{Z})=\mathbf{0},
    \end{aligned}
\end{equation}
where $\lambda_Z>0$ is a tradeoff parameter. We apply the Frobenius norm on the reconstruction loss to alleviate the noise effect. $\ell_1$-norm is applied to enforce the sparsity of the self-representation matrix $\mathbf{Z}$.
Then we construct the affinity matrix $\mathbf{S}$ as the weighted sum of $\mathbf{W}$ and $|\mathbf{Z}|$
\begin{equation}
\label{eq:construct S}
\mathbf{S} = \alpha_1 |\mathbf{Z}| + \alpha_2 \mathbf{W},
\end{equation} 
where $\alpha_1>0$ and $\alpha_2>0$. 

\begin{proposition}
\label{prop:laplacian equation}
\textit{Given any matrix $\mathbf{S}\in\mathbb{R}^{n\times n}$, we define $\mathbf{L}_{\mathbf{S}}$ as the Laplacian matrix of $\frac{|\mathbf{S}|+|\mathbf{S}|^\top}{2}$, i.e., $\mathbf{L}_{\mathbf{S}}=\mathbf{D}_{\mathbf{S}}-\frac{|\mathbf{S}|+|\mathbf{S}|^\top}{2}$, where $\mathbf{D}_{\mathbf{S}}$ is a diagonal matrix with the $i$-th diagonal element being $\sum_{j=1}^n \frac{|s_{ij}|+|s_{ji}|}{2}$. Then we can obtain the following equation}
\begin{equation}
    \text{Tr}\left(\mathbf{H}\mathbf{L}_{\mathbf{S}}\mathbf{H}^\top\right)=\frac{1}{2}\sum_{i,j=1}^n|s_{ij}|\|\mathbf{h}_i-\mathbf{h}_j\|^2.
\end{equation}
\end{proposition}

The proof of Proposition \ref{prop:laplacian equation} can be found in the Appendix. Then by integrating (\ref{eq:trace raito2}),  (\ref{eq:ssc}), (\ref{eq:construct S}) and Proposition \ref{prop:laplacian equation}, we formulate a unified optimization framework to learn the affinity matrix $|\mathbf{Z}|$ and the low-dimensional representations $\mathbf{H}$ simultaneously
\begin{equation}
\label{optim:SSC-TR-bridge-v0}
\begin{aligned}
\min_{\mathbf{Z},\mathbf{H}} ~ &\frac{1}{2} \|\mathbf{X}-\mathbf{XZ}\|^2 + \lambda_Z \|\mathbf{Z}\|_1  +  \frac{\alpha_1 \text{Tr}(\mathbf{HL}_\mathbf{Z}\mathbf{H}^\top)} {\text{Tr}(\mathbf{HL}_\mathbf{C}\mathbf{H}^\top)} \\
 &\liang{}{\qquad}+   \frac{ \alpha_2 \text{Tr}\left(\mathbf{H}\left(\mathbf{L}_\mathbf{W}+\lambda_M\mathbf{L}_\mathbf{M}\right)\mathbf{H}^\top\right)}{ \text{Tr}(\mathbf{HL}_\mathbf{C}\mathbf{H}^\top)}, \\
 & \text{s.t.}~ \text{diag}(\mathbf{Z})=\mathbf{0}, ~ \mathbf{HH}^\top = \mathbf{I},
\end{aligned}
\end{equation}
where $\lambda_Z>0$, $\alpha_1>0$, $\alpha_2>0$ and $\lambda_M>0$ are tradeoff parameters.
Introducing an auxiliary variable $\mathbf{A}$, we consider the penalized problem of \eqref{optim:SSC-TR-bridge-v0}:
\begin{equation}
\label{optim:SSC-TR-bridge-v1}
\begin{aligned}
&\min_{\mathbf{A},\mathbf{Z},\mathbf{H}} ~ \frac{1}{2} \|\mathbf{X}-\mathbf{XA}\|^2 + \frac{\lambda}{2}\|\mathbf{A}-\mathbf{Z}\|^2 + \lambda_Z \|\mathbf{Z}\|_1  \\
 &+  \frac{\alpha_1 \text{Tr}(\mathbf{HL}_\mathbf{Z}\mathbf{H}^\top)} {\text{Tr}(\mathbf{HL}_\mathbf{C}\mathbf{H}^\top)} +   \frac{ \alpha_2 \text{Tr}\left(\mathbf{H}\left(\mathbf{L}_\mathbf{W}+\lambda_M\mathbf{L}_\mathbf{M}\right)\mathbf{H}^\top\right)}{ \text{Tr}(\mathbf{HL}_\mathbf{C}\mathbf{H}^\top)}, \\
 & \text{s.t.}~ \text{diag}(\mathbf{Z})=\mathbf{0}, ~ \mathbf{HH}^\top = \mathbf{I},
\end{aligned}
\end{equation}
where $\lambda>0$ is a parameter with a relatively large value. As will be seen in Section \ref{section:model optimization}, the term $\frac{\lambda}{2}\|\mathbf{A}-\mathbf{Z}\|^2$ makes the subproblems for updating $\mathbf{A}$ and $\mathbf{Z}$ strongly convex and thus the solutions are unique and stable. This is also beneficial to the convergence analysis.

\subsection{Numerical Algorithm}
\label{section:model optimization}
 In this section, we show how to solve the nonconvex problem (\ref{optim:SSC-TR-bridge-v1}). There are three blocks of variables in problem (\ref{optim:SSC-TR-bridge-v1}) and  we adopt the alternating minimization method that cyclically updates $\{\mathbf{A}\}$, $\{\mathbf{Z}\}$, $\{\mathbf{H}\}$.

$\bullet$ First, fix $\mathbf{A}=\mathbf{A}^k,\mathbf{Z}=\mathbf{Z}^k$, and update $\mathbf{H}^{k+1}$ by 
\begin{equation}
    \label{eq:update H}
    \mathbf{H}^{k+1} = \underset{\mathbf{HH}^\top = \mathbf{I}}{\arg \min}~ \frac{\text{Tr}\left(\mathbf{HL}_{\widetilde{\mathbf{W}}}\mathbf{H}^\top\right)}{\text{Tr}\left(\mathbf{HL}_\mathbf{C}\mathbf{H}^\top\right)},
\end{equation}
where 
\begin{equation}
    \label{eq:fused affinity}
    \begin{aligned}
    \widetilde{\mathbf{W}} &= \alpha_1 |\mathbf{Z}| + \alpha_2(\mathbf{W}+\lambda_M\mathbf{M}), \\
    \mathbf{L}_{\widetilde{\mathbf{W}}} &= \alpha_1 \mathbf{L}_\mathbf{Z} + \alpha_2 \left(\mathbf{L}_\mathbf{W}+\lambda_M\mathbf{L}_\mathbf{M}\right).
    \end{aligned}
\end{equation}
For (\ref{eq:update H}), it is equivalent to 
\begin{equation}
    \label{eq:update H max}
    \mathbf{H}^{k+1} = \underset{\mathbf{HH}^\top = \mathbf{I}}{\arg \max}~ \frac{\text{Tr}\left(\mathbf{HL}_\mathbf{C}\mathbf{H}^\top\right)}{\text{Tr}\left(\mathbf{HL}_{\widetilde{\mathbf{W}}}\mathbf{H}^\top\right)}.
\end{equation}
The trace ratio problem (\ref{eq:update H max}) has been efficiently solved by the ALGORITHM 4.1 in \cite{ngo2012trace}, which is also shown in Algorithm \ref{alg:update H} in this paper. 
\begin{algorithm}[!t]
    \caption{Update $\mathbf{H}$ by Solving (\ref{eq:update H max}) or (\ref{eq:update Hnew})}
    \label{alg:update H}
    \begin{algorithmic}
    \State \textbf{Input:} $\mathbf{B}=\mathbf{L}_{\mathbf{C}}$, $\mathbf{E}=\mathbf{L}_{\widetilde{\mathbf{W}}}$ for (\ref{eq:update H max}) or $\mathbf{E}=\mathbf{N}_{\widetilde{\mathbf{W}}}$ for (\ref{eq:update Hnew}), dimension $k$ of low-dimensional representations $\mathbf{H}$.
    \State \textbf{Initialize:} $\mathbf{H}\in \mathbb{R}^{k\times n}$, $\mathbf{HH^\top}=\mathbf{I}$, $\rho := \frac{\text{Tr}\left(\mathbf{H}\mathbf{B}\mathbf{H}^\top\right)}{\text{Tr}\left(\mathbf{H}{\mathbf{E}}\mathbf{H}^\top\right)}$.
    \While{not converged}
    \State 1) Obtain the $k$ largest eigenvalues of $\mathbf{B} - \rho \mathbf{E}$ and associated orthonormal eigenvectors $[\mathbf{v}_1,\mathbf{v}_2,\ldots,\mathbf{v}_k]^\top \equiv \mathbf{H}$;
    \State 2) Set $\rho := \frac{\text{Tr}\left(\mathbf{HB}\mathbf{H}^\top\right)}{\text{Tr}\left(\mathbf{HE}\mathbf{H}^\top\right)}$;
    \EndWhile
     \State \textbf{Output:} $\mathbf{H}$.
    \end{algorithmic}
\end{algorithm}

\begin{algorithm}[!t]
    \caption{Solve (\ref{optim:SSC-TR-bridge-v1}) by Alternating Minimization}
    \label{alg:DGSLv1}
    \begin{algorithmic}
    \State \textbf{Input:} Feature matrix $\mathbf{X}$, must-link matrix $\mathbf{M}$, cannot-link  matrix $\mathbf{C}$ and matrix $\mathbf{W}$, parameters $\lambda_Z$, $\alpha_1$, $\alpha_2$, $\lambda$ and $\lambda_M$.
    \State \textbf{Initialize:} $k=0$, $\mathbf{A=0}$, $\mathbf{Z=0}$, $\mathbf{H=0}$.
    \While{not converged}
    \State 1) Compute $\mathbf{H}^{k+1}$ by solving (\ref{eq:update H max}) using Algorithm \ref{alg:update H};
    \State 2) Compute $\mathbf{A}^{k+1}$ according to (\ref{solutoin:A});
    \State 3) Compute $\mathbf{Z}^{k+1}$ according to (\ref{solution:Z});
    \State 4) k=k+1;
    \EndWhile
     \State \textbf{Output:} $\mathbf{A}=\mathbf{A}^k$, $\mathbf{Z}=\mathbf{Z}^k$ and $\mathbf{H}=\mathbf{H}^k$.
    \end{algorithmic}
\end{algorithm}

$\bullet$ Second, fix $\mathbf{H}=\mathbf{H}^{k+1}$, $\mathbf{Z}=\mathbf{Z}^k$, and update $\mathbf{A}^{k+1}$ by 
\begin{equation}
\label{eq:update A}
\mathbf{A}^{k+1} = \underset{\mathbf{A}}{\arg \min} ~ \frac{1}{2} \|\mathbf{X}-\mathbf{XA}\|^2 + \frac{\lambda}{2} \|\mathbf{A}-\mathbf{Z}\|^2.
\end{equation}
Since the  objective function in (\ref{eq:update A}) is smooth and strongly convex, the closed form solution of $\mathbf{A}^{k+1}$ can be obtained by
\begin{equation}
    \label{solutoin:A}
    \mathbf{A}^{k+1} = \left(\mathbf{X}^\top\mathbf{X} + \lambda \mathbf{I}\right)^{-1}\left(\mathbf{X}^\top\mathbf{X}+\lambda \mathbf{Z}\right).
\end{equation}

$\bullet$ Third, fix $\mathbf{H}=\mathbf{H}^{k+1}$,  $\mathbf{A}=\mathbf{A}^{k+1}$,  and update $\mathbf{Z}^{k+1}$ by 
\begin{equation}
\label{eq:update Z}
\begin{aligned}
\mathbf{Z}^{k+1} = \underset{\mathbf{Z}}{\arg \min} ~ &\frac{\alpha_1 \text{Tr}(\mathbf{HL}_\mathbf{Z} \mathbf{H}^\top)} { \text{Tr}(\mathbf{HL}_\mathbf{C}\mathbf{H}^\top)} + \frac{\lambda}{2} \|\mathbf{Z}-\mathbf{A}\|^2 + \lambda_Z \|\mathbf{Z}\|_1 ,\\
  \text{s.t.}~ &\text{diag}(\mathbf{Z}) = \mathbf{0}.
\end{aligned}
\end{equation}
\begin{algorithm}[!t]
    \caption{Solve (\ref{optim:SSC-TR-bridge-v1}) by Alternating Minimization and Normalization Operations}
    \label{alg:DGSLv1 with normalization operation}
    \begin{algorithmic}
    \State \textbf{Input:} Feature matrix $\mathbf{X}$, must-link matrix $\mathbf{M}$, cannot-link  matrix $\mathbf{C}$ and matrix $\mathbf{W}$, parameters $\lambda_Z$, $\alpha_1$, $\alpha_2$, $\lambda$ and $\lambda_M$.
    \State \textbf{Initialize}: $k=0$, $\mathbf{A=0}$, $\mathbf{Z=0}$, $\mathbf{H=0}$.
    \While{not converged}
    \State 1) Obtain $\widetilde{\mathbf{W}}$ by (\ref{eq:refine tildeW});
    \State 2) Compute $\mathbf{H}^{k+1}$ by solving (\ref{eq:update Hnew}) using Algorithm \ref{alg:update H};
    \State 3) Compute $\mathbf{A}^{k+1}$ according to (\ref{solutoin:A});
    \State 4) Obtain $\boldsymbol{\Theta}$ by (\ref{eq:threshold});
    \State 5) Compute $\mathbf{Z}^{k+1}$ according to (\ref{solution:Z});
    \State 6) k=k+1;
    \EndWhile
    \State \textbf{Output:} $\mathbf{A}=\mathbf{A}^k$, $\mathbf{Z}=\mathbf{Z}^k$ and $\mathbf{H}=\mathbf{H}^k$.
    \end{algorithmic}
\end{algorithm}
Since $\operatorname{Tr}\left(\mathbf{H L}_{\mathbf{Z}} \mathbf{H}^{\top}\right)=\frac{1}{2} \sum_{i, j=1}^{n}\left|z_{i j}\right|\left\|\mathbf{h}_{i}-\mathbf{h}_{j}\right\|^{2}$ from Proposition \ref{prop:laplacian equation} and $\|\mathbf{Z}\|_1:=\sum_{i,j=1}^n|z_{ij}|$, we can rewrite (\ref{eq:update Z}) as
\begin{equation}
\label{eq:update Z1}
\mathbf{Z}^{k+1} = \underset{\mathbf{Z}}{\arg\min} ~\frac{1}{2} \|\mathbf{Z}-\mathbf{A}\|^2 + \|\boldsymbol{\Theta} \odot \mathbf{Z}\|_1, ~\text{s.t.} ~\text{diag}(\mathbf{Z})=  \mathbf{0},
\end{equation}
where $\odot$ is the Hadamard product and $\boldsymbol{\Theta} = (\boldsymbol{\Theta}_{ij})$, with
\begin{equation}
\boldsymbol{\Theta}_{ij} = \frac{\alpha_1 \|\mathbf{h}_{i}-\mathbf{h}_{j}\|^2}{2\lambda \text{Tr}(\mathbf{HL}_\mathbf{C} \mathbf{H}^\top)} + \frac{\lambda_Z}{\lambda}.
\end{equation}
The closed form solution for (\ref{eq:update Z1}) can be calculated by
\begin{equation}
    \label{solution:Z}
    \mathbf{Z}^{k+1} = \hat{\mathbf{Z}}^{k+1} - \text{diag}(\hat{\mathbf{Z}}^{k+1}),
\end{equation}
where 
$
    \hat{\mathbf{Z}}^{k+1}_{ij} = \text{sgn}(\mathbf{A}_{ij}) \max(|\mathbf{A}_{ij}| - \boldsymbol{\Theta}_{ij},0)
$
and $\text{sgn}(a):=1$ (resp. 0, -1) if $a>0$ (resp. $a=0$, $a<0$). 
The whole procedure of the alternating minimization scheme for (\ref{optim:SSC-TR-bridge-v1}) is given in Algorithm \ref{alg:DGSLv1}. 

Furthermore, to boost the performance of our model, we adopt normalization operations during the updating procedure for $\mathbf{A,Z,H}$. Specifically, for the update of $\mathbf{H}^{k+1}$, we first construct the  matrix $\widetilde{\mathbf{W}}$ as 
\begin{equation}
\label{eq:refine tildeW}
    \widetilde{\mathbf{w}}_i = \alpha_1 \frac{|\mathbf{z}_i|}{\|\mathbf{z}_i\|_\infty} + \alpha_2(\mathbf{w}_i+\lambda_M\mathbf{m}_i),~i=1,\ldots,n,
\end{equation}
where $\widetilde{\mathbf{w}}_i$ is the $i$-th column of $\widetilde{\mathbf{W}}$.
Then we obtain $\mathbf{H}^{k+1}$ by solving the optimization problem
\begin{equation}
    \label{eq:update Hnew}
   \mathbf{H}^{k+1} = \underset{\mathbf{HH}^\top = \mathbf{I}}{\arg \max}~ \frac{\text{Tr}\left(\mathbf{HL}_\mathbf{C}\mathbf{H}^\top\right)}{\text{Tr}\left(\mathbf{HN}_{\widetilde{\mathbf{W}}}\mathbf{H}^\top\right)},
\end{equation}
where $\mathbf{N}_{\widetilde{\mathbf{W}}}$ is the normalization of the Laplacian matrix $\mathbf{L}_{\widetilde{\mathbf{W}}}$, i.e.,
$\mathbf{N}_{\widetilde{\mathbf{W}}} =\mathbf{D}_{\widetilde{\mathbf{W}}}^{-1/2} \mathbf{L}_{\widetilde{\mathbf{W}}}\mathbf{D}_{\widetilde{\mathbf{W}}}^{-1/2}$, and $\mathbf{D}_{\widetilde{\mathbf{W}}}$ is a diagonal matrix with its $i$-th diagonal element being $\sum_{j=1}^n\frac{\widetilde{\mathbf{W}}_{ij}+\widetilde{\mathbf{W}}_{ji}}{2}$. 
For the update of $\mathbf{Z}^{k+1}$, we reconstruct the matrix $\boldsymbol{\Theta} = (\boldsymbol{\Theta}_{ij})$ as
\begin{equation}
\label{eq:threshold}
\boldsymbol{\Theta}_{ij} = \frac{\alpha_1 \left\|\frac{\mathbf{h}_{i}}{\left\|\mathbf{h}_{i}\right\|}-\frac{\mathbf{h}_{j}}{\left\|\mathbf{h}_{j}\right\|}\right\|^2}{2\lambda \text{Tr}(\mathbf{HL}_\mathbf{C} \mathbf{H}^\top)} + \frac{\lambda_Z}{\lambda}.
\end{equation}
The whole procedure of the alternating minimization scheme with normalization operations for (\ref{optim:SSC-TR-bridge-v1}) is given in Algorithm \ref{alg:DGSLv1 with normalization operation}.

We denote the objective function of (\ref{optim:SSC-TR-bridge-v1}) as $f(\mathbf{A,Z,H})$. Let $S_{1}=\{\mathbf{Z} \mid \text{diag}(\mathbf{Z})=\mathbf{0}\}$ and $S_{2}=\{\mathbf{H} \mid \mathbf{HH}^\top = \mathbf{I}\}$,  and we denote the indicator functions of $S_1$ and $S_2$ as $\iota_{S_{1}}(\mathbf{Z})$ and $\iota_{S_{2}}(\mathbf{H})$. Then we give the convergence guarantee for Algorithm \ref{alg:DGSLv1}.

\begin{proposition}
\label{main proposition}
\textit{The sequence $\left\{\mathbf{H}^{k}, \mathbf{A}^{k}, \mathbf{Z}^{k}\right\}$ generated by Algorithm \ref{alg:DGSLv1} has the following properties:}

\textit{(1) The objective $f\left(\mathbf{A}^{k}, \mathbf{Z}^{k}, \mathbf{H}^{k}\right)+\iota_{S_{1}}\left(\mathbf{Z}^{k}\right)+\iota_{S_{2}}\left(\mathbf{H}^{k}\right)$ is monotonically decreasing. Moreover,}
\begin{equation*}
\begin{aligned}
& f\left(\mathbf{A}^{k+1}, \mathbf{Z}^{k+1}, \mathbf{H}^{k+1}\right)+\iota_{S_{1}}\left(\mathbf{Z}^{k+1}\right)+\iota_{S_{2}}\left(\mathbf{H}^{k+1}\right) \\
\leq & f\left(\mathbf{A}^{k}, \mathbf{Z}^{k}, \mathbf{H}^{k}\right)+\iota_{S_{1}}\left(\mathbf{Z}^{k}\right)+\iota_{S_{2}}\left(\mathbf{H}^{k}\right) \\
&-\frac{\lambda}{2}\left\|\mathbf{A}^{k+1}-\mathbf{A}^{k}\right\|^{2}-\frac{\lambda}{2}\left\|\mathbf{Z}^{k+1}-\mathbf{Z}^{k}\right\|^{2};
\end{aligned}
\end{equation*}

\textit{(2) $\mathbf{A}^{k+1}-\mathbf{A}^{k} \rightarrow 0, \mathbf{Z}^{k+1}-\mathbf{Z}^{k} \rightarrow 0$;}

\textit{(3) The sequences $\{\mathbf{A}^k\}, \{\mathbf{Z}^k\}$ and $\{\mathbf{H}^k\}$ are bounded.}
\end{proposition}

\begin{theorem} 
\label{main theorem}\textit{The sequence $\{\mathbf{A}^k,\mathbf{Z}^k,\mathbf{H}^{k}\}$ generated by Algorithm \ref{alg:DGSLv1} has at least one limit point and any limit point $(\mathbf{A}^*,\mathbf{Z}^*,\mathbf{H}^*)$ of $\{\mathbf{A}^k,\mathbf{Z}^k,\mathbf{H}^{k}\}$ is  a stationary point of (\ref{optim:SSC-TR-bridge-v1}).}
\end{theorem}

Please refer to the Appendix for the proofs of Theorem \ref{main theorem} and Proposition \ref{main proposition}.

\textbf{Remark.} For Algorithm \ref{alg:DGSLv1}, the closed form solutions for $\mathbf{A}^{k+1}$ and $\mathbf{Z}^{k+1}$ can be obtained by (\ref{solutoin:A}) and (\ref{solution:Z}). For the update of $\mathbf{H}^{k+1}$, the ALGORITHM 4.1 in \cite{ngo2012trace} has been proven in \cite{zhang2010fast} to converge globally to the optimal solution of (\ref{eq:update H max}). 

\subsection{Complexity Analysis}
Three sub-problems are included in Algorithm \ref{alg:DGSLv1 with normalization operation}. The computation complexity of updating $\mathbf{H}$ is O($\eta n^3$), where $\eta$ is the maximum number of spectral decompositions to calculate $\mathbf{H}$. The computation complexity of updating $\mathbf{A}$ is O($n^3$). The computation complexity of updating $\mathbf{Z}$ is O($kn^2$). The computation complexity for the normalization operations is O($n^2$). Thus, the overall computation complexity of Algorithm $\ref{alg:DGSLv1 with normalization operation}$ is O($T\eta n^3$), where $T$ is the maximum number of iterations of alternating minimization. We set $\eta=20$ and $T=50$ in this paper. 

In comparison, the compared methods  CSP \cite{wang2014constrained},  SL \cite{kamvar2003spectral} and LSGR \cite{yang2014unified} have computational complexity O($n^3$). CSCAP \cite{lu2008constrained} has computation complexity O($n_c n^2$+$n^3$), where $n_c$ is the number of cannot-link constraints in $\mathcal{C}$. NNLRS \cite{zhuang2012non}, S$^2$LRR \cite{li2015learning}, S$^3$R \cite{li2015learning} and NNLRR \cite{fang2015robust} have computational complexity O($Tn^3$) while DCSSC \cite{wang2018unified} has computational complexity O($T(dn^2+n^3)$), where $T$ is the total number of iterations.

\begin{figure}[!t]
  \centering
   \subfloat[]{\includegraphics[width=3.2in]{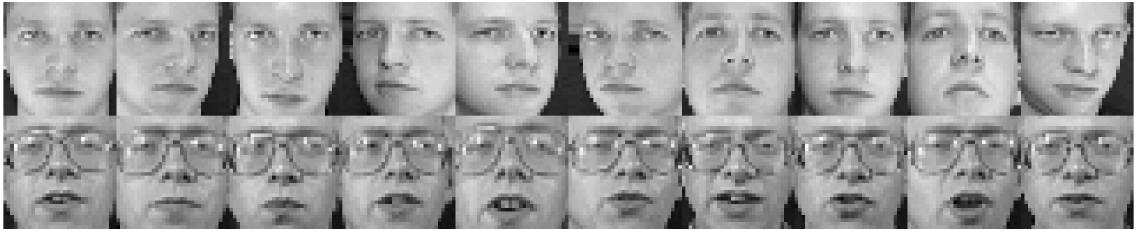}
  }
  \\
  \subfloat[]{\includegraphics[width=3.2in]{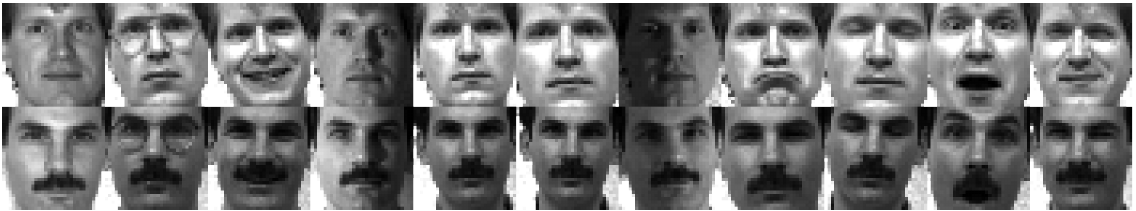}}
\\
\subfloat[]{\includegraphics[width=3.2in]{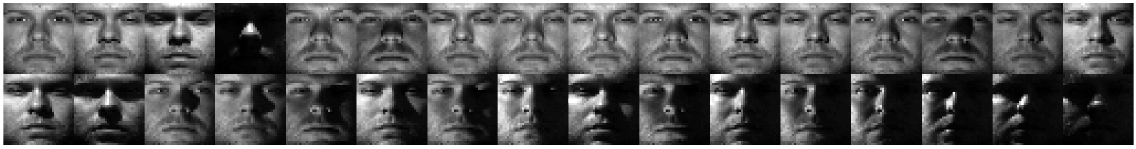}}
\\
\subfloat[]{\includegraphics[width=3.2in]{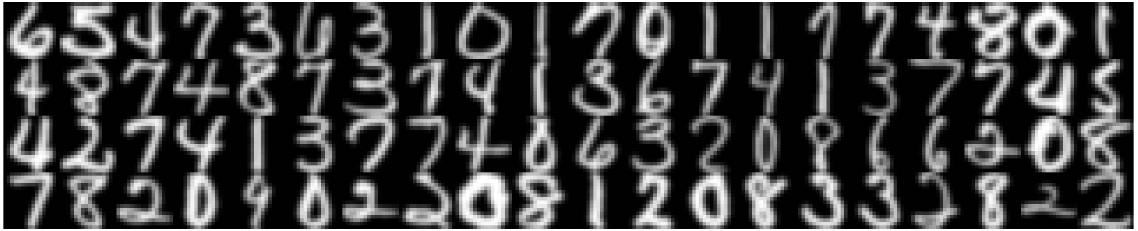}}
\\
\subfloat[]{\includegraphics[width=1.65in]{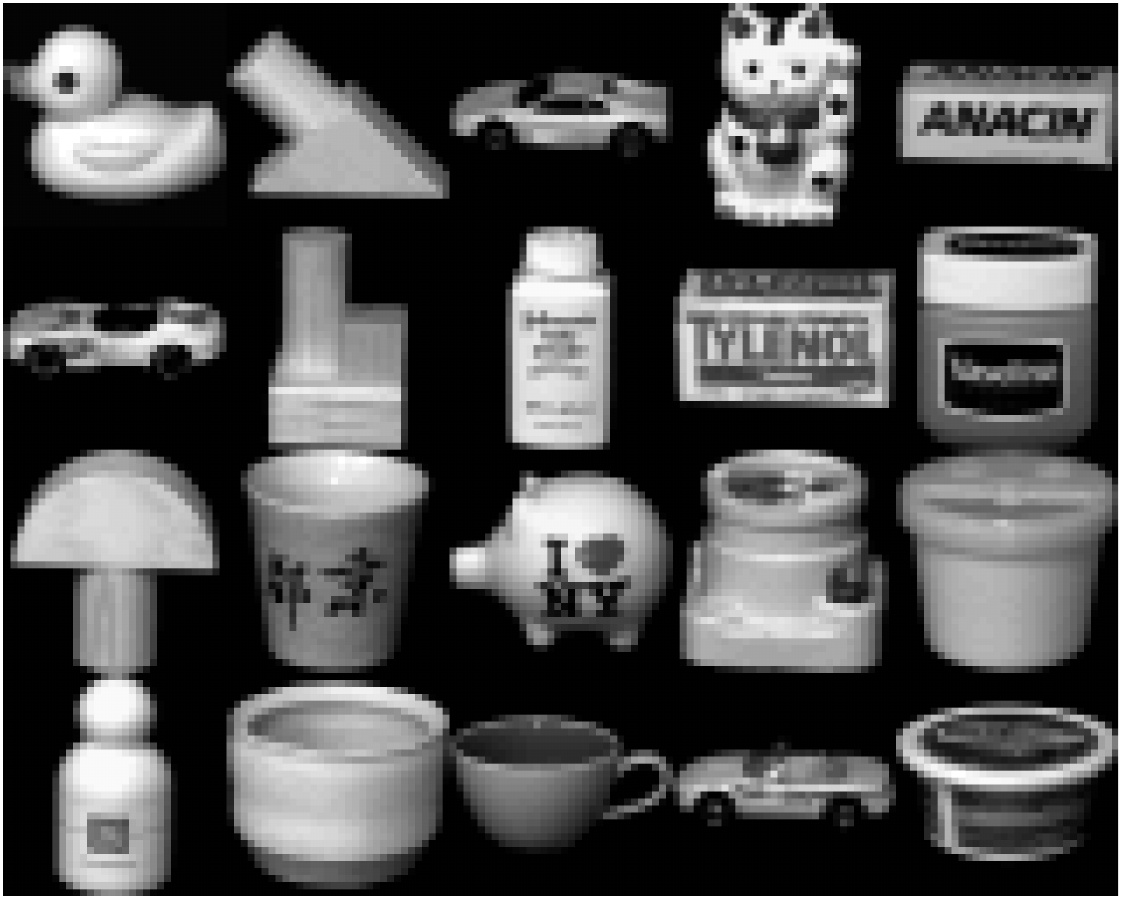}}
\hfil
\subfloat[]{\includegraphics[width=1.8in]{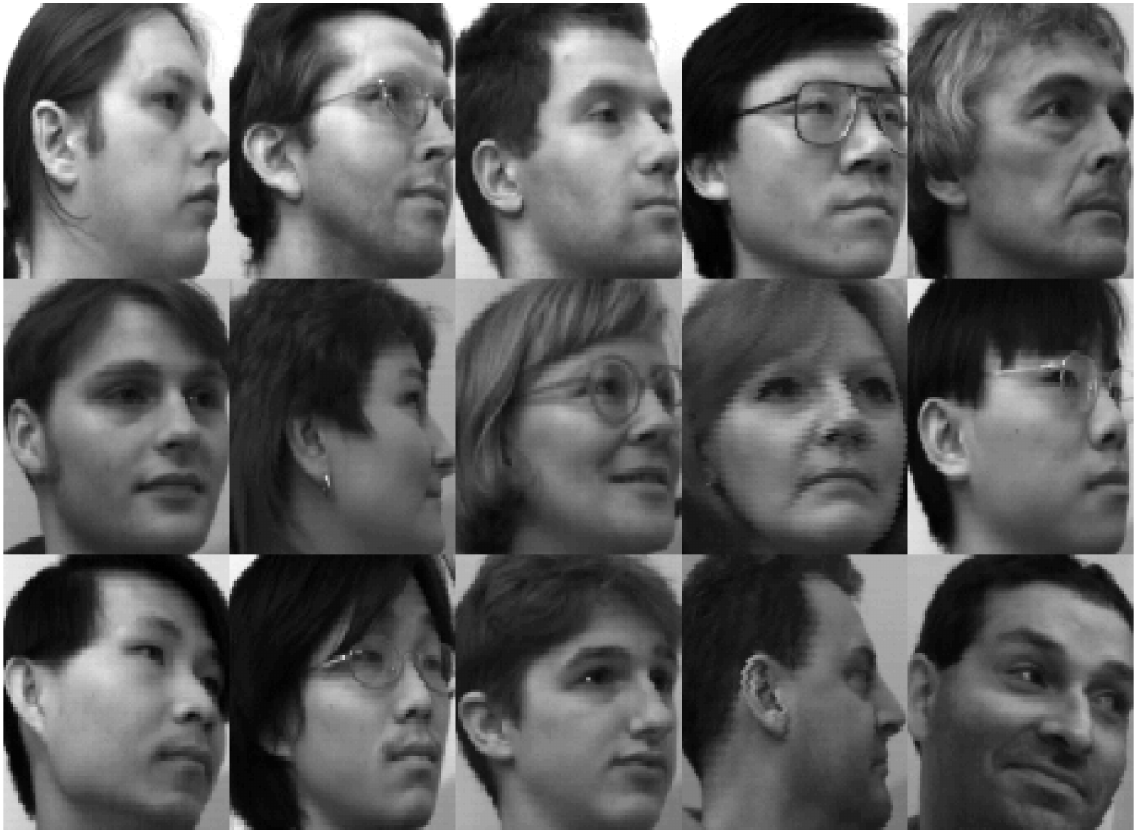}}
\caption{Sample images of the datasets. (a) ORL. (b) Yale. (c) Extended Yale B. (d) USPS. (e) COIL20. (f) UMIST.}
\label{fig:Visualization}
\end{figure}

\section{Experiments}
In this section, we evaluate our proposed Dynamic Graph Structure Learning (DGSL) method by comparing the clustering performance of DGSL with state-of-the-art semi-supervised graph-based clustering methods on eight benchmark datasets.
\subsection{Experimental Settings}
\subsubsection{Datasets}
Eight benchmark datasets including face images, handwritten digit images, object images and spoken letters are used in the experiments, i.e., Extended Yale B \cite{cai2007spectral}\footnote{\url{http://www.cad.zju.edu.cn/home/dengcai/Data/FaceData.html}}, Yale \cite{he2005face}\footnote{\url{http://www.cad.zju.edu.cn/home/dengcai/Data/FaceData.html}}, COIL20\footnote{\url{https://www.cs.columbia.edu/CAVE/software/softlib/coil-20.php}}, UMIST\footnote{\url{https://cs.nyu.edu/~roweis/data.html}}, Isolet \cite{cai2011speed}\footnote{\url{http://www.cad.zju.edu.cn/home/dengcai/Data/MLData.html}}, ORL \cite{cai2006orthogonal}\footnote{\url{http://www.cad.zju.edu.cn/home/dengcai/Data/FaceData.html}},  MNIST\footnote{\url{http://yann.lecun.com/exdb/mnist/}} and USPS\footnote{\url{https://www.csie.ntu.edu.tw/~cjlin/libsvmtools/datasets/multiclass.html#usps}}. Table \ref{tab:description for the data} shows the important statistics of the datasets, and Fig. \ref{fig:Visualization} shows sample images in these datasets.
\begin{itemize}
\item \textbf{ORL}
contains 400 grayscale images of 40 individuals. The images are taken with different lighting conditions, facial expressions, and facial details.
Each image is resized to 32$\times$32 pixels.
\item \textbf{Yale}
 contains 165 grayscale images of 15 individuals. Images for each individual are taken with different facial expressions or configurations.
 Each image is resized to 32$\times$32 pixels.
\item \textbf{MNIST} contains grayscale images of handwritten digits~$0\sim 9$. Each image is of size 28$\times$28. We randomly select 1000 images, with 100 images per digit.
\item \textbf{USPS}
contains 9,298  handwritten digit images. Each image is of size 16$\times$16. We randomly select 1000 images, with 100 images per digit.

\item \textbf{Extended Yale B}  consists of 2,414 frontal face images of 38 individuals. Each image is downsampled to 32 $\times$ 32 pixels. We follow the setting in \cite{wang2018unified}: the first 18 persons with 1,134 images are used in our experiments. 
\item \textbf{COIL20} consists of 1,440 images of 20 objects. The images of each object are taken five degrees apart as the object is rotated on a turntable. Each image is resized to 32$\times$32 pixels.
\item \textbf{UMIST} contains 575 total images of 20 individuals. 
Each image is cropped into size $112\times 92$.
\item \textbf{Isolet} contains 150 speakers who spoke twice the name of each letter of the alphabet. The speakers are grouped into five sets, with 30 speakers in each set, denoted as isolet1 to isolet5.
\end{itemize}
\begin{table}[H]
\caption{Statistics of the Eight Datasets}
\label{tab:description for the data}
\centering
\setlength\tabcolsep{4pt} 
\begin{tabular}{c c c c c}
\toprule
Datasets & $\#$Samples & $\#$Dimensions & $\#$Classes & Type \\
\midrule
Yale & 165& 1,024 & 15& face \\
ORL & 400 &1,024 & 40 & face \\
UMIST & 575 & 10,304 & 20 & face \\
MNIST & 70,000 &784 & 10 & digit \\
USPS & 9,298 &256 & 10 & digit \\
COIL20 & 1,440 &1,024 & 20 &object \\
Isolet & 7,797 & 617 & 26 & speech \\
Extended Yale B & 2,414 & 1,024 & 38 & face\\
\bottomrule
\end{tabular}
\end{table}

\subsubsection{Compared Methods}
We compare the proposed DGSL with four state-of-the-art semi-supervised graph-based clustering methods that use pairwise constraints as supervision, i.e., SL \cite{kamvar2003spectral}, LSGR \cite{yang2014unified}, 
CSCAP \cite{lu2008constrained},
CSP \cite{wang2014constrained}, and five state-of-the-art semi-supervised subspace clustering methods that use partial labels as supervision, i.e., NNLRS \cite{zhuang2012non}, NNLRR \cite{fang2015robust}, S$^3$R \cite{li2015learning}, S$^2$LRR \cite{li2015learning}, DCSSC \cite{wang2018unified}. 

\begin{table*}[!t]
\caption{Clustering Performance (ACC$\%$ $\pm$ STD$\%$)  on Different Datasets}
\label{tab:mode 1 ACC}
\centering
\begin{tabular}{c ccc ccc ccc}
    \toprule
\multirow{2}{*}{ACC} 
        & \multicolumn{3}{c}{ORL} & \multicolumn{3}{c}{Yale} & \multicolumn{3}{c}{COIL20}   \\
    \cmidrule(lr){2-4} \cmidrule(lr){5-7} \cmidrule(lr){8-10}
        & 2  & 3 & 4  & 2 & 3  & 4         &2 & 6 & 10\\
    \midrule
    SL \cite{kamvar2003spectral}    & 66.8 $\pm$ 2.9 & 74.2 $\pm$ 2.2 & 81.7 $\pm$ 2.7 &  47.5 $\pm$ 2.8 & 55.2 $\pm$ 3.1 & 65.6 $\pm$ 3.3  & 71.0 $\pm$ 3.1 & 72.8 $\pm$ 3.3  & 76.2 $\pm$ 3.2     \\
     CSP \cite{wang2014constrained}   & 78.4 $\pm$ 1.7 & 83.4 $\pm$ 5.1 & 87.5 $\pm$ 6.4  & 59.2 $\pm$ 1.8 & 61.4 $\pm$ 3.9 & 65.4 $\pm$ 3.4  & 80.0 $\pm$ 4.7 & 83.4 $\pm$ 5.1 & 87.7 $\pm$ 6.1    \\
     CSCAP \cite{lu2008constrained}   & 82.3 $\pm$ 2.0 & 89.5 $\pm$ 2.4 & 92.3 $\pm$ 2.2 & 61.4 $\pm$ 3.3 & 65.4 $\pm$ 3.6 & 68.7 $\pm$ 3.0  & 70.9 $\pm$ 2.7 & 74.3 $\pm$ 2.9 & 76.0 $\pm$ 2.6          \\
     LSGR \cite{yang2014unified}   & 81.4 $\pm$ 2.8 & 86.4 $\pm$ 2.3 & 90.4 $\pm$ 1.9 &  63.2 $\pm$ 3.4 & 68.1 $\pm$ 2.9 & 73.6 $\pm$ 3.6 & 81.0 $\pm$ 1.4 & 82.4 $\pm$ 2.7 & 85.6 $\pm$ 1.0         \\
     DGSL   & \textbf{90.4} $\pm$ 2.0    & \textbf{94.5} $\pm$ 2.0   & \textbf{96.4} $\pm$ 1.2  & \textbf{64.8} $\pm$ 3.3   & \textbf{73.5} $\pm$ 2.6   & \textbf{80.0} $\pm$ 2.2  & \textbf{85.1} $\pm$ 0.7 & \textbf{93.4} $\pm$ 2.5 & \textbf{97.2} $\pm$ 1.1    \\
    \midrule
\multirow{2}{*}{ACC} 
        & \multicolumn{3}{c}{UMIST} & \multicolumn{3}{c}{isolet1} & \multicolumn{3}{c}{isolet2}   \\
    \cmidrule(lr){2-4} \cmidrule(lr){5-7} \cmidrule(lr){8-10}
        & 2  & 3 & 4  & 5 & 7  & 10         &5 & 7 & 10\\
    \midrule
    SL \cite{kamvar2003spectral}    & 82.4 $\pm$ 3.6 & 83.6 $\pm$ 4.1 & 83.7 $\pm$ 6.0  & 66.6 $\pm$ 1.1    & 67.5 $\pm$ 1.2   &  68.5 $\pm$ 1.4 & 63.3 $\pm$ 0.7 & 63.9 $\pm$ 0.6  & 65.0 $\pm$ 0.5          \\
     CSP \cite{wang2014constrained}   & 84.3 $\pm$ 2.4 & 85.5 $\pm$ 1.9 & 85.8 $\pm$ 2.1   & 73.5 $\pm$ 2.7   & 74.9 $\pm$ 3.7    & 76.4 $\pm$ 4.5  & 70.0 $\pm$ 1.6 & 71.6 $\pm$ 2.0 & 72.8 $\pm$ 3.7             \\
     CSCAP \cite{lu2008constrained}   & 82.9 $\pm$   3.2 & 85.1 $\pm$ 4.0 & 87.2 $\pm$ 4.2   & 74.3 $\pm$ 1.6    & 75.4 $\pm$ 2.0   & 77.8 $\pm$ 1.4  & 67.8 $\pm$ 1.1 & 69.3 $\pm$ 2.0 & 70.5 $\pm$ 1.6         \\
     LSGR \cite{yang2014unified}   & 80.0 $\pm$ 1.5 & 81.8 $\pm$ 4.0 & 82.7 $\pm$ 5.9  & 69.7 $\pm$ 1.6   & 70.6 $\pm$ 1.5  & 72.7 $\pm$ 1.4  & 63.9 $\pm$ 0.9 & 65.0 $\pm$ 0.9 & 67.0 $\pm$ 1.2           \\
     DGSL   & \textbf{89.1} $\pm$ 2.6   & \textbf{92.5} $\pm$ 3.1   & \textbf{94.3} $\pm$ 1.8  & \textbf{82.9} $\pm$ 1.9    & \textbf{86.4} $\pm$ 1.6  & \textbf{89.1} $\pm$ 1.0  & \textbf{74.4} $\pm$ 2.5 & \textbf{79.8} $\pm$ 2.4 & \textbf{83.6} $\pm$ 1.9    \\
    \midrule
\multirow{2}{*}{ACC} 
        & \multicolumn{3}{c}{Extended Yale B} & \multicolumn{3}{c}{MNIST} & \multicolumn{3}{c}{USPS}   \\
    \cmidrule(lr){2-4} \cmidrule(lr){5-7} \cmidrule(lr){8-10}
        & 4  & 7 & 10  & 5 & 7  & 10         &5 & 7 & 10\\
    \midrule
    SL \cite{kamvar2003spectral}   &  87.0 $\pm$ 2.5 & 88.8 $\pm$ 2.2 & 90.3 $\pm$ 3.9  &  66.0 $\pm$ 3.9 & 69.8 $\pm$ 3.9 & 76.0 $\pm$ 5.6 & 84.3 $\pm$ 5.8 & 86.5 $\pm$ 5.7 & 87.6 $\pm$ 5.0    \\
     CSP \cite{wang2014constrained}   &  87.8 $\pm$ 4.3 & 88.3 $\pm$ 4.2  & 88.8 $\pm$ 3.3   & 79.0 $\pm$ 4.0  & 83.2 $\pm$ 1.8 &   85.0 $\pm$ 1.5 & 88.2 $\pm$ 0.9 & 89.4 $\pm$ 0.9 & 90.7 $\pm$ 0.8 \\
     CSCAP \cite{lu2008constrained}   & 88.0 $\pm$ 3.3 & 89.7 $\pm$ 3.4 & 90.7 $\pm$ 3.9 & 70.6 $\pm$ 5.0 & 72.7 $\pm$ 4.9 & 77.4 $\pm$ 4.6 & 86.1 $\pm$ 5.2 & 88.1 $\pm$ 4.5 & 89.8 $\pm$ 3.8   \\
     LSGR \cite{yang2014unified}   & 86.2 $\pm$ 3.3 & 88.2 $\pm$ 3.5 & 89.3 $\pm$ 2.5  & 67.1 $\pm$ 3.3 & 72.5 $\pm$ 5.5 & 76.6 $\pm$ 5.6 & 84.3 $\pm$ 5.6 & 85.5 $\pm$ 6.4 & 86.7 $\pm$ 5.4      \\
     DGSL   & \textbf{95.4} $\pm$ 0.8 & \textbf{95.6} $\pm$ 0.6 & \textbf{96.4}  $\pm$ 0.6 &  \textbf{83.7} $\pm$ 2.6 & \textbf{86.2} $\pm$ 1.8 & \textbf{88.6} $\pm$ 1.6 & \textbf{91.5} $\pm$ 0.8 & \textbf{92.3} $\pm$ 0.8 & \textbf{93.0} $\pm$ 0.9\\
    \bottomrule
    \end{tabular}
\end{table*}

\subsubsection{Evaluation Metrics}
For the clustering performance evaluation, we use two commonly used metrics \cite{li2017robust}, i.e., Accuracy (ACC) and Normalized Mutual Information  (NMI). For the above two metrics, a higher value implies a better clustering performance. 

\subsubsection{Implementation Details}
We first introduce how we decide the parameters in (\ref{optim:SSC-TR-bridge-v1}), i.e., $\alpha_1,\alpha_2,\lambda,\lambda_M$ and $\lambda_Z$. The parameter $\lambda$ is fixed as a relatively large number, i.e., $\lambda=100$. The parameter $\lambda_Z$ is tuned from $\{0,0.5,1.0,1.5,2.0\}$. For the parameter $\alpha_1$, we set $\alpha_1 = 2\tau\lambda\text{Tr}(\mathbf{H}^1\mathbf{L}_\mathbf{C}{\mathbf{H}^1}^\top)$,  where $\mathbf{H}^1$ is obtained by solving  (\ref{eq:update Hnew}) with $\widetilde{\mathbf{W}}=\mathbf{W}+\lambda_M\mathbf{M}$, and we tune $\tau$ from $\{0.01,0.02,0.03,0.05,0.075,0.1,0.2,0.3\}$. We set $\lambda_M=10$ for most of datasets except $\lambda_M=100$ for Extended Yale B, COIL20 and UMIST, $\lambda_M=1$ for Isolet. For the parameter $\alpha_2$, we set $\frac{\alpha_2}{\alpha_1}=0.2$ for $\lambda_M=1,10$ and $\frac{\alpha_2}{\alpha_1}=0.02$ for $\lambda_M=100$. Then we introduce how we construct the affinity matrix $\mathbf{W}$. We set $m=7,l=5$ for most of the datasets, except $m=51,l=21$ for isolet1 and isolet2, $m=5,l=3$ for UMIST, and $m=4,l=3$ for Extended Yale B. For the grayscale images with grayscale values distributed in $0\sim 255$, we normalize the grayscale values by dividing them by $255$. To obtain the final clustering results, we  conduct K-means on $\{\frac{\mathbf{h}_i}{\|\mathbf{h}_i\|}\}_{i=1}^n$ for the output $\mathbf{H}=[\mathbf{h}_1,\ldots,\mathbf{h}_n]$ of Algorithm \ref{alg:DGSLv1 with normalization operation}.

For the compared methods SL, CSP, LSGR, and CSCAP, one first needs to construct the affinity matrix. We construct the affinity matrix for MNIST, USPS, COIL20, isolet1, and isolet2 by (\ref{eq:construct W})  with the same setting of $m$ and $l$ as our approach. We construct the affinity matrix for ORL, Yale, UMIST, and Extended Yale B by first solving 
\begin{equation}
    \min_{\mathbf{A},\mathbf{Z}} ~ \frac{1}{2} \|\mathbf{X}-\mathbf{XA}\|^2 + \frac{\lambda}{2}\|\mathbf{A}-\mathbf{Z}\|^2 + \lambda_Z \|\mathbf{Z}\|_1 ~\text{s.t.}~\text{diag}(\mathbf{Z})=\mathbf{0}
\end{equation}
using alternating minimization, then we construct the $i$-th column of the affinity matrix as $\frac{|\mathbf{z}_i|}{\|\mathbf{z}_i\|_\infty}$, $i=1,\ldots,n$. For LSGR, we report the best performance among the three types of semi-supervised spectral clustering algorithms in \cite{yang2014unified}, i.e., $\mathcal{F}_{\text{ADJ}}$, $\mathcal{F}_{\text{LAP}}$ and $\mathcal{F}_{\text{NLAP}}$, which are based on the affinity matrix, Laplacian matrix and normalized Laplacian matrix, respectively. For CSP, we report the performance of the constrained spectral clustering algorithm for $K$-way partition in \cite{wang2014constrained}. For CSCAP, we report the performance of the constrained clustering algorithm for more than two classes in \cite{lu2008constrained}. We normalize the obtained low-dimensional representations of data points into unit Euclidean length before K-means clustering for the four compared methods that use pairwise constraints as supervisory information.

\begin{table*}[!t]
\caption{Clustering Performance (NMI$\%$ $\pm$ STD$\%$)  on Different Datasets}
\label{tab:mode 1 NMI}
\centering
\begin{tabular}{c ccc ccc ccc}
    \toprule
\multirow{2}{*}{NMI} 
        & \multicolumn{3}{c}{ORL} & \multicolumn{3}{c}{Yale} & \multicolumn{3}{c}{COIL20}   \\
    \cmidrule(lr){2-4} \cmidrule(lr){5-7} \cmidrule(lr){8-10}
        & 2 & 3  & 4  & 2 & 3  & 4          &2 & 6 &10\\
    \midrule
    SL \cite{kamvar2003spectral}    & 83.5 $\pm$ 1.2 & 87.6 $\pm$ 1.1 & 91.6 $\pm$ 1.3 & 53.7 $\pm$ 2.0 & 60.5 $\pm$ 2.7 & 68.9 $\pm$ 2.4   & 87.3 $\pm$ 1.6 & 87.7 $\pm$ 1.3 & 89.0 $\pm$ 1.3           \\
     CSP \cite{wang2014constrained}   & 89.5 $\pm$ 0.8 & 88.9 $\pm$ 3.5 &  91.7 $\pm$ 4.1 & 62.7 $\pm$ 1.8 & 64.4 $\pm$ 3.0 & 67.5 $\pm$ 2.9  & 89.1 $\pm$ 1.6 & 91.5 $\pm$ 1.6 &   92.3 $\pm$ 2.2        \\
     CSCAP \cite{lu2008constrained}   & 90.9 $\pm$ 0.7 & 94.1 $\pm$ 1.1 &  95.2 $\pm$ 1.3 &  63.3 $\pm$ 3.0 & 66.8 $\pm$ 2.6 & 70.1 $\pm$ 2.6 & 86.8 $\pm$ 1.3 & 87.8 $\pm$ 1.7 & 88.9 $\pm$ 1.4            \\
     LSGR \cite{yang2014unified}    & 90.8 $\pm$ 1.1 & 93.5 $\pm$ 0.9 & 95.6 $\pm$ 0.7 & \textbf{65.9} $\pm$ 2.5 & 70.4 $\pm$ 2.3 & 75.6 $\pm$ 2.4  & 92.0 $\pm$ 0.8 & 93.1 $\pm$ 0.9 & 94.3 $\pm$ 0.7            \\
     DGSL   & \textbf{94.1} $\pm$ 0.9   & \textbf{96.3} $\pm$ 1.1    & \textbf{97.4} $\pm$ 0.8    & 64.7 $\pm$ 2.8    & \textbf{71.2} $\pm$ 2.5   & \textbf{77.4} $\pm$ 2.1   & \textbf{94.7} $\pm$ 0.7 & \textbf{96.9} $\pm$ 0.7 & \textbf{97.4} $\pm$ 0.7          \\
   \midrule
\multirow{2}{*}{NMI} 
        & \multicolumn{3}{c}{UMIST} & \multicolumn{3}{c}{isolet1} & \multicolumn{3}{c}{isolet2}   \\
    \cmidrule(lr){2-4} \cmidrule(lr){5-7} \cmidrule(lr){8-10}
        & 2 & 3  & 4  & 5 & 7  & 10          &5 & 7 &10\\
    \midrule
    SL \cite{kamvar2003spectral}    &  87.5 $\pm$ 1.8 & 88.7 $\pm$ 1.8 & 89.7 $\pm$ 2.3  & 78.6 $\pm$ 0.3    & 79.1 $\pm$ 0.5   & 79.9 $\pm$ 0.5 & 75.5 $\pm$ 0.7 & 76.0 $\pm$ 0.5 & 76.9 $\pm$ 0.6          \\
     CSP \cite{wang2014constrained}   & 88.1 $\pm$ 1.3 & 88.7 $\pm$ 0.9 & 89.0 $\pm$ 1.2   & 82.3 $\pm$ 0.6   & 83.2 $\pm$ 0.7   & 84.2 $\pm$ 0.9 & 80.3 $\pm$ 0.6 & 81.3 $\pm$ 0.8 &   82.7 $\pm$ 0.7       \\
     CSCAP \cite{lu2008constrained}   & 87.9 $\pm$ 1.6 & 89.3 $\pm$ 1.6 & 91.0 $\pm$ 1.8    & 80.6 $\pm$ 0.7   & 81.9 $\pm$ 0.6    & 83.2 $\pm$ 0.9  & 78.0 $\pm$ 0.5 & 78.5 $\pm$ 0.6 & 79.3 $\pm$ 0.7        \\
     LSGR \cite{yang2014unified}    & 85.1 $\pm$ 0.8 & 88.5 $\pm$ 1.7 & 89.9 $\pm$ 2.1    & 79.0 $\pm$ 0.6   & 79.7 $\pm$ 0.7 & 81.0 $\pm$ 0.6  & 75.2 $\pm$ 0.5 & 76.0 $\pm$ 0.4 & 76.8 $\pm$ 0.4             \\
     DGSL   & \textbf{93.3} $\pm$ 1.0    & \textbf{94.9} $\pm$ 1.0    & \textbf{95.6} $\pm$ 1.0    & \textbf{85.6} $\pm$ 0.7    & \textbf{87.1} $\pm$ 0.9   & \textbf{88.7} $\pm$ 0.7   & \textbf{81.6} $\pm$ 1.0 & \textbf{83.5} $\pm$ 1.1 & \textbf{85.3} $\pm$ 0.9            \\
    \midrule
\multirow{2}{*}{NMI} 
        & \multicolumn{3}{c}{Extended Yale B} & \multicolumn{3}{c}{MNIST} & \multicolumn{3}{c}{USPS}   \\
    \cmidrule(lr){2-4} \cmidrule(lr){5-7} \cmidrule(lr){8-10}
        & 4 & 7  & 10  & 5 & 7  & 10          &5 & 7 &10\\
    \midrule
    SL \cite{kamvar2003spectral}    & 90.5 $\pm$ 0.7 & 91.2 $\pm$ 0.8 & 91.9 $\pm$ 1.1  & 66.7 $\pm$ 1.9 & 69.2 $\pm$ 1.6 & 72.9 $\pm$  2.3  &80.1 $\pm$ 2.1 & 81.6 $\pm$ 2.3 & 83.0 $\pm$ 1.7   \\
     CSP \cite{wang2014constrained}    & 90.3 $\pm$ 1.6 & 90.5 $\pm$ 1.8 & 91.0 $\pm$ 1.3   & 71.2 $\pm$ 2.5 & 73.0 $\pm$ 1.9 & 75.0 $\pm$ 1.8 & 79.9 $\pm$ 1.0 & 81.4 $\pm$ 1.1 & 83.3 $\pm$ 1,0\\
     CSCAP \cite{lu2008constrained}   &  89.9 $\pm$ 1.1 & 89.6 $\pm$ 1.0 &  91.6 $\pm$ 1.1 &  69.4 $\pm$ 2.6 & 70.4 $\pm$ 2.2 & 74.0 $\pm$ 2.7 & 80.8 $\pm$ 2.1 & 82.1 $\pm$ 2.1 & 84.0 $\pm$ 1.4    \\
     LSGR \cite{yang2014unified}   & 89.2 $\pm$ 1.1 & 90.2 $\pm$ 1.1 & 91.1 $\pm$ 1.0  &  67.4 $\pm$ 1.6 & 69.9 $\pm$ 2.2 & 73.6 $\pm$ 2.4 & 80.4 $\pm$ 2.2 & 81.7 $\pm$ 2.0 & 82.9 $\pm$ 1.8     \\
     DGSL   &  \textbf{93.4} $\pm$ 0.9 & \textbf{93.6} $\pm$ 0.7 &  \textbf{94.6} $\pm$ 0.7   & \textbf{76.6} $\pm$ 1.5 & \textbf{78.6} $\pm$ 1.6 & \textbf{80.6} $\pm$ 1.5 & \textbf{85.4} $\pm$ 0.8 & \textbf{86.3} $\pm$ 1.0 & \textbf{87.2} $\pm$ 1.0 \\
    \bottomrule
\end{tabular}
\end{table*}

\begin{table*}[!t]
\caption{Clustering Performance (ACC$\%$ $\pm$ STD$\%$)  on Different Datasets}
\label{tab:compare with methods using partial labels}
\centering
\begin{tabular}{c ccc ccc ccc}
    \toprule
\multirow{2}{*}{ACC} &  
        \multicolumn{3}{c}{COIL20} & \multicolumn{3}{c}{Extended Yale B} & \multicolumn{3}{c}{Yale}   \\
    \cmidrule(lr){2-4} \cmidrule(lr){5-7} \cmidrule(lr){8-10}
       & 2  & 6  & 10  & 4  & 7 & 10          & 2 & 3 & 4\\
    \midrule
    \multicolumn{10}{c}{Using Partial Labels as Supervision} \\
    \addlinespace
     NNLRS \cite{zhuang2012non}  & 74.9 $\pm$ 3.0   & 84.5 $\pm$ 2.8    & 89.4 $\pm$ 3.0    & 75.5 $\pm$ 2.6   & 84.8 $\pm$ 2.0    & 89.0 $\pm$ 3.0   & 56.7 $\pm$ 3.5 & 67.0 $\pm$ 4.1 & 71.8 $\pm$ 3.6          \\
    S$^2$LRR \cite{li2015learning} & 50.2 $\pm$ 6.9    & 80.0 $\pm$ 2.5    & 84.2 $\pm$ 2.3    & 93.3 $\pm$ 0.8   & 93.7 $\pm$ 0.7    & 95.5 $\pm$ 0.4  & 60.5 $\pm$ 4.2 & 72.5 $\pm$ 5.1 & 75.7 $\pm$ 4.8           \\
     S$^3$R \cite{li2015learning}   & 79.4 $\pm$ 4.0   & 88.3 $\pm$ 3.5    & 92.8 $\pm$ 0.3    & 87.8 $\pm$ 4.7   & 92.2 $\pm$ 2.4   & 92.7 $\pm$ 1.4   & 58.9 $\pm$ 6.6 & 68.5 $\pm$ 5.1 & 70.5 $\pm$ 4.4           \\
     NNLRR \cite{fang2015robust}   & 76.4 $\pm$ 4.7    & 88.5 $\pm$ 1.2   & 91.3 $\pm$ 1.2   & 72.0 $\pm$ 1.7   & 85.6 $\pm$ 2.0    & 88.0 $\pm$ 0.8    & 58.6 $\pm$ 3.1 & 67.8 $\pm$ 3.6 & 73.8 $\pm$ 3.4      \\
     DCSSC \cite{wang2018unified}  & 79.4 $\pm$ 3.6   & \textbf{94.3} $\pm$ 0.7   & 94.9 $\pm$ 0.2    & 93.2 $\pm$ 0.9   & 95.5 $\pm$ 0.9  & \textbf{97.1} $\pm$ 0.6    & \textbf{66.7} $\pm$ 2.3 & \textbf{74.6} $\pm$ 3.2 & \textbf{80.0} $\pm$ 1.3         \\
     
    \addlinespace
    \multicolumn{10}{c}{Using Pairwise Constraints as Supervision} \\
    \addlinespace
       DGSL & \textbf{85.1} $\pm$ 0.7   & 93.4 $\pm$ 2.5    & \textbf{97.2} $\pm$ 1.1    & \textbf{95.4} $\pm$ 0.8    & \textbf{95.6} $\pm$ 0.6   & 96.4 $\pm$ 0.6    & 64.8 $\pm$ 3.3 & 73.5 $\pm$ 2.6 & \textbf{80.0} $\pm$ 2.2       \\
    \bottomrule
\end{tabular}
    \end{table*}

\begin{figure*}[!t]
\centering
\subfloat[UMIST]{\includegraphics[width=1.8in]{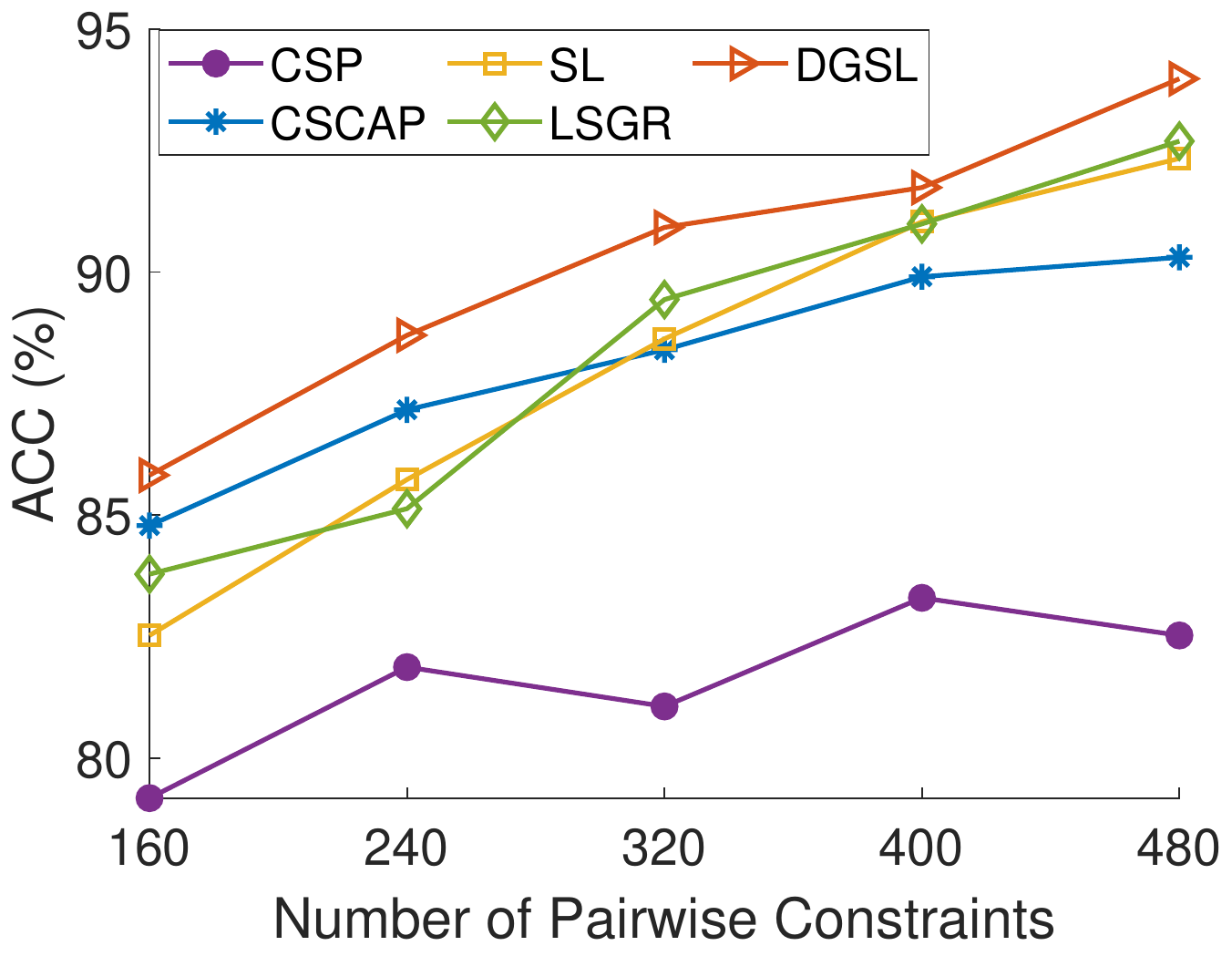}%
}
\hfil
\subfloat[UMIST]{\includegraphics[width=1.75in]{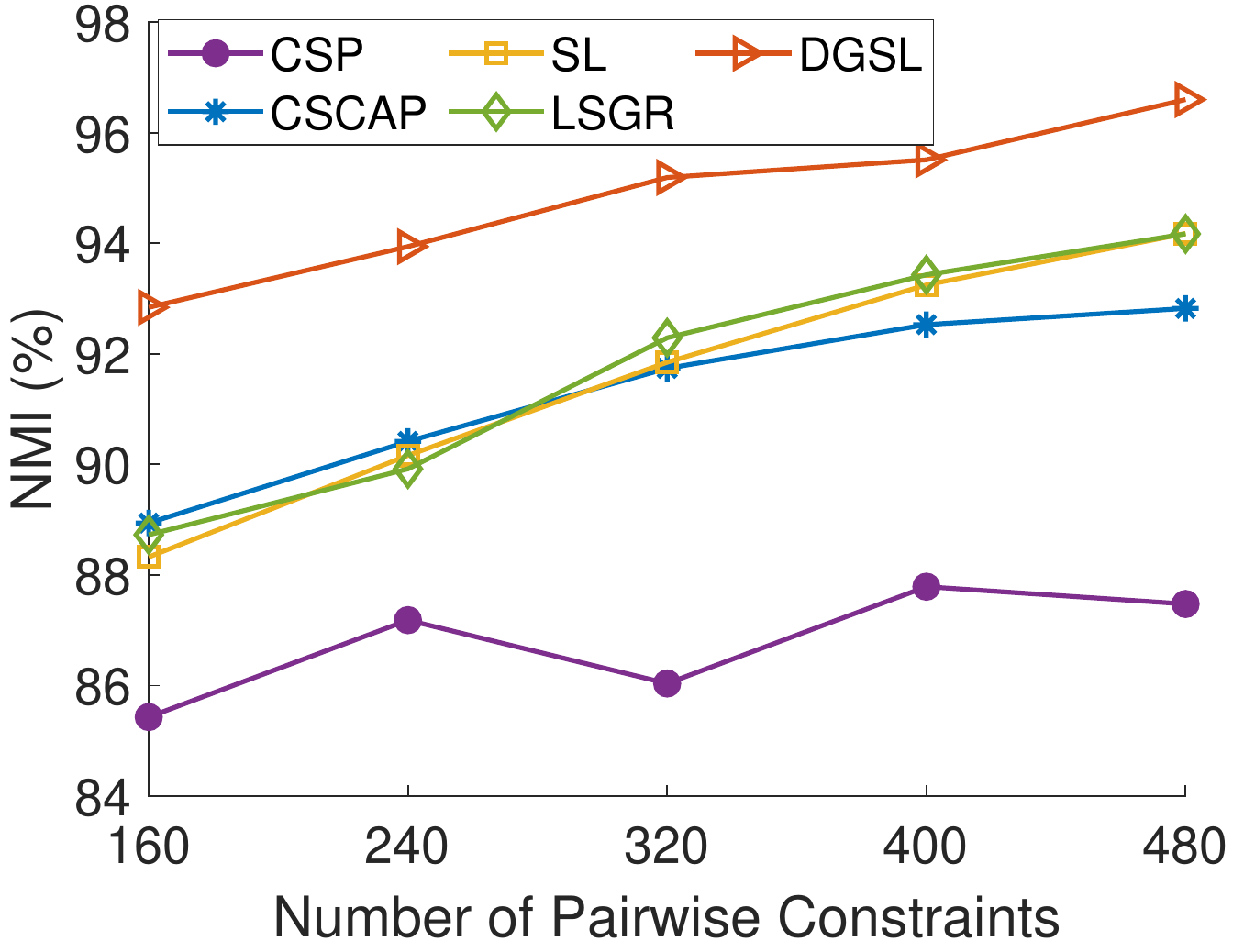}}%
\hfil
\subfloat[isolet1]{\includegraphics[width=1.8in]{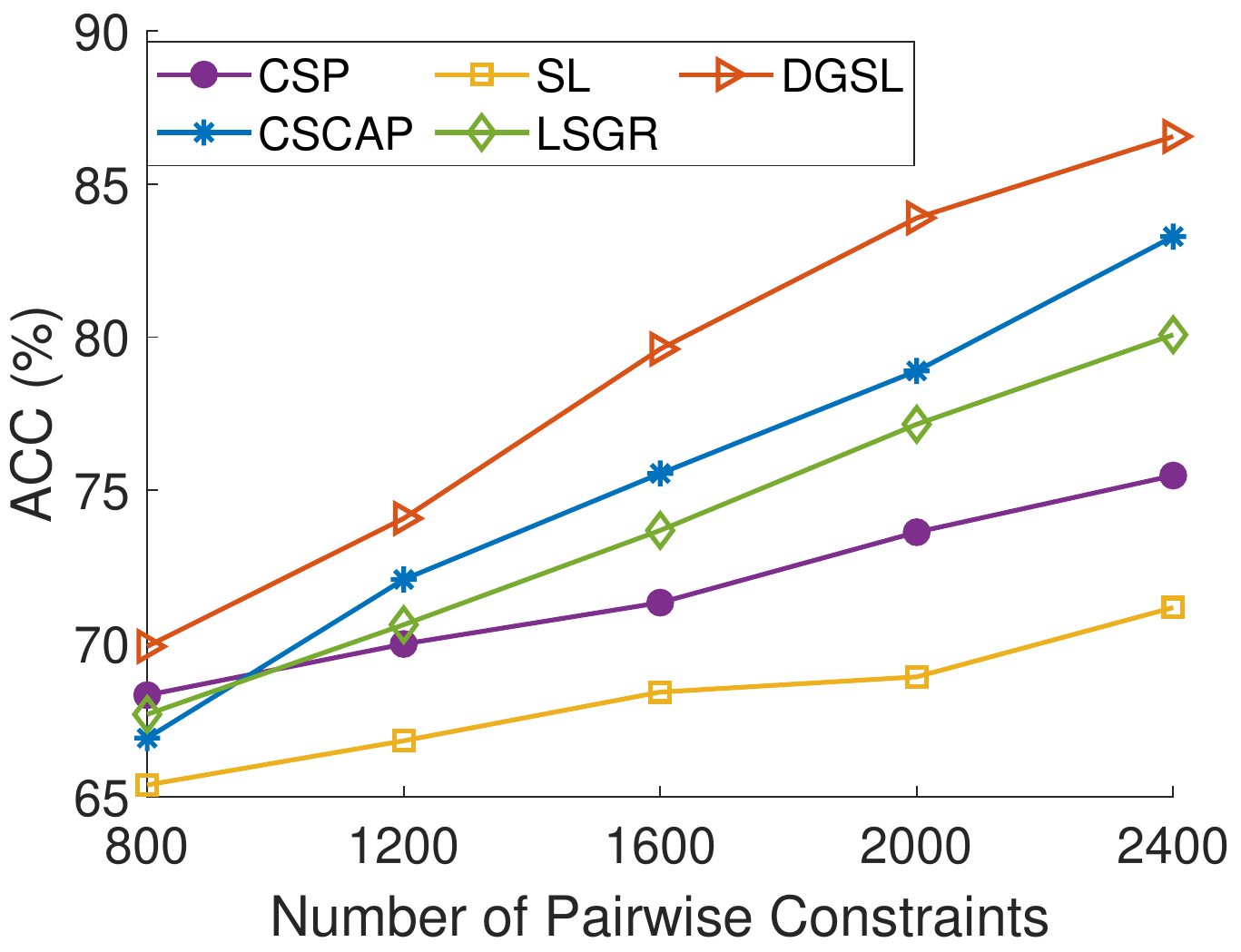}%
}
\hfil
\subfloat[isolet1]{\includegraphics[width=1.75in]{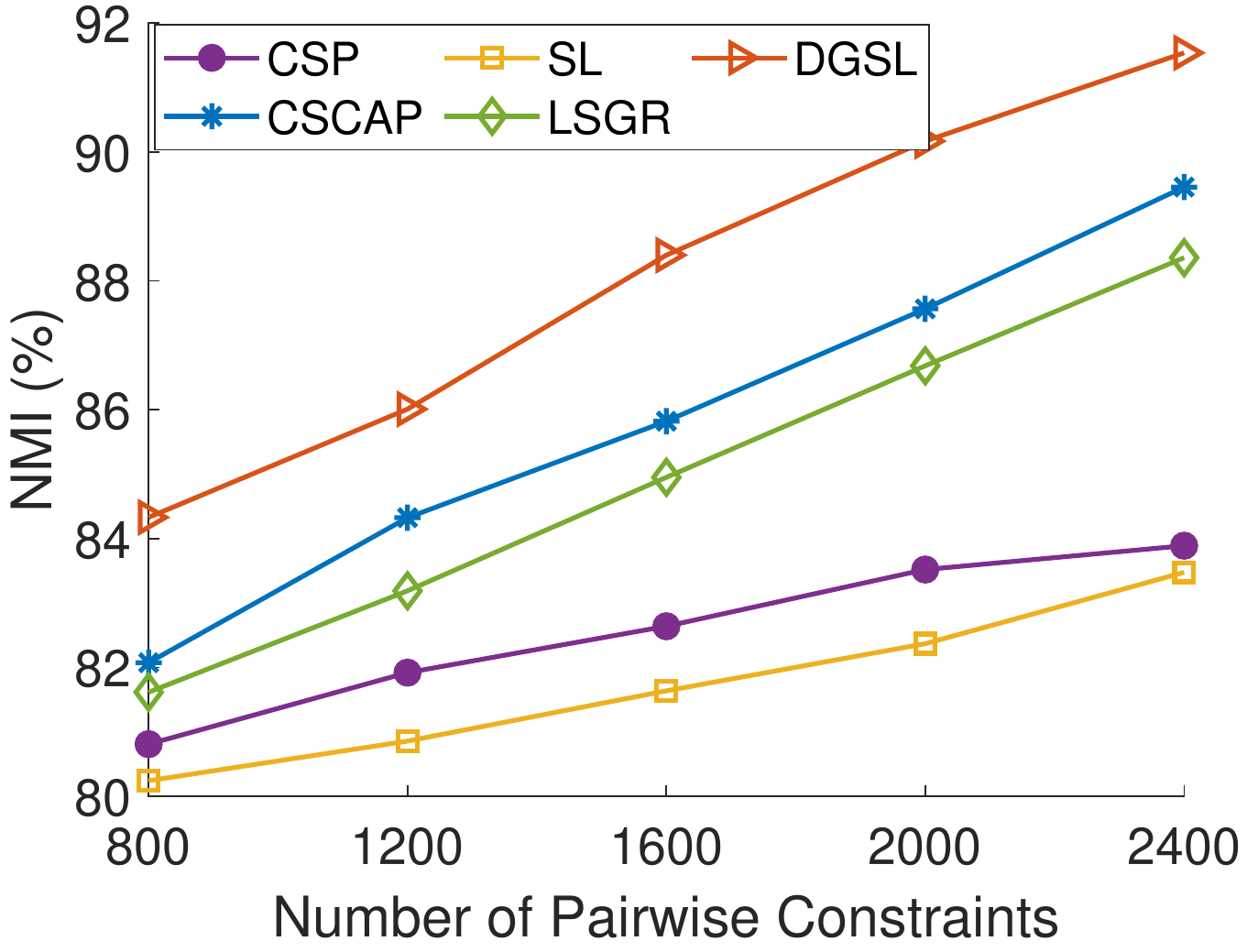}}%
\\
\subfloat[ORL]{\includegraphics[width=1.8in]{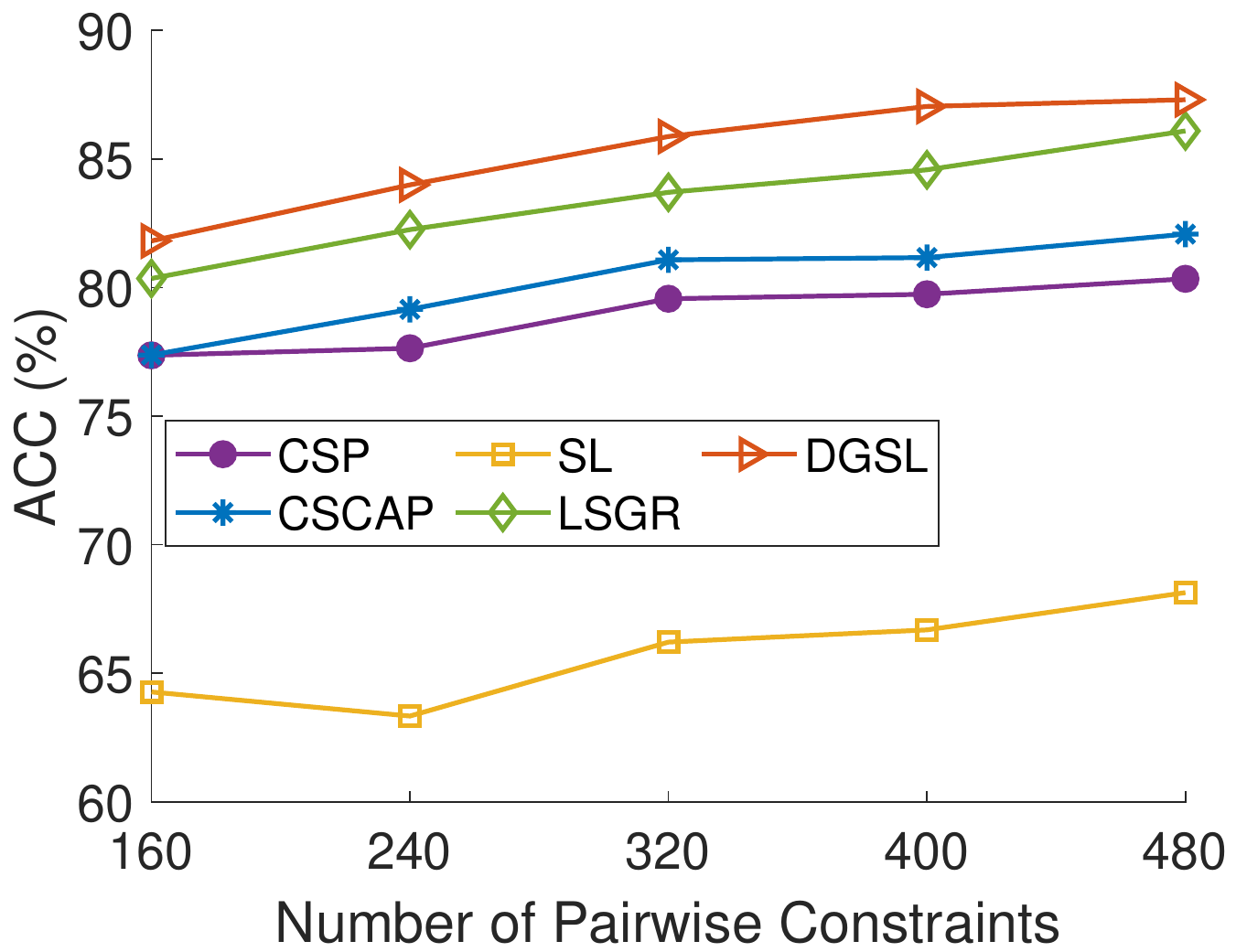}%
}
\hfil
\subfloat[ORL]{\includegraphics[width=1.75in]{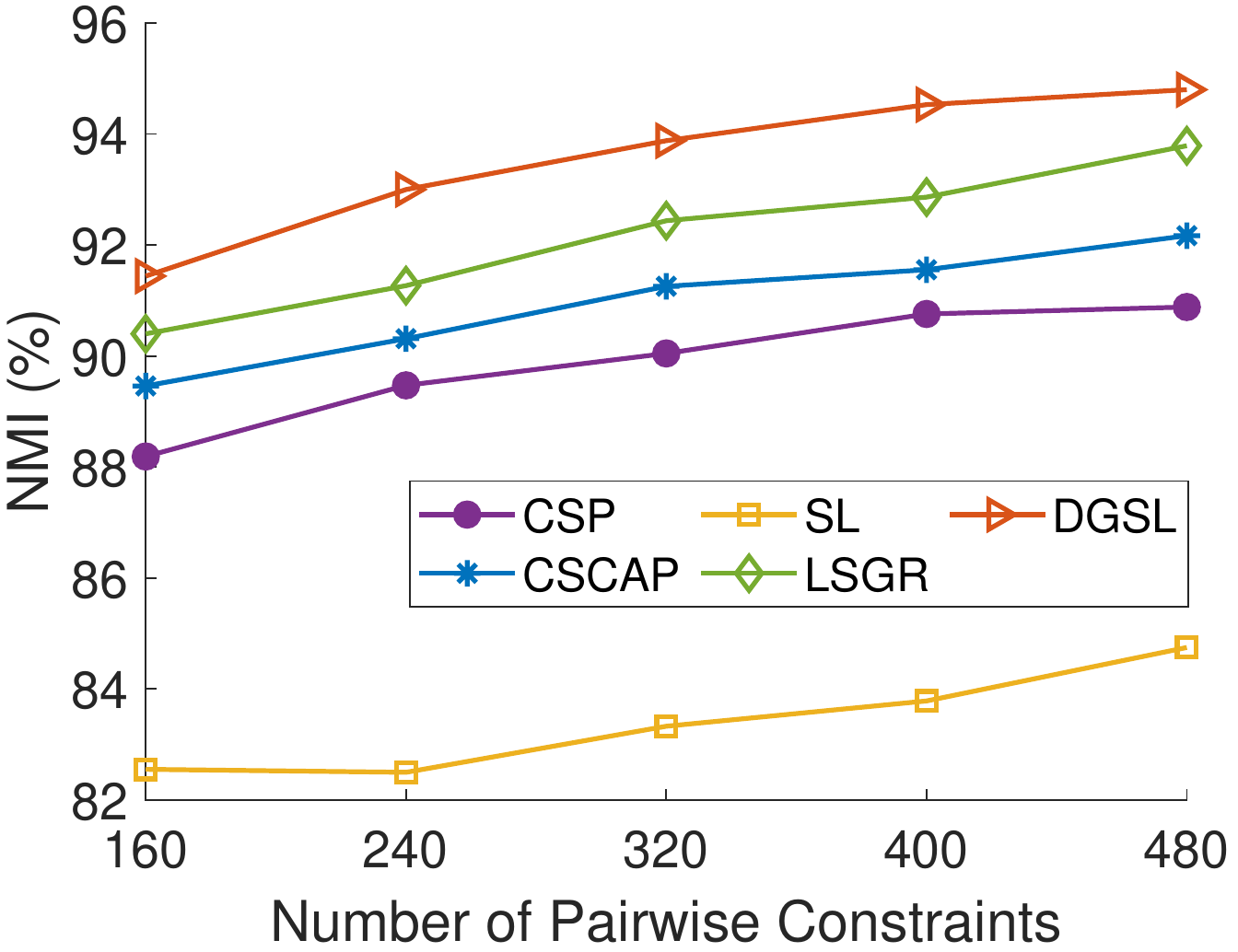}%
}
\hfil
\subfloat[Yale]{\includegraphics[width=1.75in]{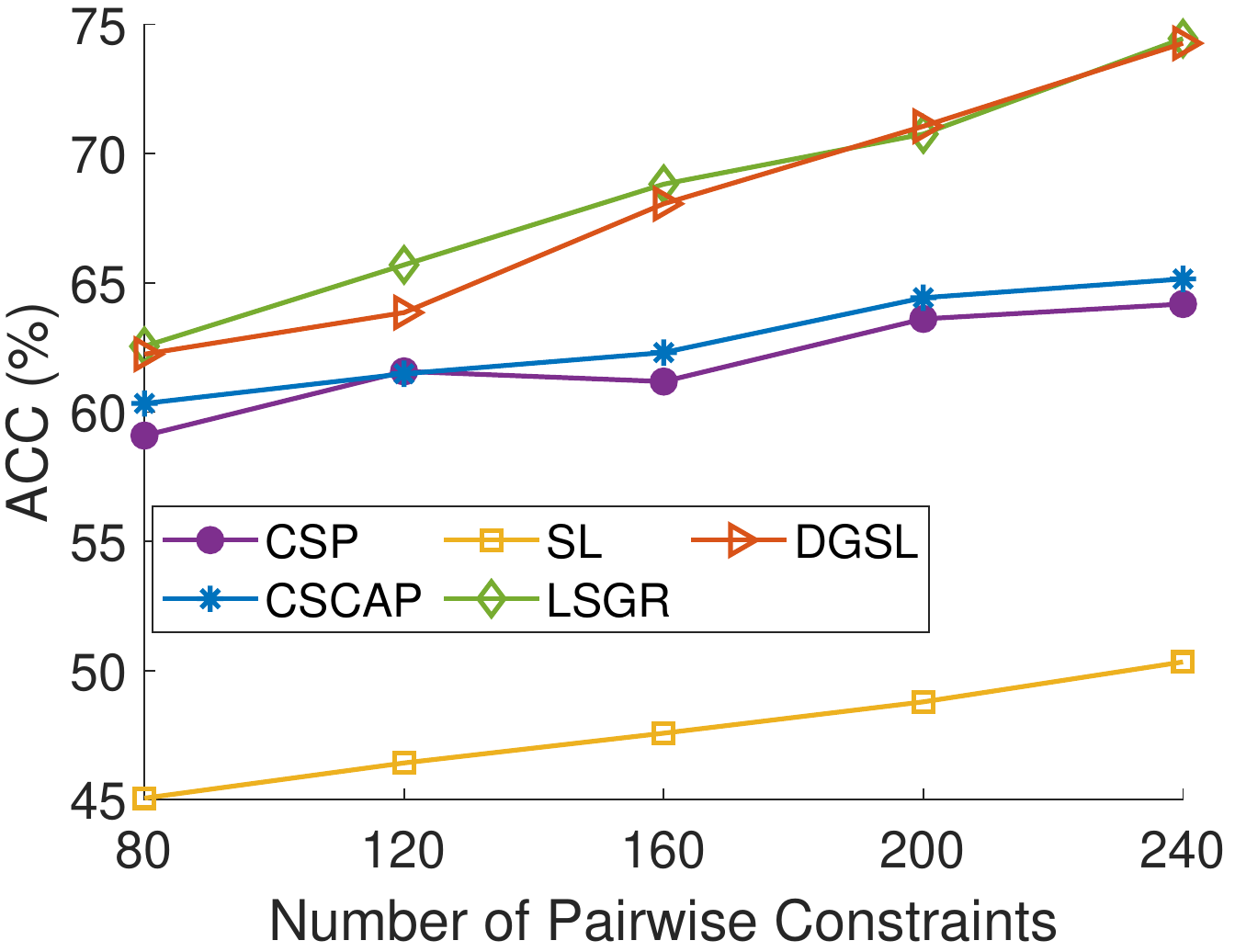}%
}
\hfil
\subfloat[Yale]{\includegraphics[width=1.75in]{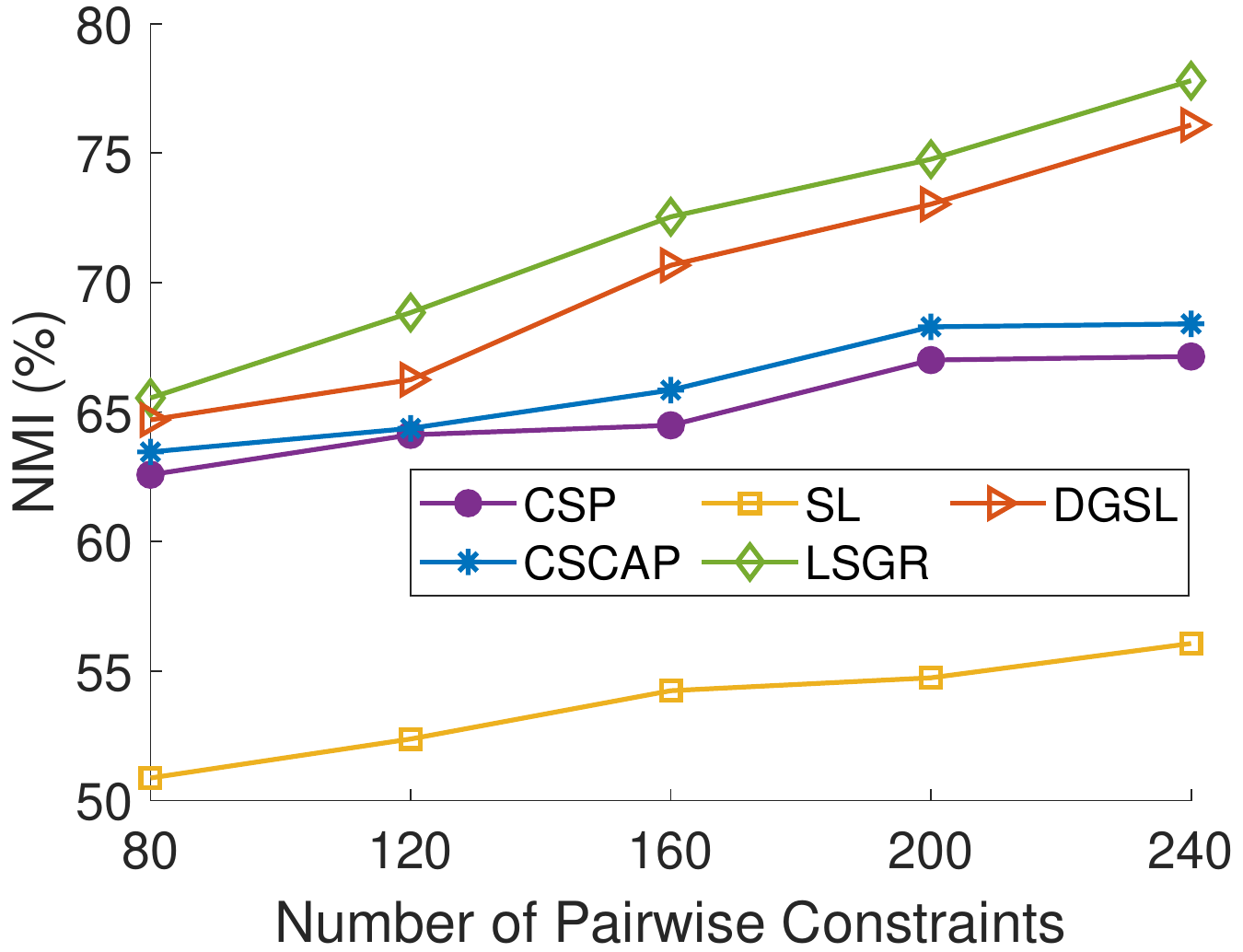}%
}
\\
\subfloat[Extended Yale B]{\includegraphics[width=1.8in]{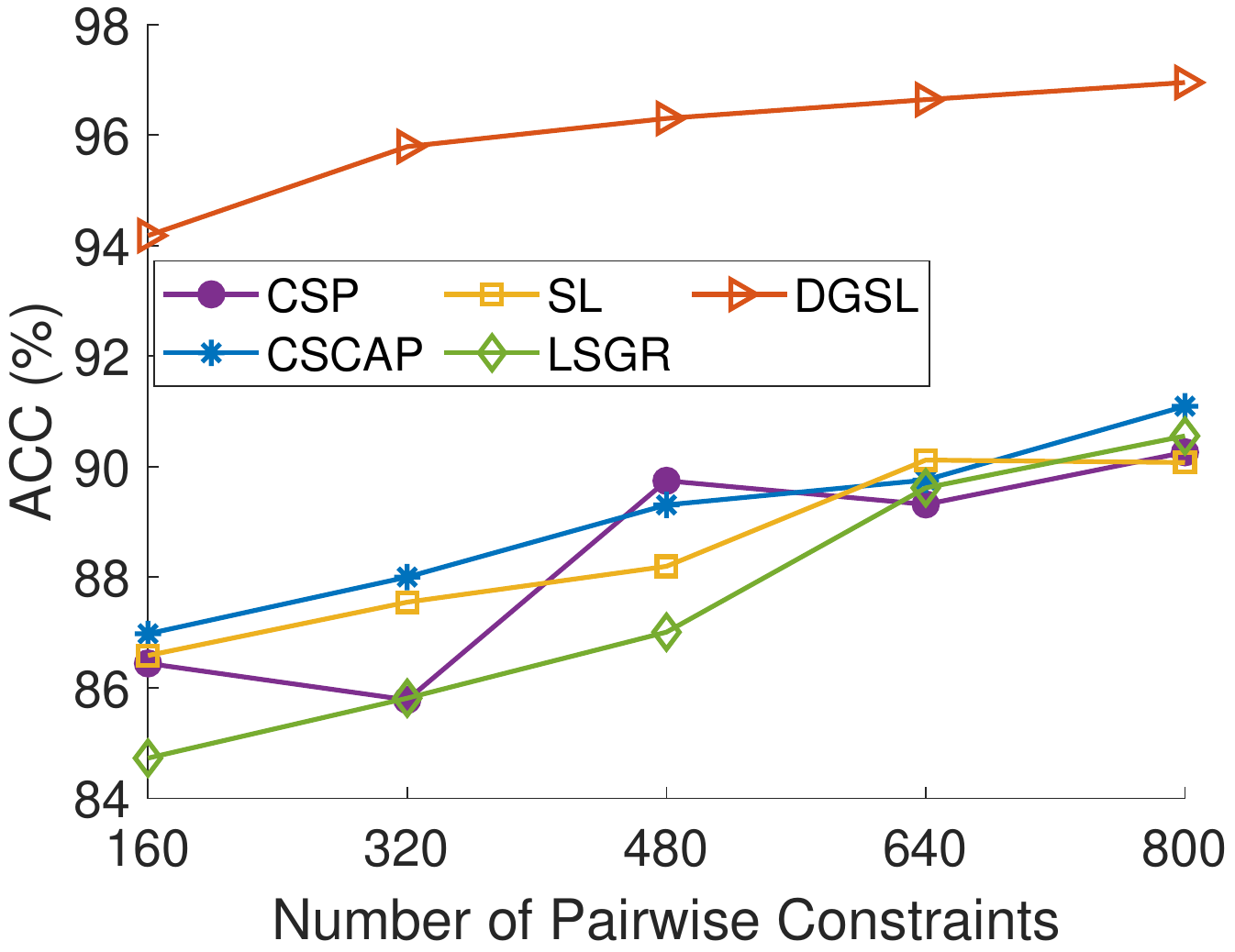}%
}
\hfil
\subfloat[Extended Yale B]{\includegraphics[width=1.75in]{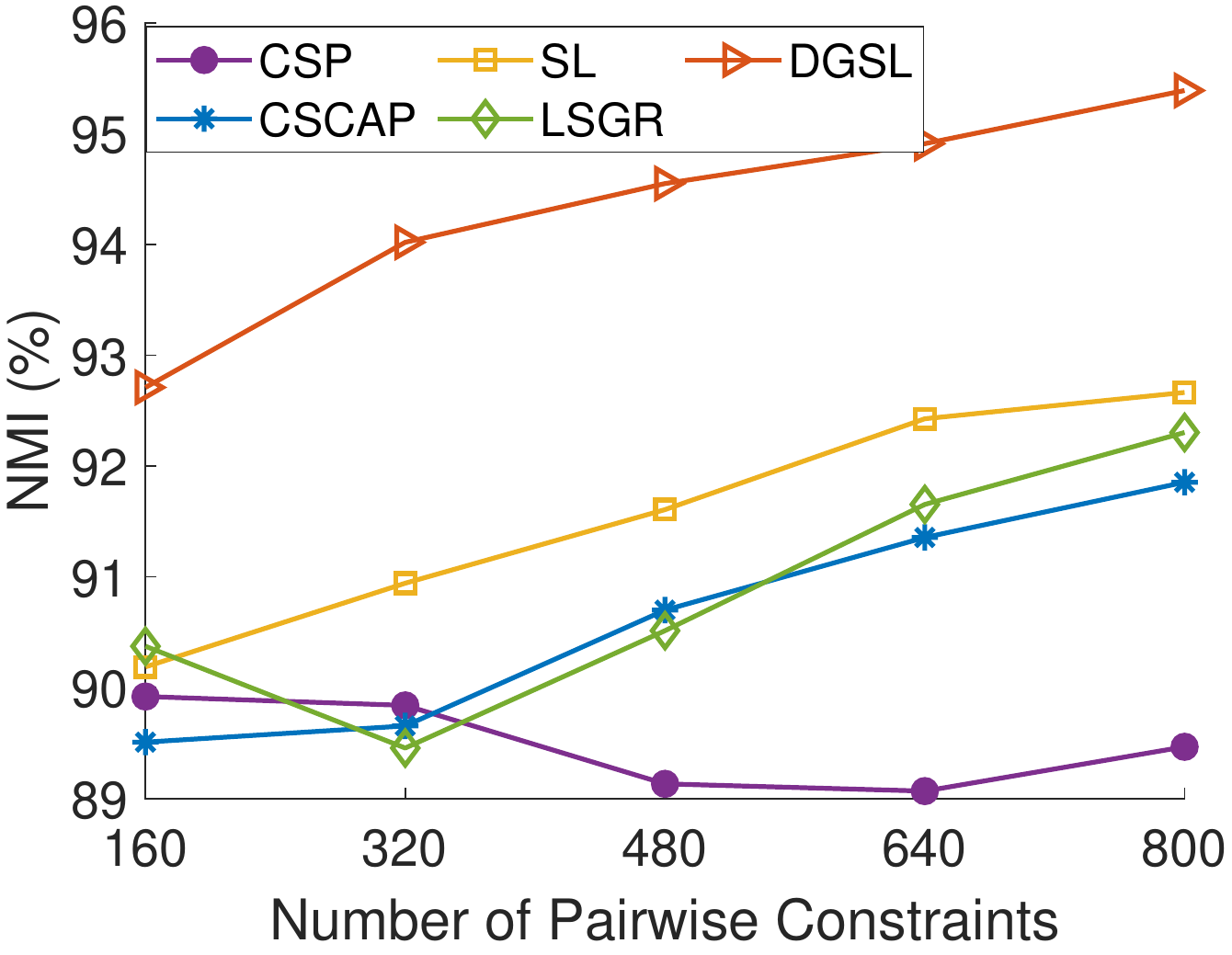}%
}
\hfil
\subfloat[COIL20]{\includegraphics[width=1.75in]{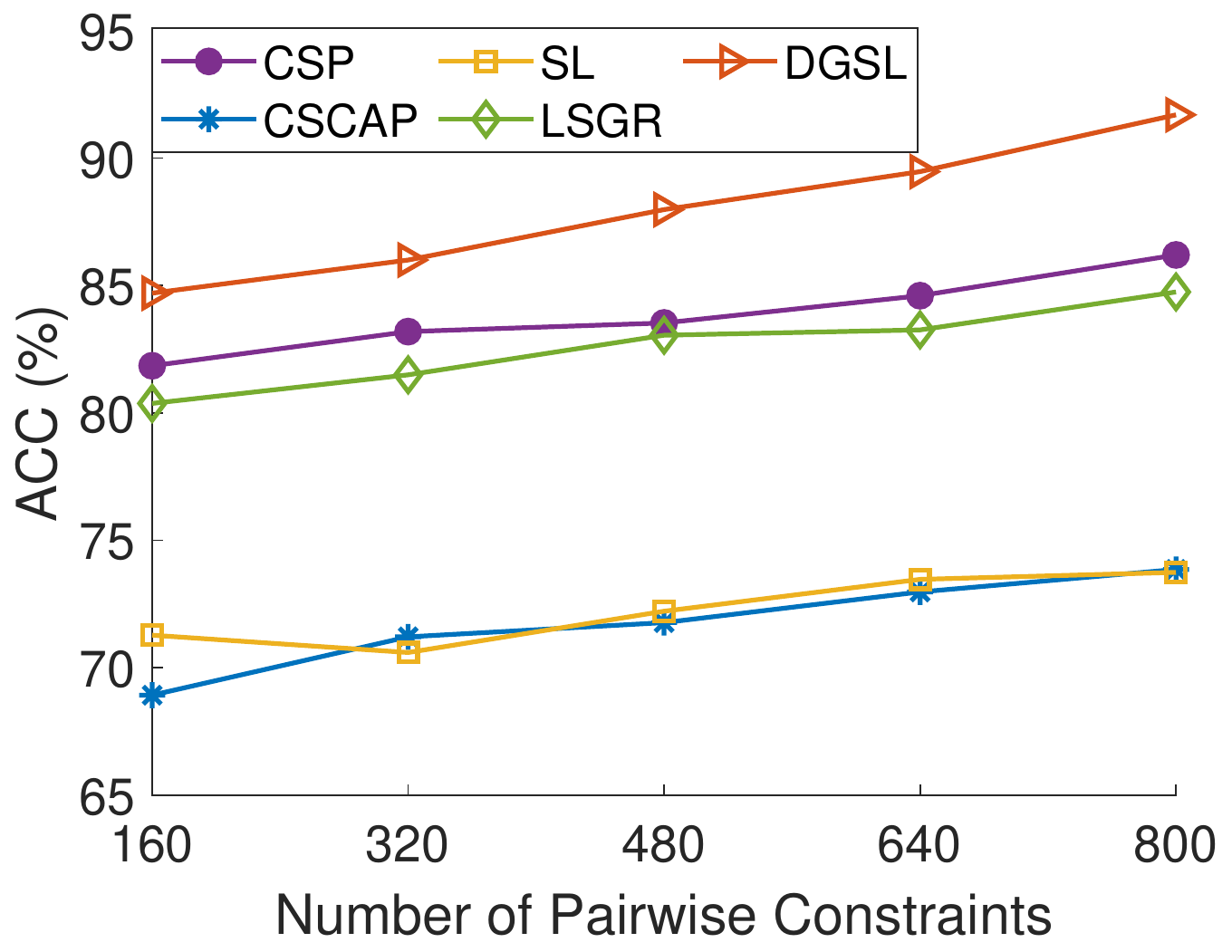}%
}
\hfil
\subfloat[COIL20]{\includegraphics[width=1.75in]{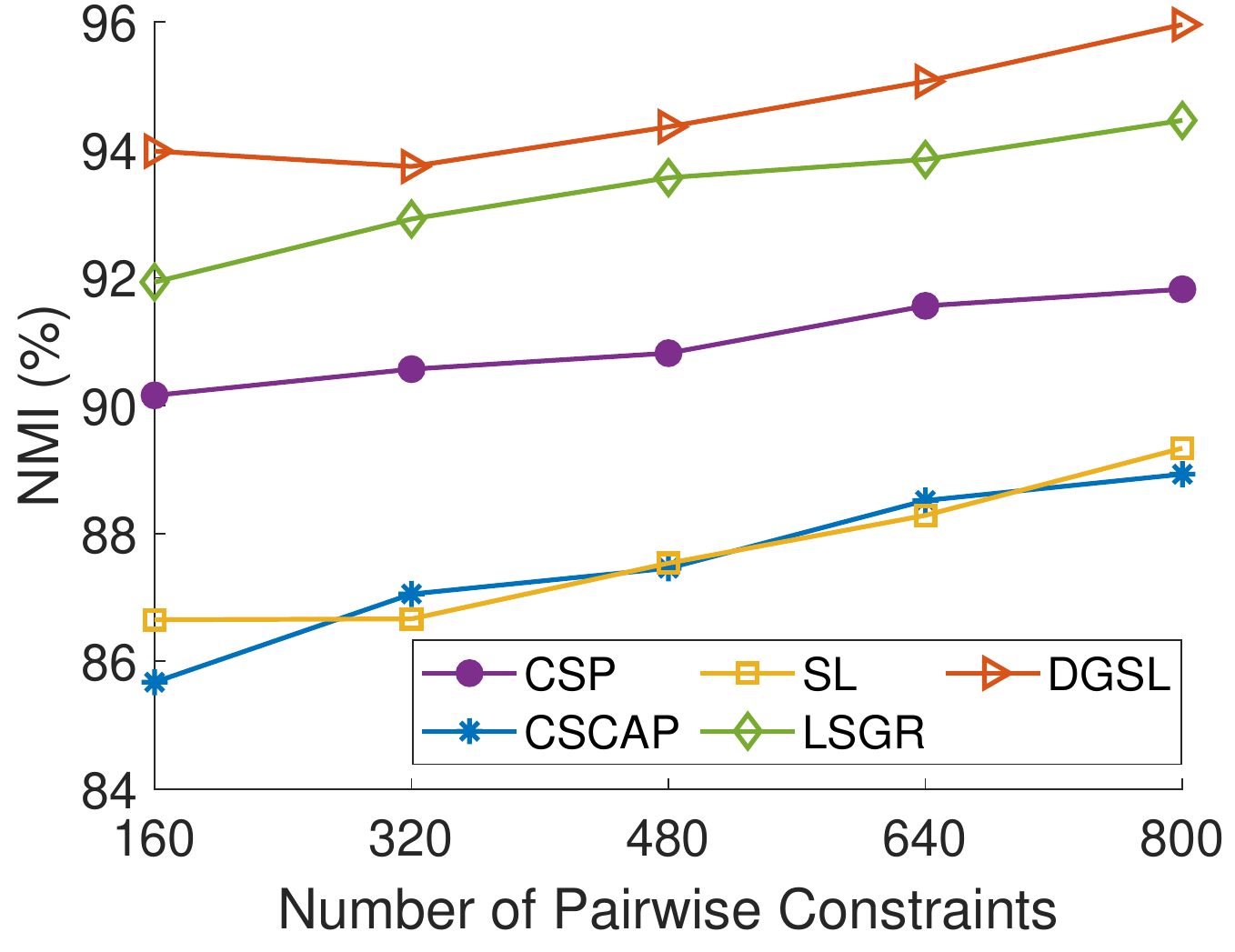}%
}
\\
\subfloat[MNIST]{\includegraphics[width=1.8in]{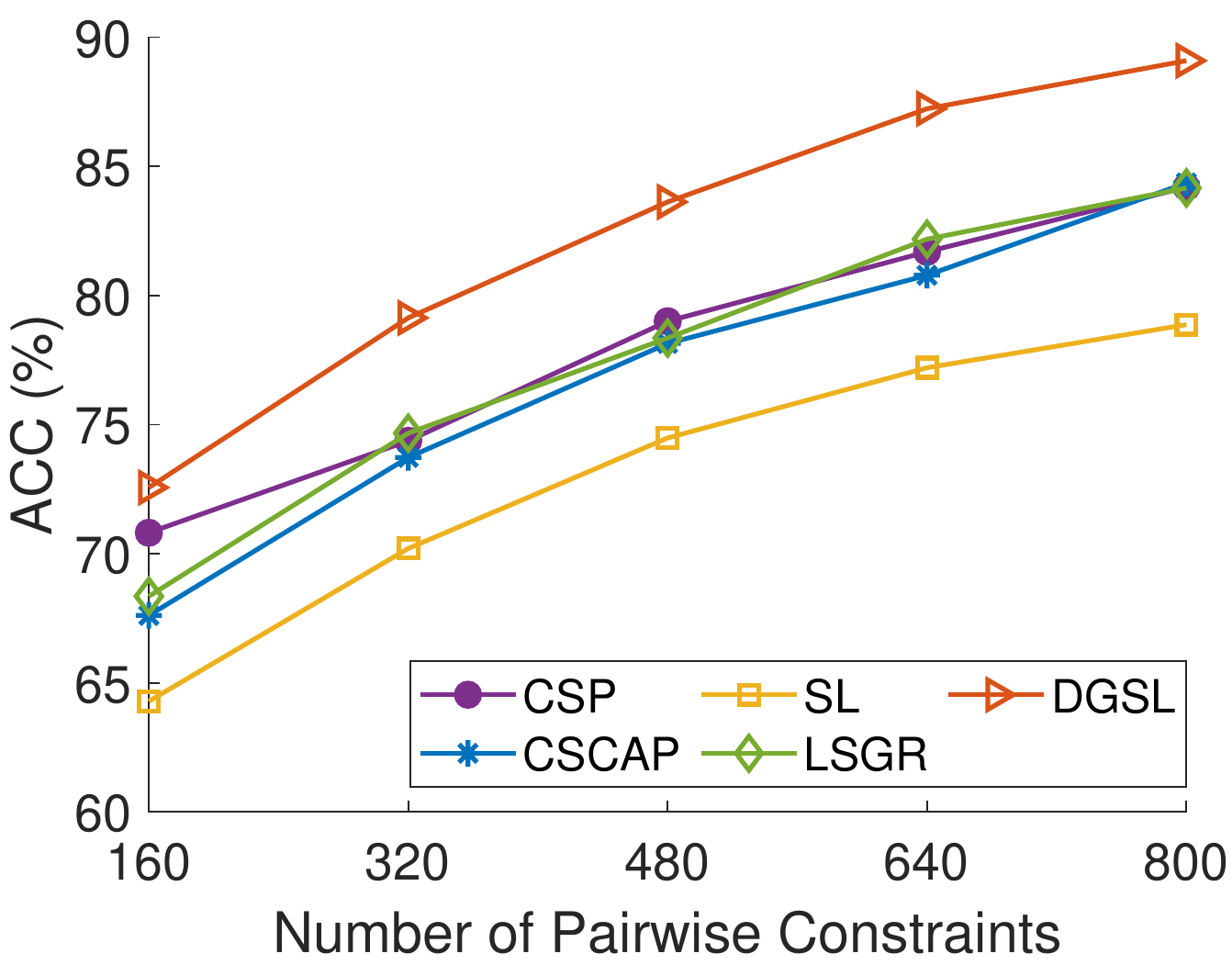}%
}
\hfil
\subfloat[MNIST]{\includegraphics[width=1.75in]{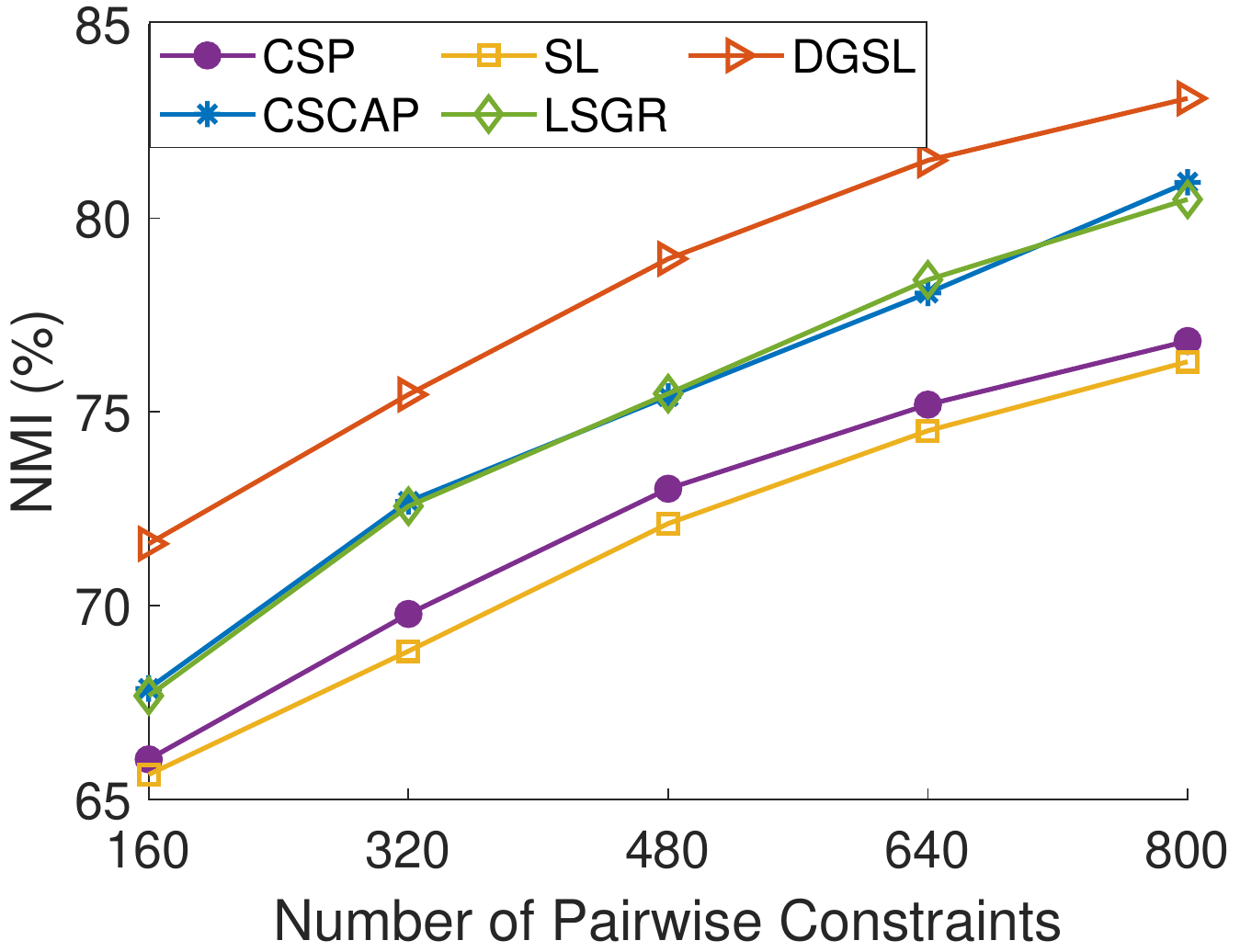}%
}
\hfil
\subfloat[USPS]{\includegraphics[width=1.75in]{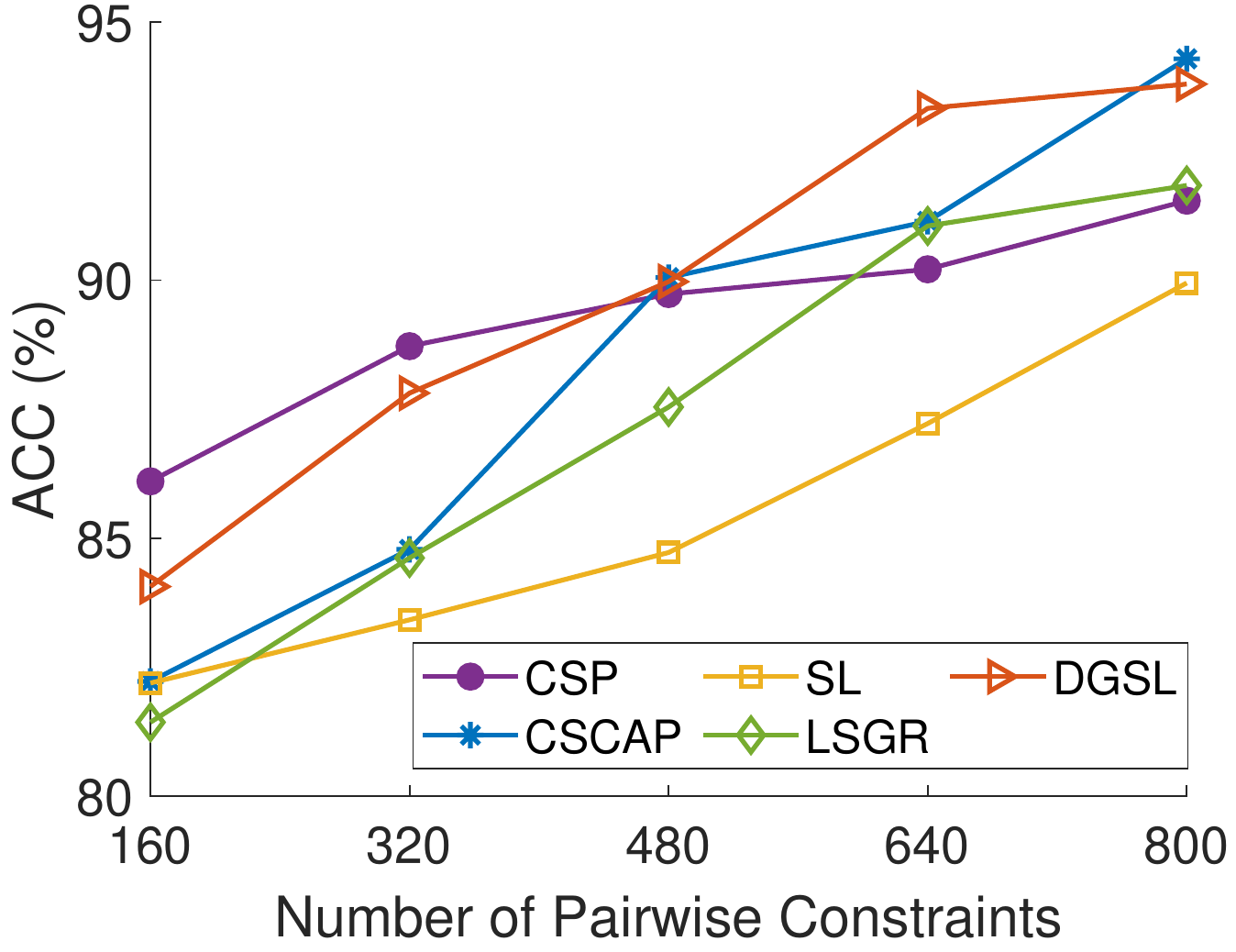}%
}
\hfil
\subfloat[USPS]{\includegraphics[width=1.75in]{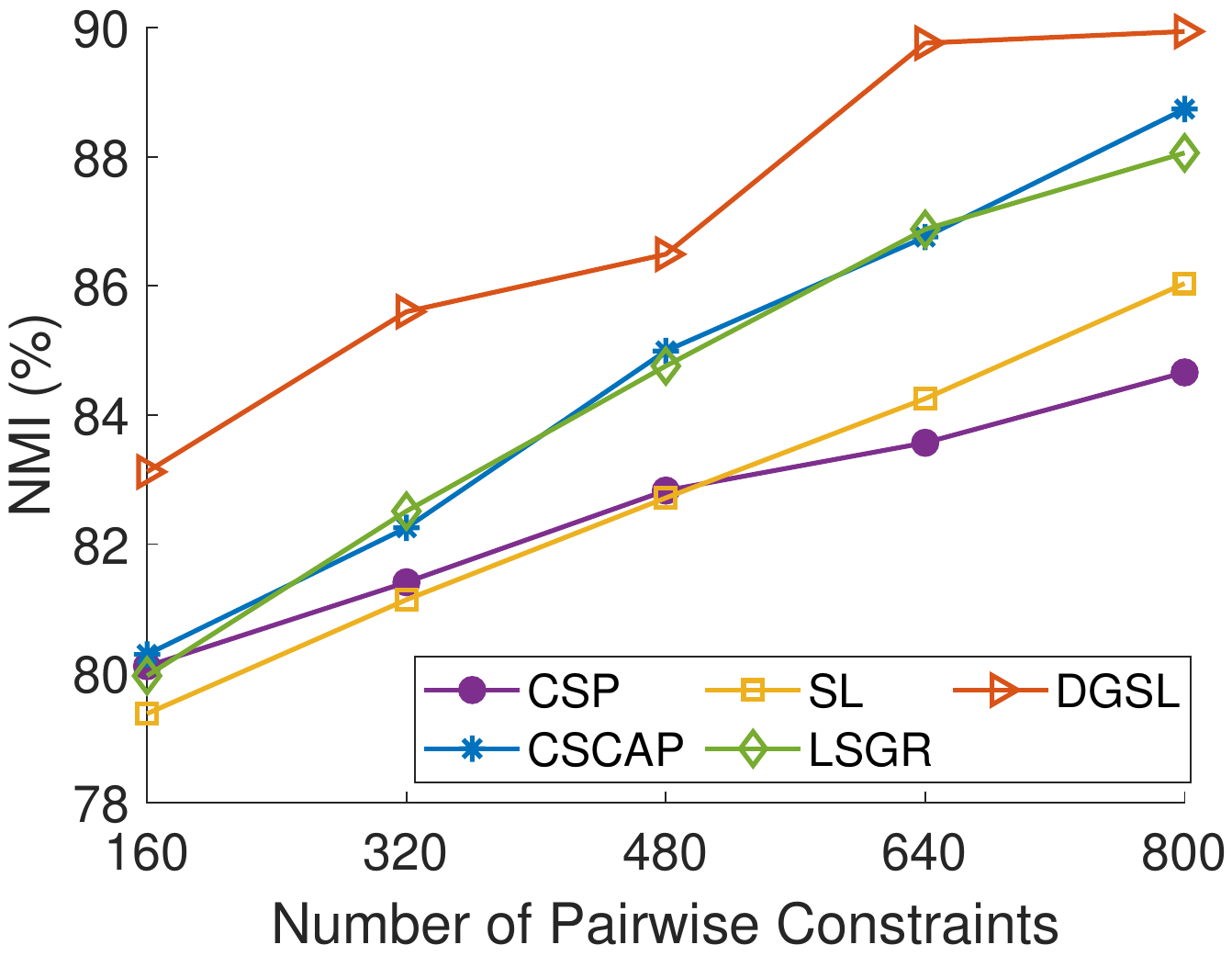}%
}
\caption{Clustering performance on different datasets with different numbers of pairwise constraints.}
\label{fig:mode 2}
\end{figure*}

\subsection{Comparisons with Methods That Use Pairwise Constraints as Supervisory Information}
\label{section:Comparisons with Methods Using Pairwise Constraints}
We compare our proposed DGSL with four state-of-the-art semi-supervised graph-based clustering methods which use pairwise constraints as supervisory information, i.e., SL \cite{kamvar2003spectral}, LSGR \cite{yang2014unified}, 
CSCAP \cite{lu2008constrained},
CSP \cite{wang2014constrained}. We use two different settings to generate the pairwise constraints. In the first setting, for any class $i$, we randomly select a subset of data points, denoted as  $S_i$, to construct the supervisory information. Then the set of must-link constraints $\mathcal{M}$ is defined as $\mathcal{M}=\{(\mathbf{x}_i,\mathbf{x}_j)|\mathbf{x}_i\in S_k,\mathbf{x}_j\in S_t,k=t\}$, and the set of cannot-link constraints $\mathcal{C}$ is defined as $\mathcal{C}=\{(\mathbf{x}_i,\mathbf{x}_j)|\mathbf{x}_i\in S_k,\mathbf{x}_j\in S_t,k\neq t\}$. In the second setting,  we randomly choose different numbers of data pairs with the same class label to generate $\mathcal{M}$ and data pairs with different class labels to generate $\mathcal{C}$.

For the first setting, we choose $f$ data points to construct $S_i$ for any class $i$, e.g., $f=2,3,4$ for ORL dataset. Each experiment is repeated 20  times independently with different supervisory information, and we report the average results, i.e., ACC ($\%$) and NMI ($\%$), as well as the standard deviations (STD$\%$). The clustering results are shown in Table \ref{tab:mode 1 ACC} and Table \ref{tab:mode 1 NMI}. The best results are in bold font. 

For the second setting, we set the number of cannot-link constraints as three times the number of must-link constraints. Each experiment is repeated 20  times with different supervisory information, and we report the average results in Fig. \ref{fig:mode 2}.

From Table \ref{tab:mode 1 ACC}, Table \ref{tab:mode 1 NMI} and Fig. \ref{fig:mode 2}, the following observations can be made: (1)  DGSL consistently achieves the highest ACC and NMI on most datasets, which verifies the effectiveness of our dynamic graph structure learning scheme; (2) The performance of our proposed DGSL improves rapidly with the increasing supervisory information on most datasets, demonstrating that DGSL can utilize the supervisory information effectively.

\subsection{Comparisons with Methods That Use Partial Labels as Supervisory Information}
We also compare DGSL with five state-of-the-art semi-supervised subspace clustering methods that use partial labels as supervisory information, i.e.,
NNLRS \cite{zhuang2012non}, NNLRR \cite{fang2015robust}, S$^3$R \cite{li2015learning}, S$^2$LRR \cite{li2015learning}   and  DCSSC \cite{wang2018unified}. 
We randomly select $f$ data points from any class $i$, denoted as $S_i$, as labeled samples to generate the supervisory information, e.g., $f=2,6,10$ for COIL20 dataset. The clustering performance of compared methods is cited from literature in \cite{wang2018unified}. The clustering performance is shown in Table \ref{tab:compare with methods using partial labels} and the best results are in bold font.   It can be seen that our proposed DGSL achieves competitive results on the three datasets. Specifically, DGSL achieves at least $5.7\%$ improvement over ACC of the compared methods for the case $f=2$ in the COIL20 dataset. Thus, for datasets COIL20, Yale, and Extended Yale B, DGSL can achieve competitive clustering performance compared with state-of-the-art semi-supervised subspace clustering methods. Moreover, DGSL adopts weaker and more flexible supervisory information, i.e., pairwise constraints, than the compared methods that leverage partial labels.

\begin{figure*}[!t]
\centering
\subfloat[ORL]{\includegraphics[width=1.8in]{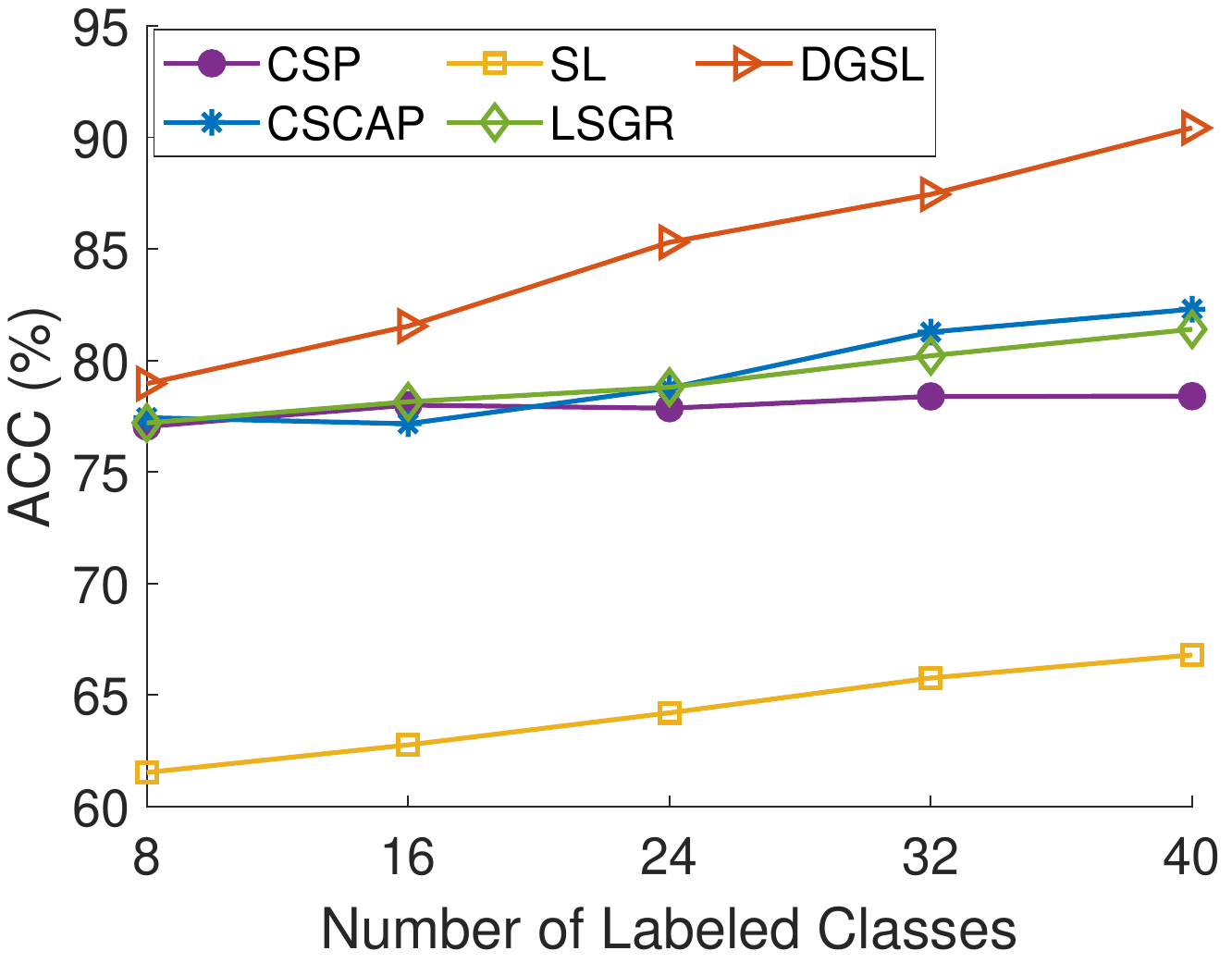}%
}
\hfil
\subfloat[ORL]{\includegraphics[width=1.75in]{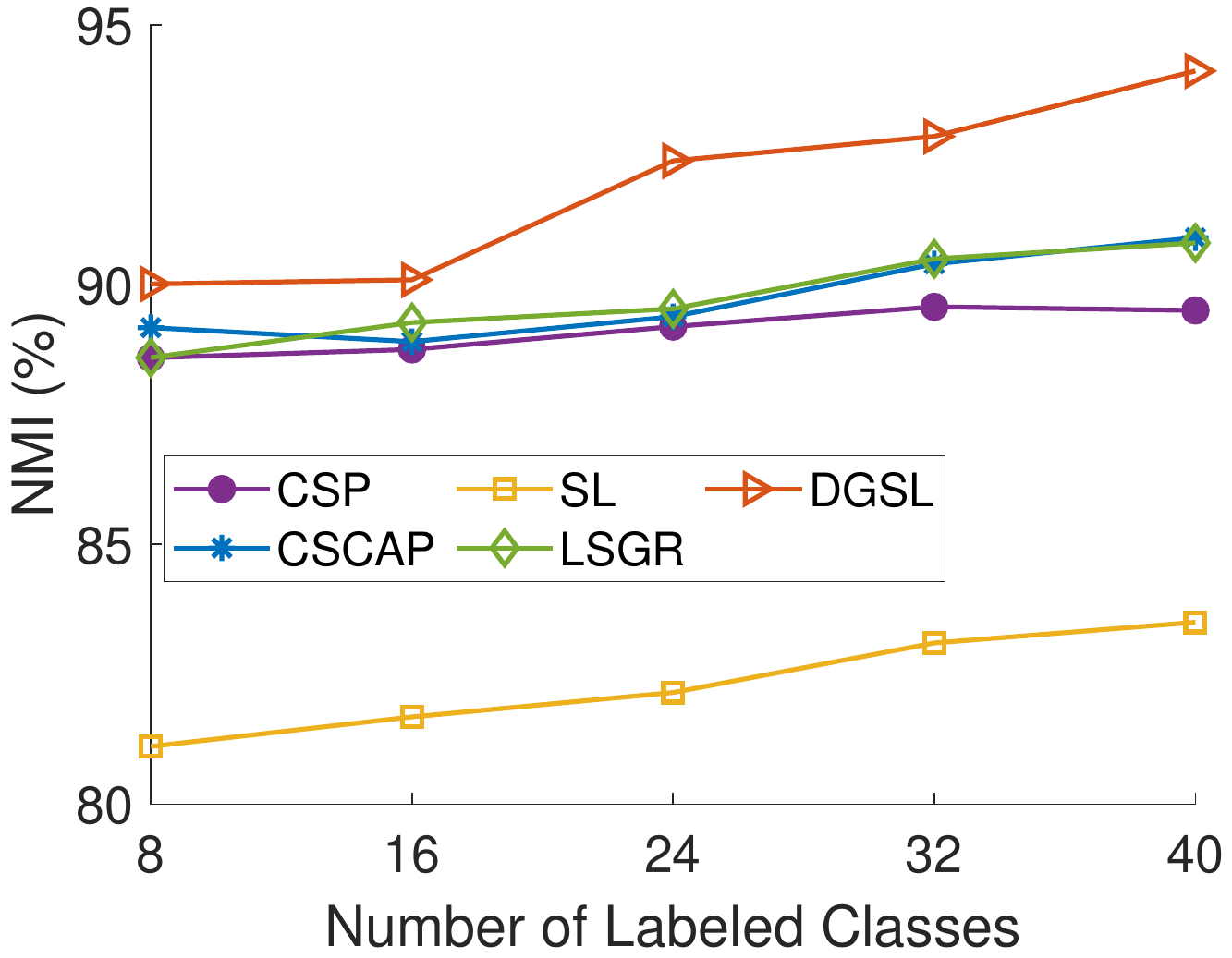}}%
\hfil
\subfloat[Yale]{\includegraphics[width=1.8in]{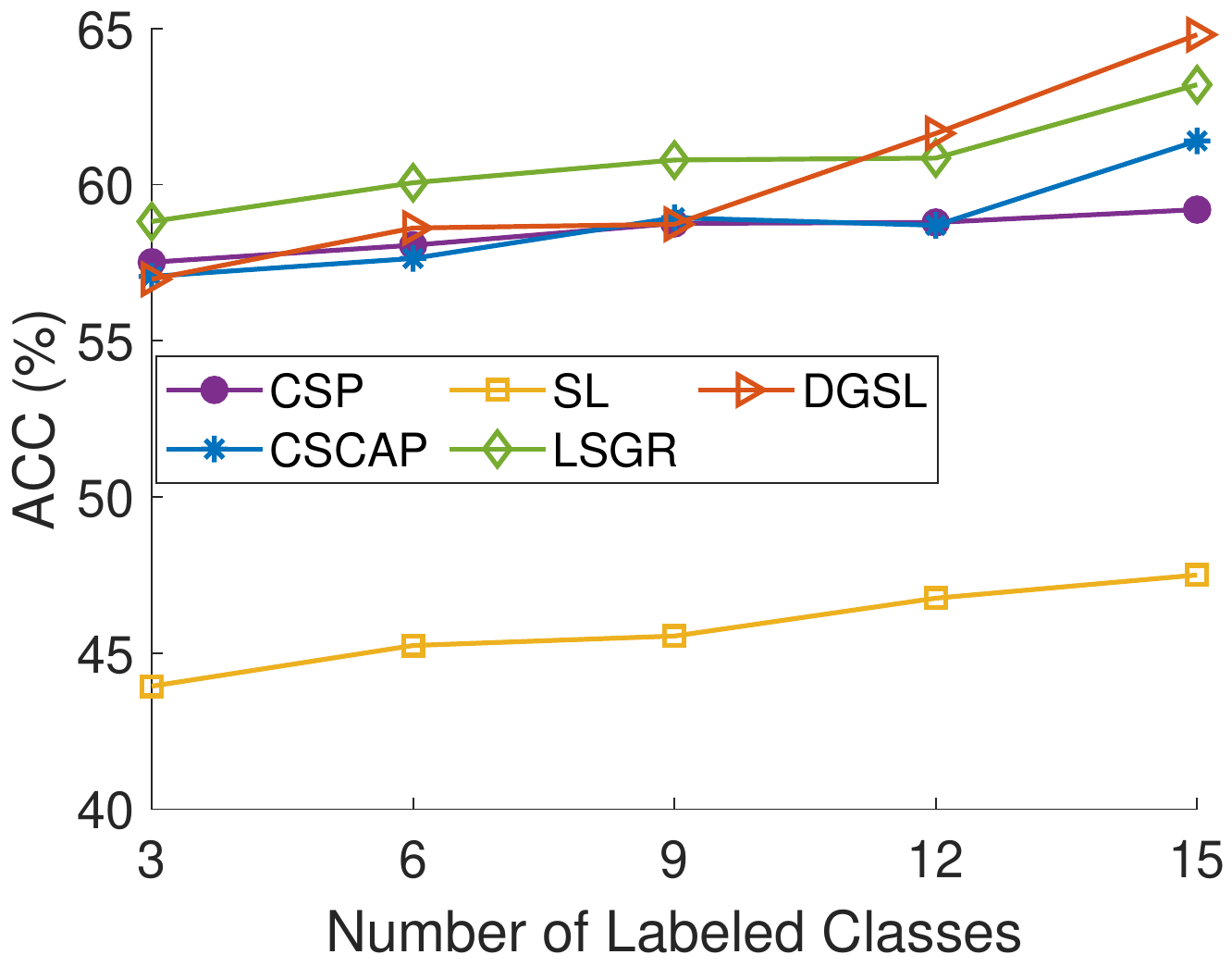}%
}
\hfil
\subfloat[Yale]{\includegraphics[width=1.75in]{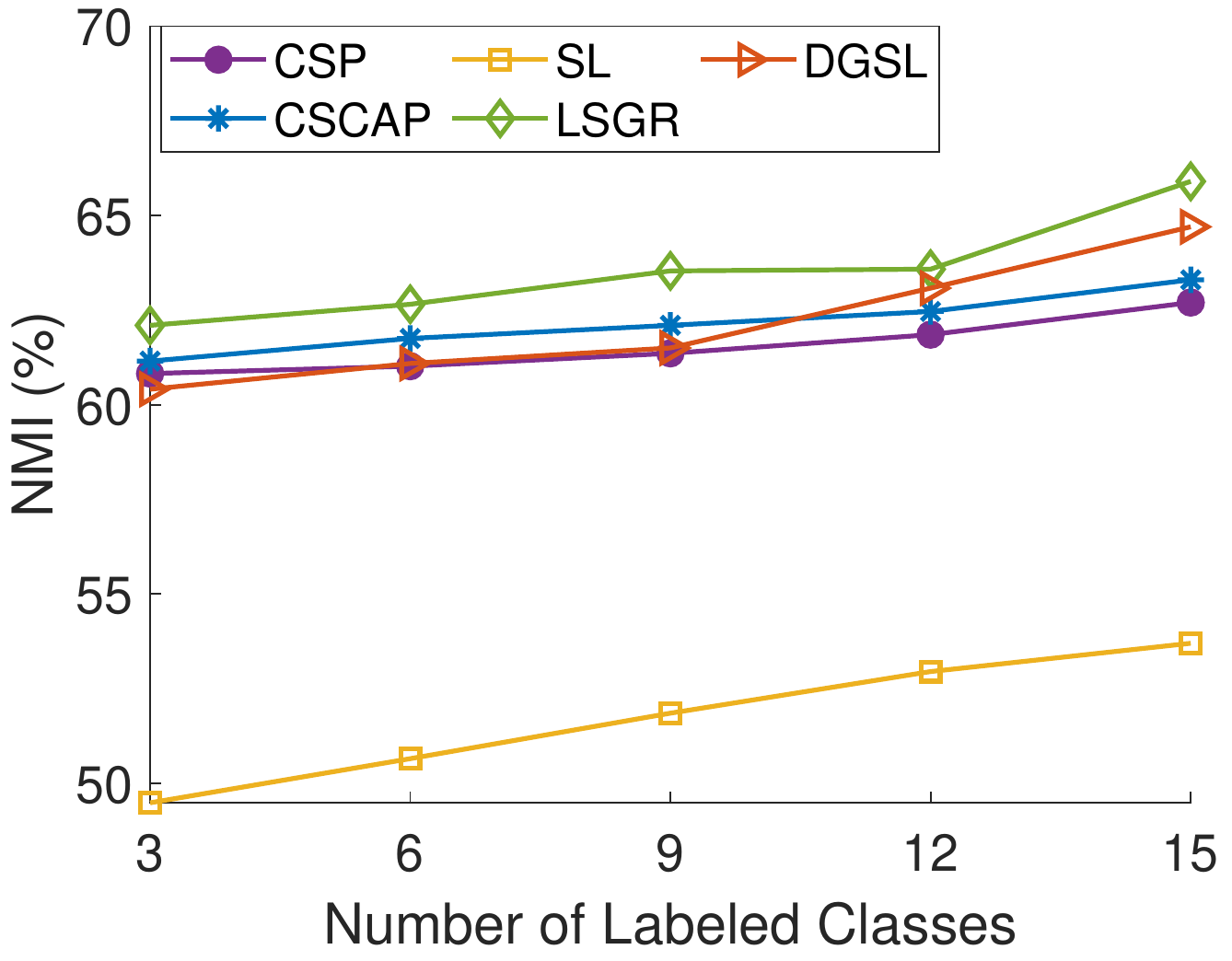}}%
\\
\subfloat[MNIST]{\includegraphics[width=1.8in]{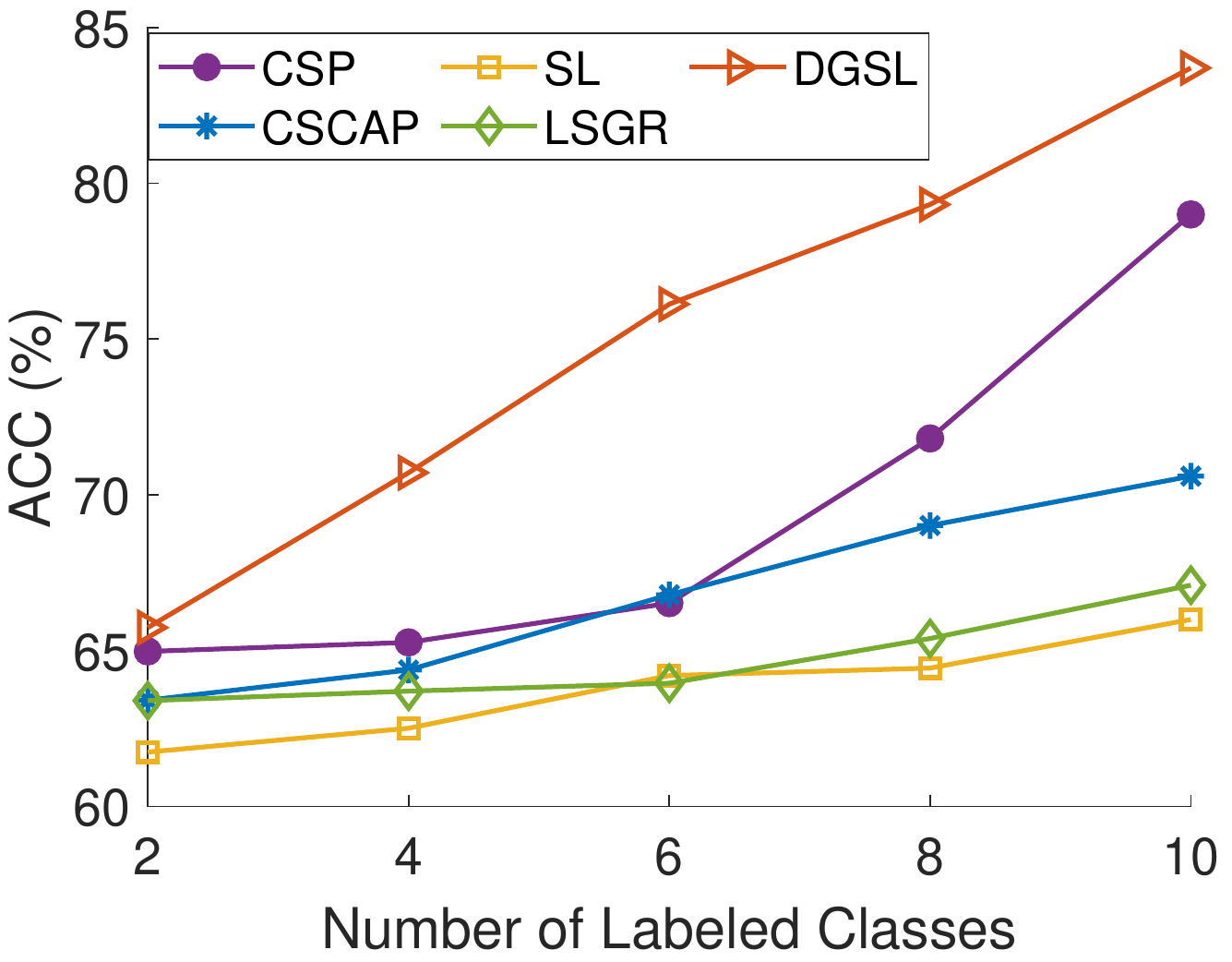}%
}
\hfil
\subfloat[MNIST]{\includegraphics[width=1.75in]{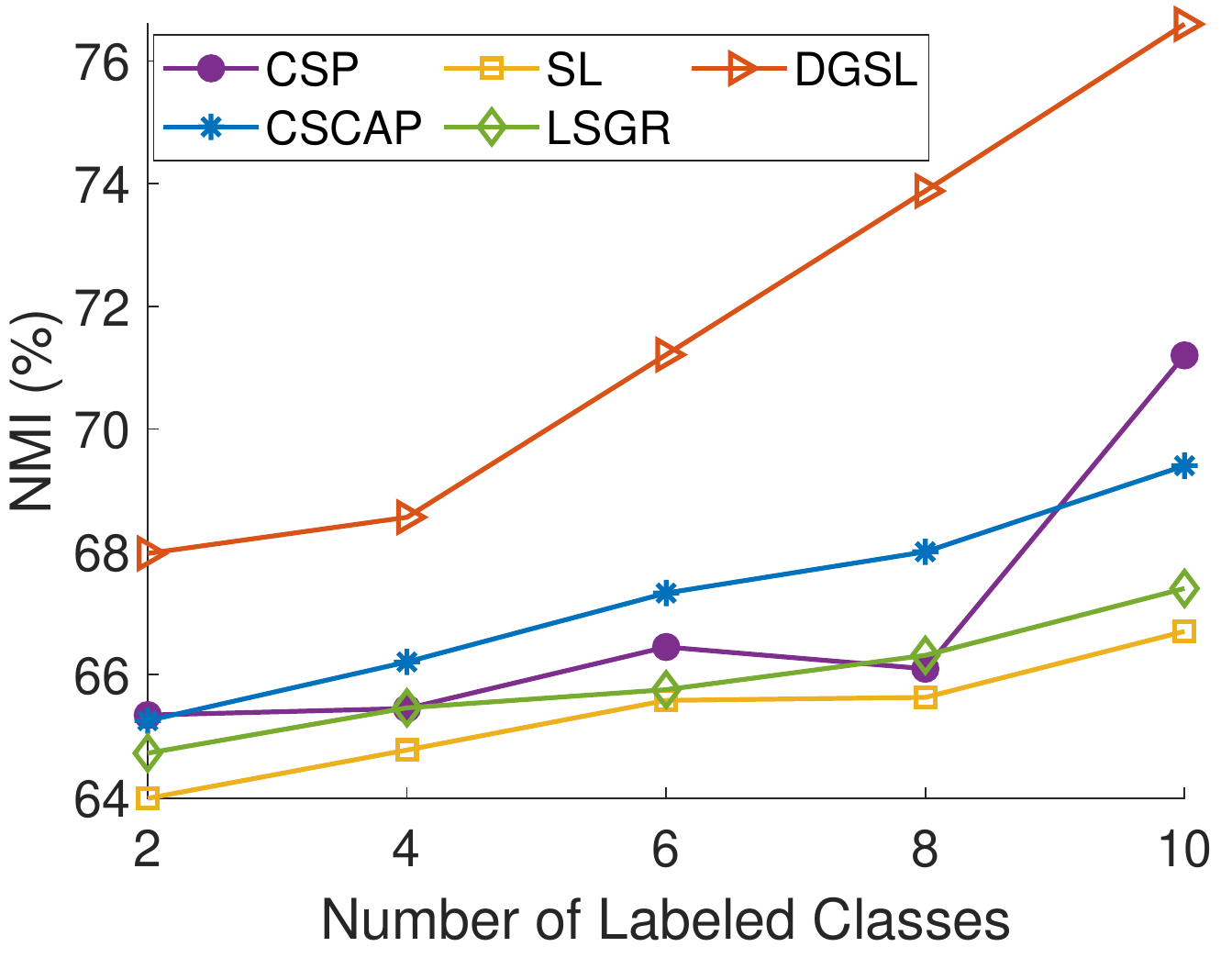}%
}
\hfil
\subfloat[isolet1]{\includegraphics[width=1.75in]{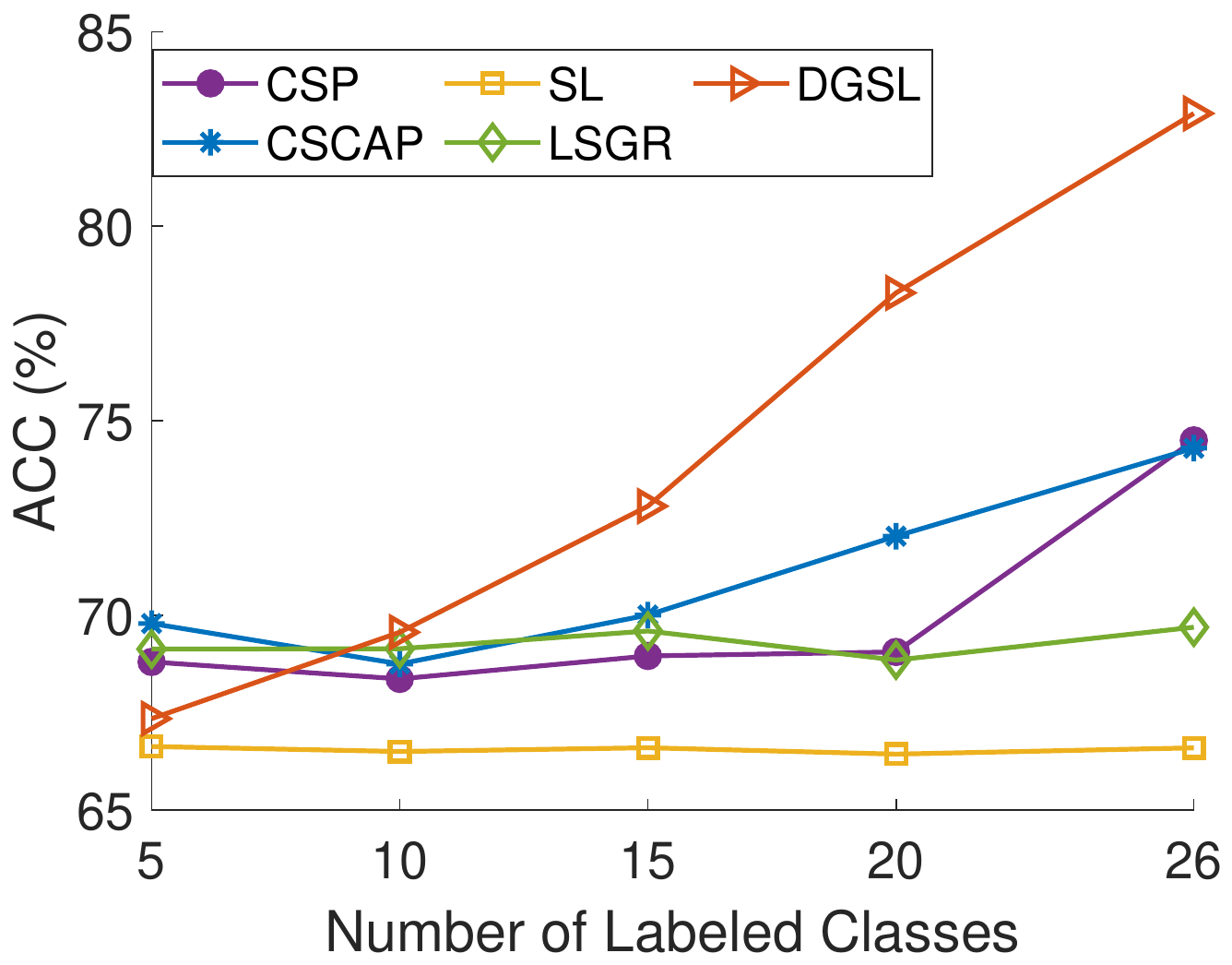}%
}
\hfil
\subfloat[isolet1]{\includegraphics[width=1.75in]{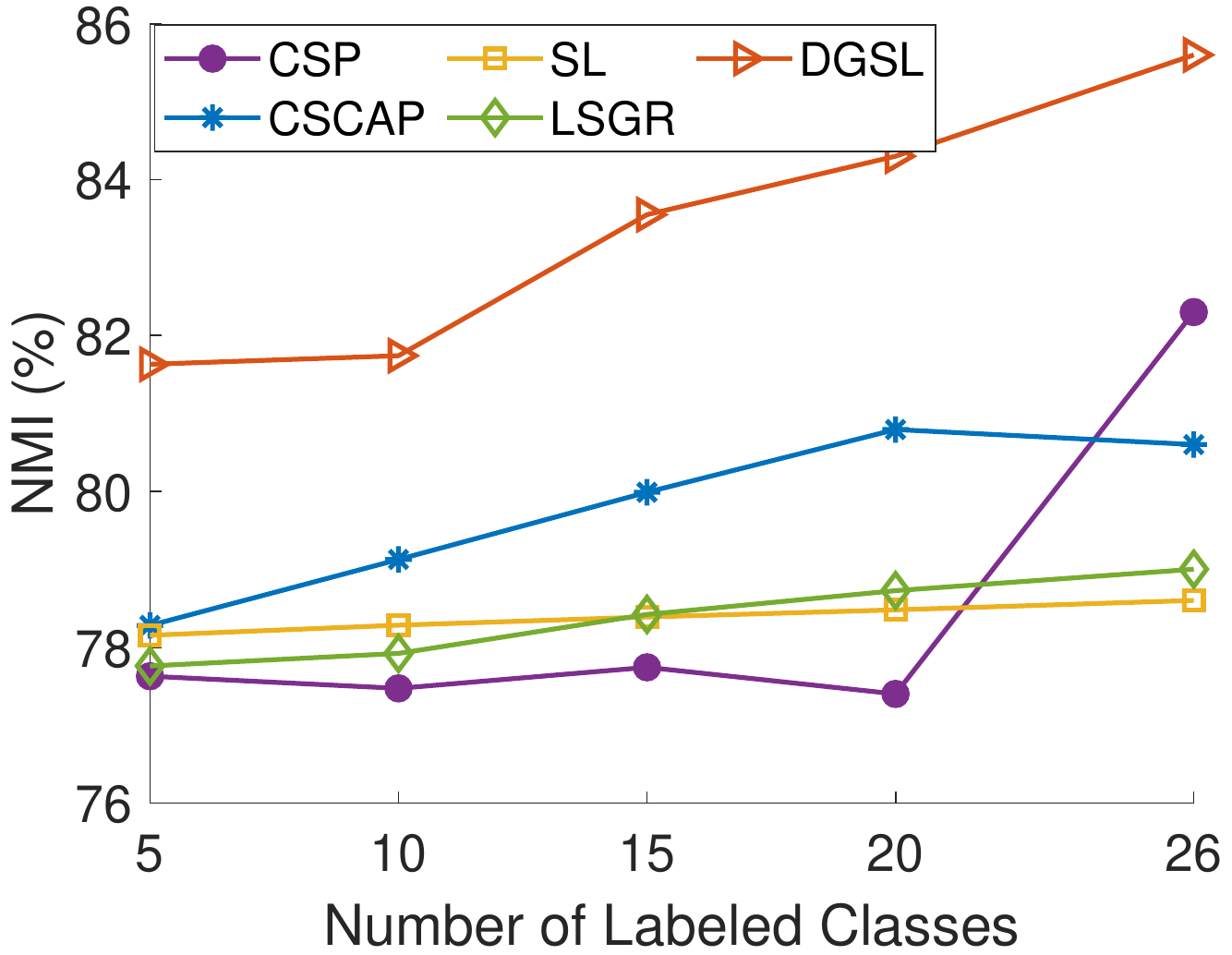}%
}
\caption{Clustering performance on different datasets with different numbers of labeled classes.}
\label{fig:incomplete}
\end{figure*}

\begin{figure*}[!t]
\centering
\subfloat[ACC versus  $\alpha_1$]{\includegraphics[width=1.75in]{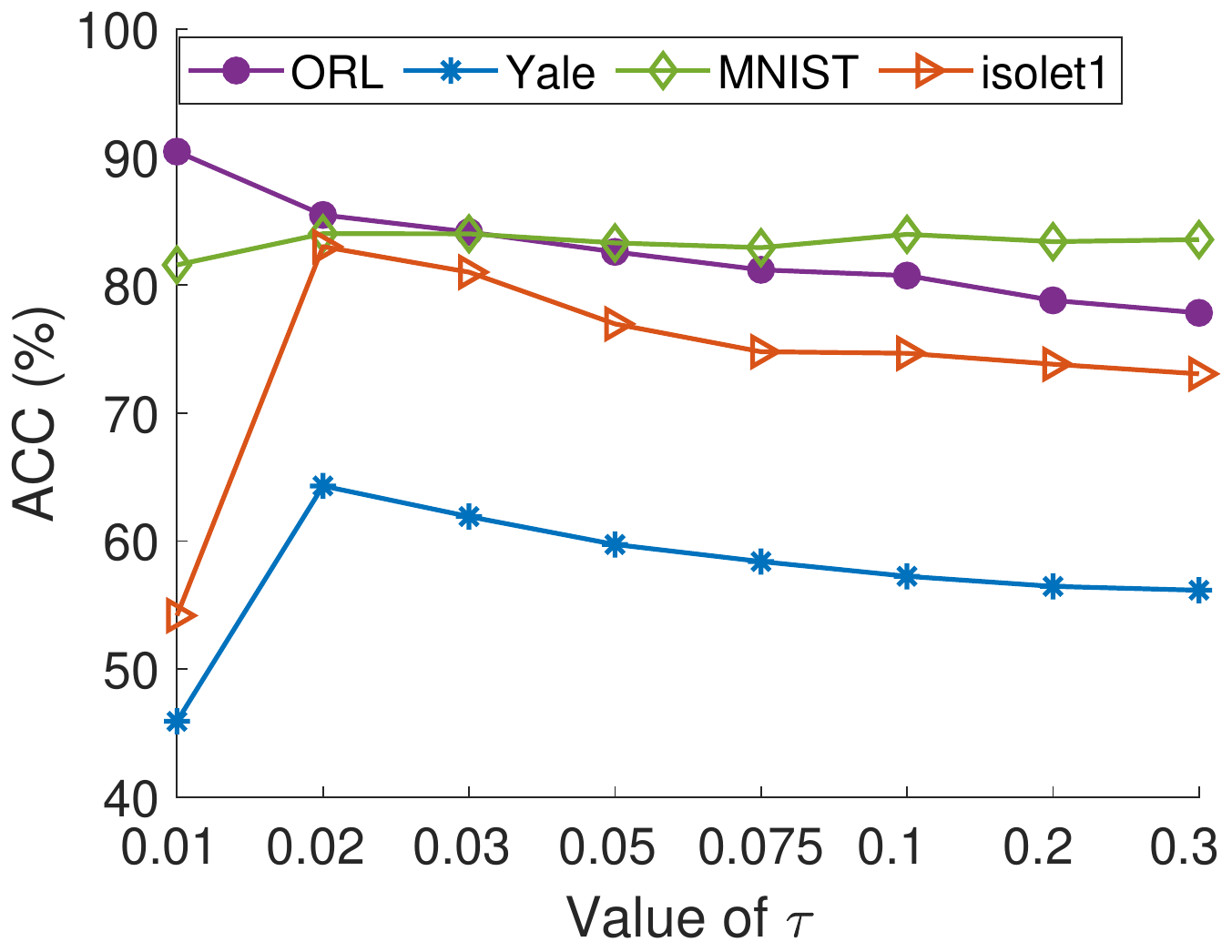}%
}
\hfil
\subfloat[NMI versus $\alpha_1$]{\includegraphics[width=1.75in]{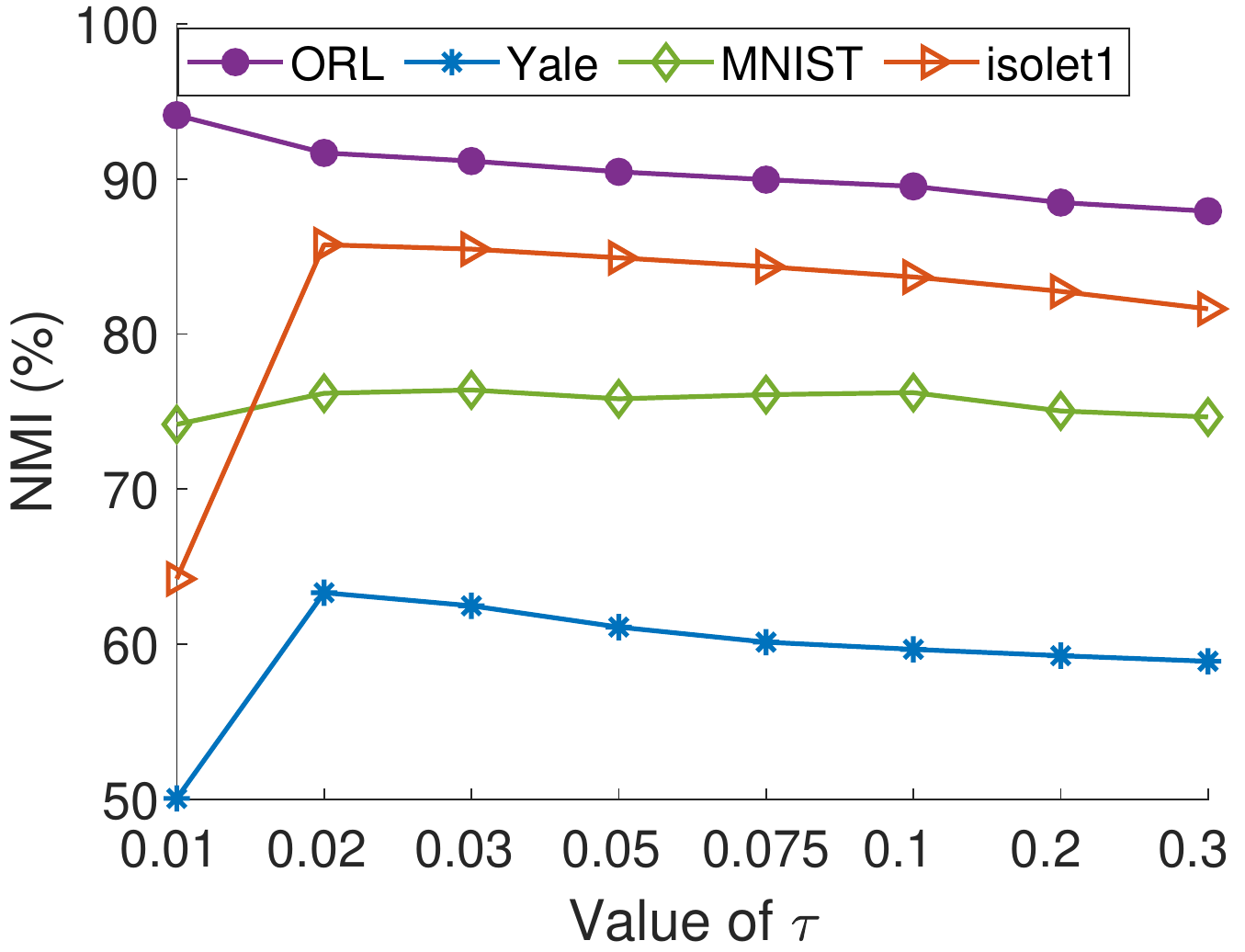}}%
\hfil
\subfloat[ACC versus $\lambda_Z$]{\includegraphics[width=1.75in]{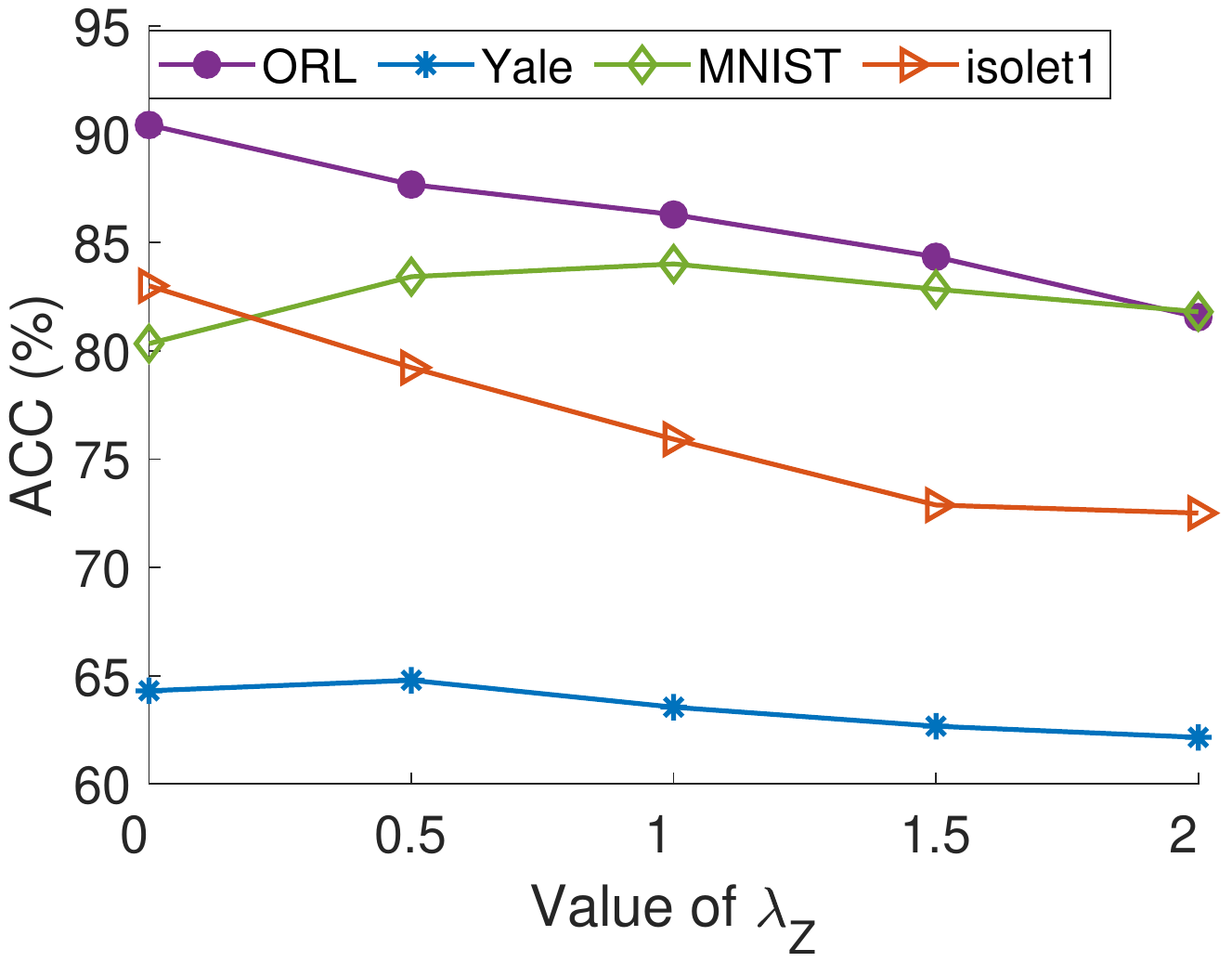}%
}
\hfil
\subfloat[NMI versus $\lambda_Z$]{\includegraphics[width=1.75in]{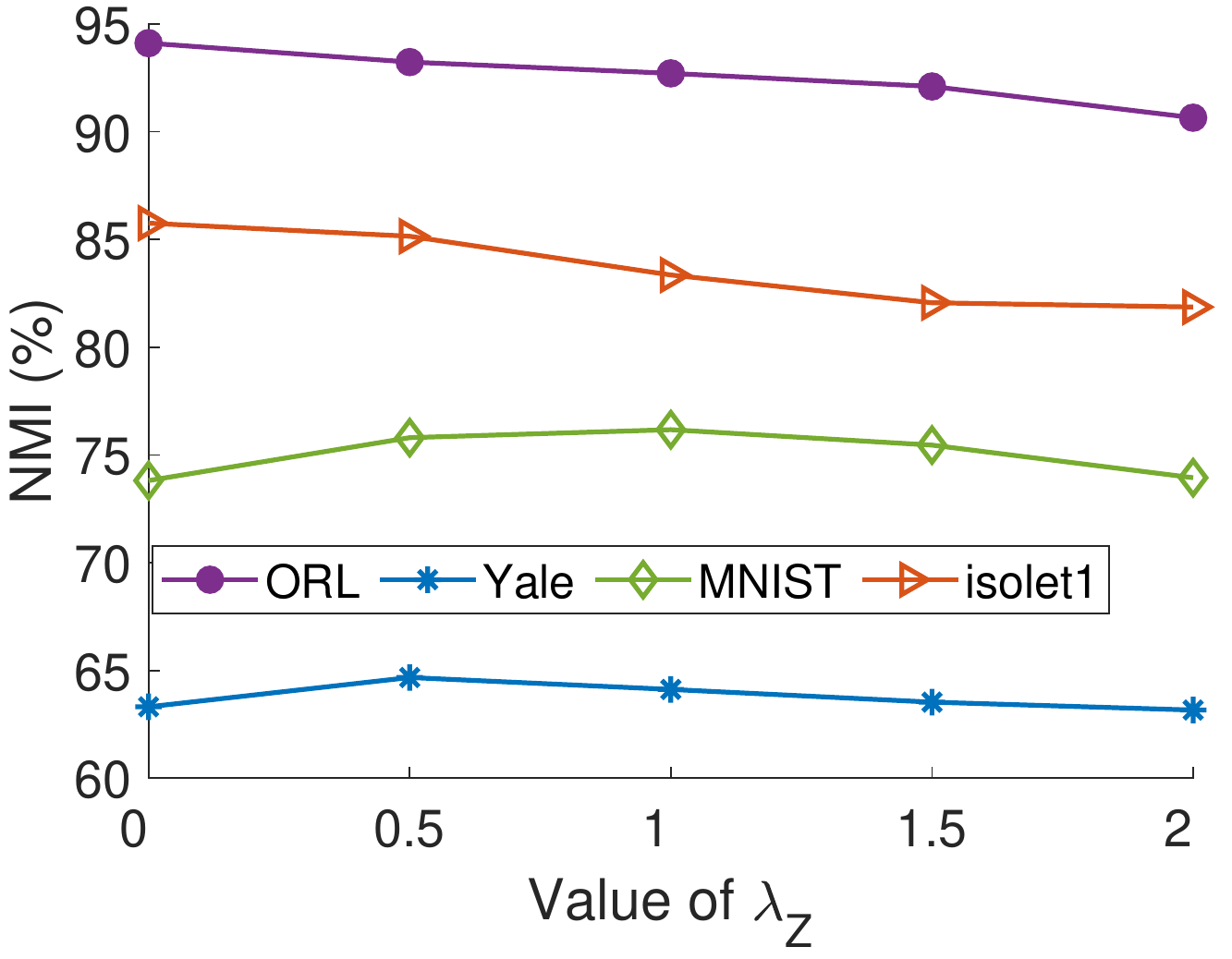}}%
\\
\subfloat[ACC versus $\alpha_2$]{\includegraphics[width=1.75in]{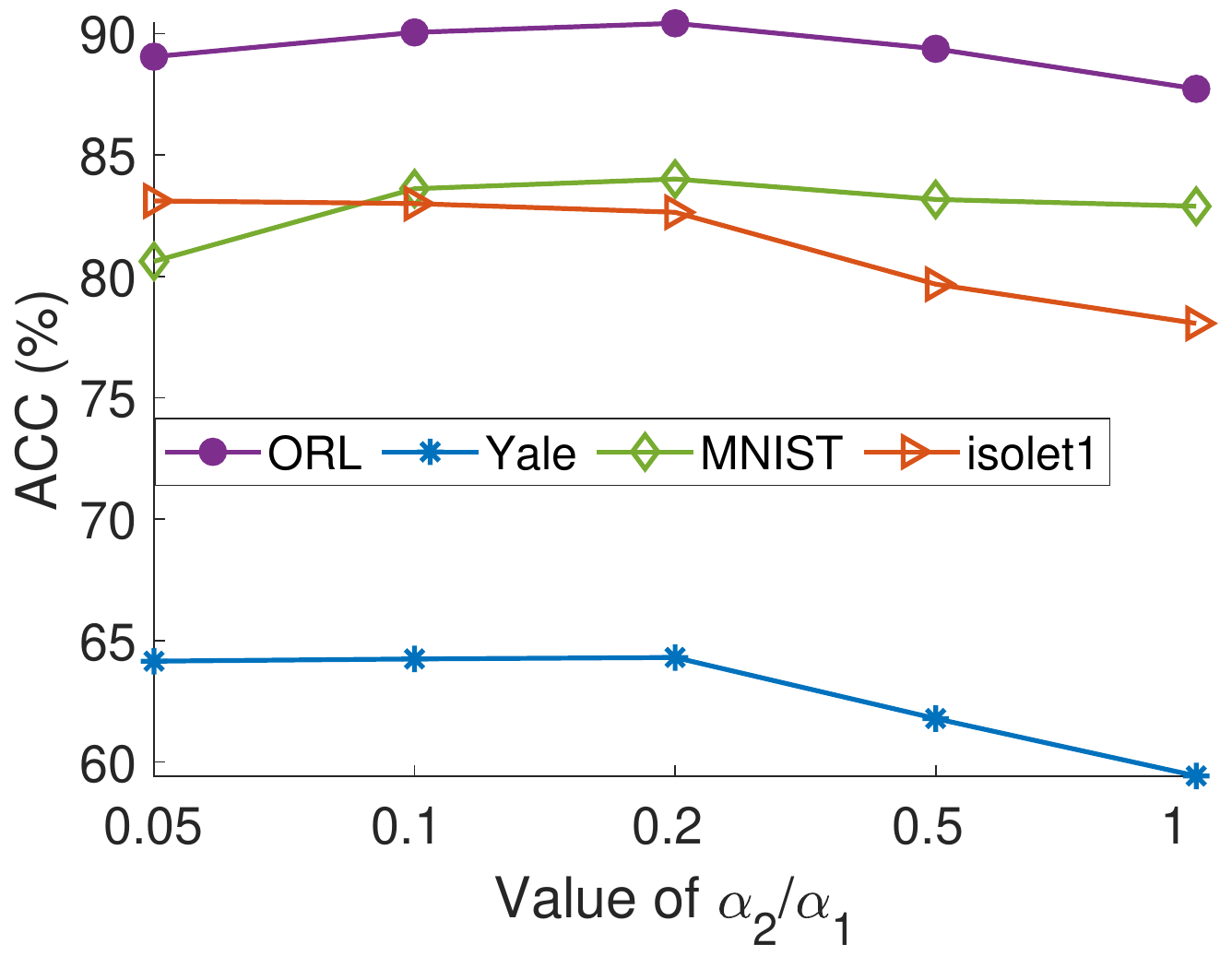}}
\hfil
\subfloat[NMI versus $\alpha_2$]{\includegraphics[width=1.75in]{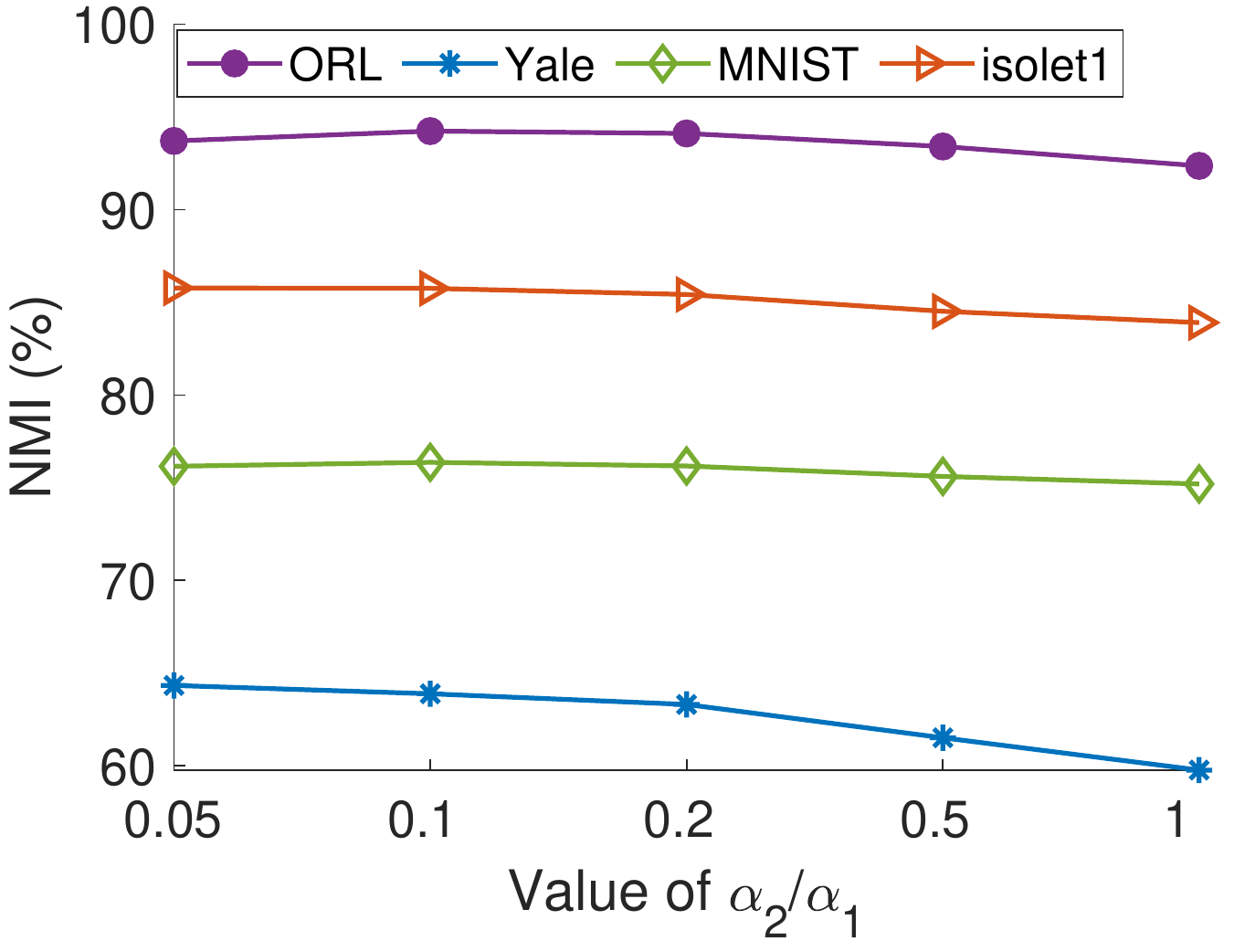}}
\hfil
\subfloat[ACC versus $\lambda_M$]{\includegraphics[width=1.75in]{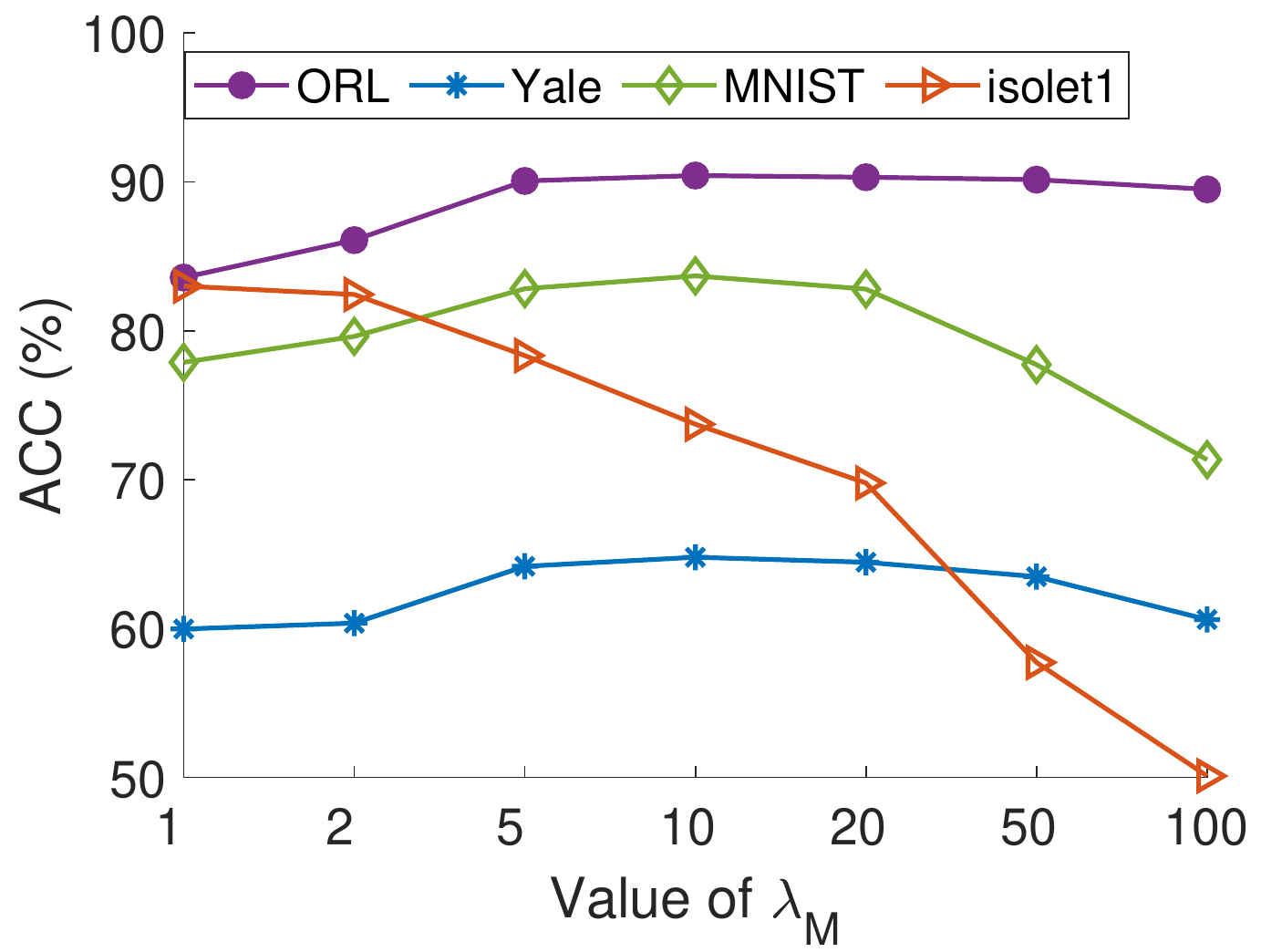}}
\hfil
\subfloat[NMI versus $\lambda_M$]{\includegraphics[width=1.75in]{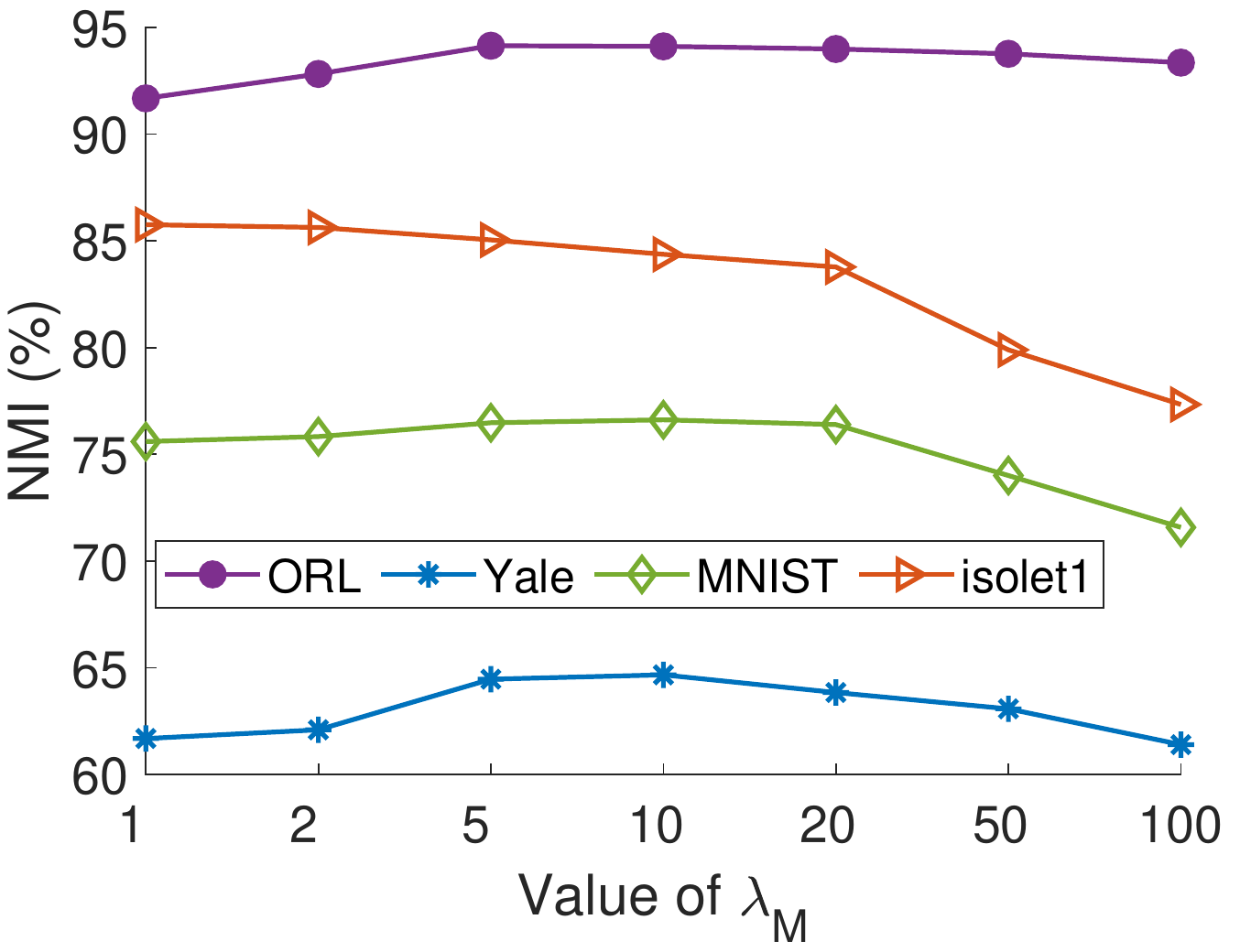}}
\caption{Clustering performance of our method on different datasets with different values of the parameters $\alpha_1$, $\lambda_Z$, $\alpha_2$ and $\lambda_M$.}
\label{fig:sensitivity of different parameters}
\end{figure*}

\begin{figure*}[!t]
\centering
\subfloat[]{\includegraphics[width=1.7in]{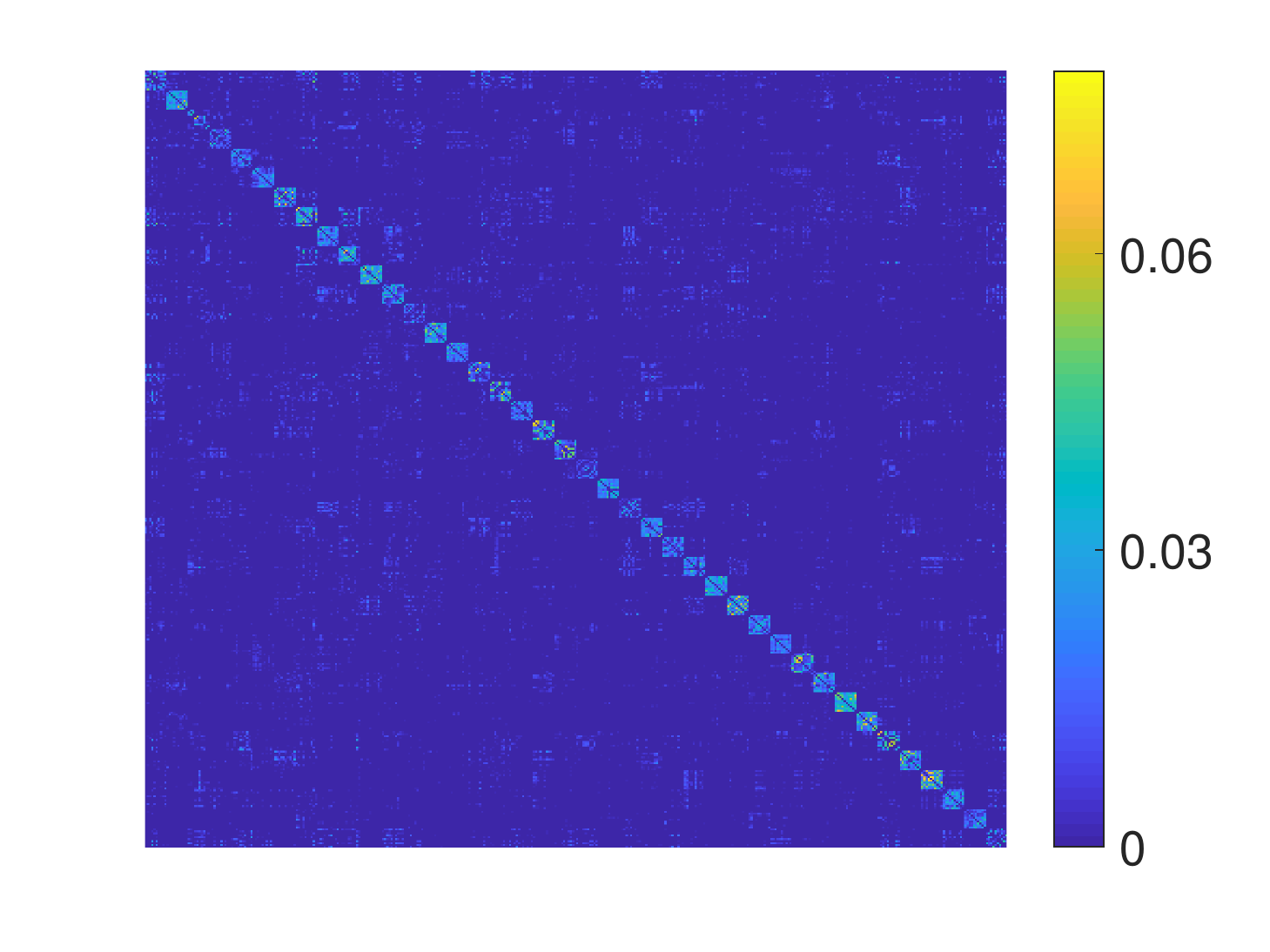}%
}
\hfil
\subfloat[]{\includegraphics[width=1.7in]{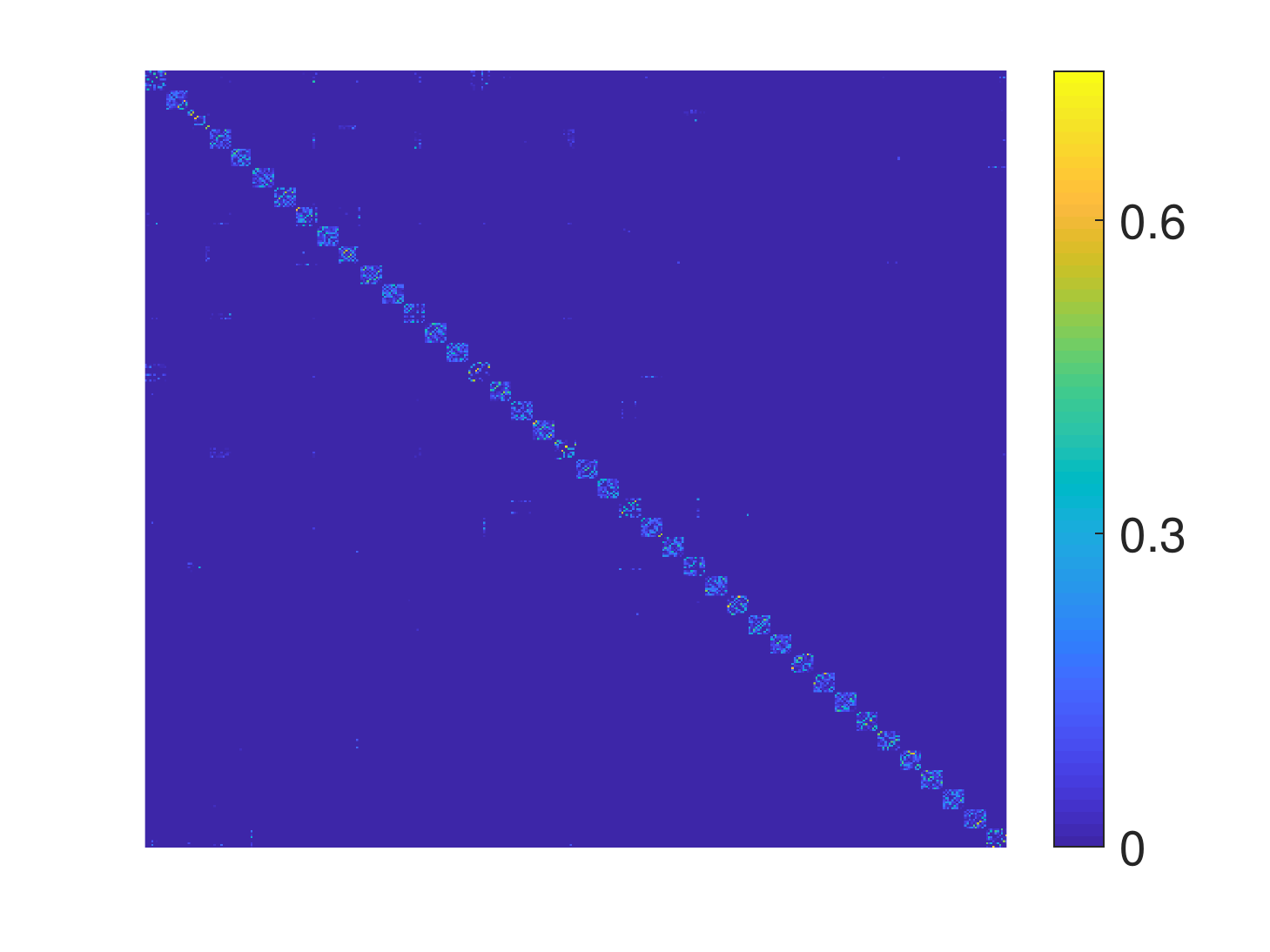}}%
\hfil
\subfloat[]{\includegraphics[width=1.7in]{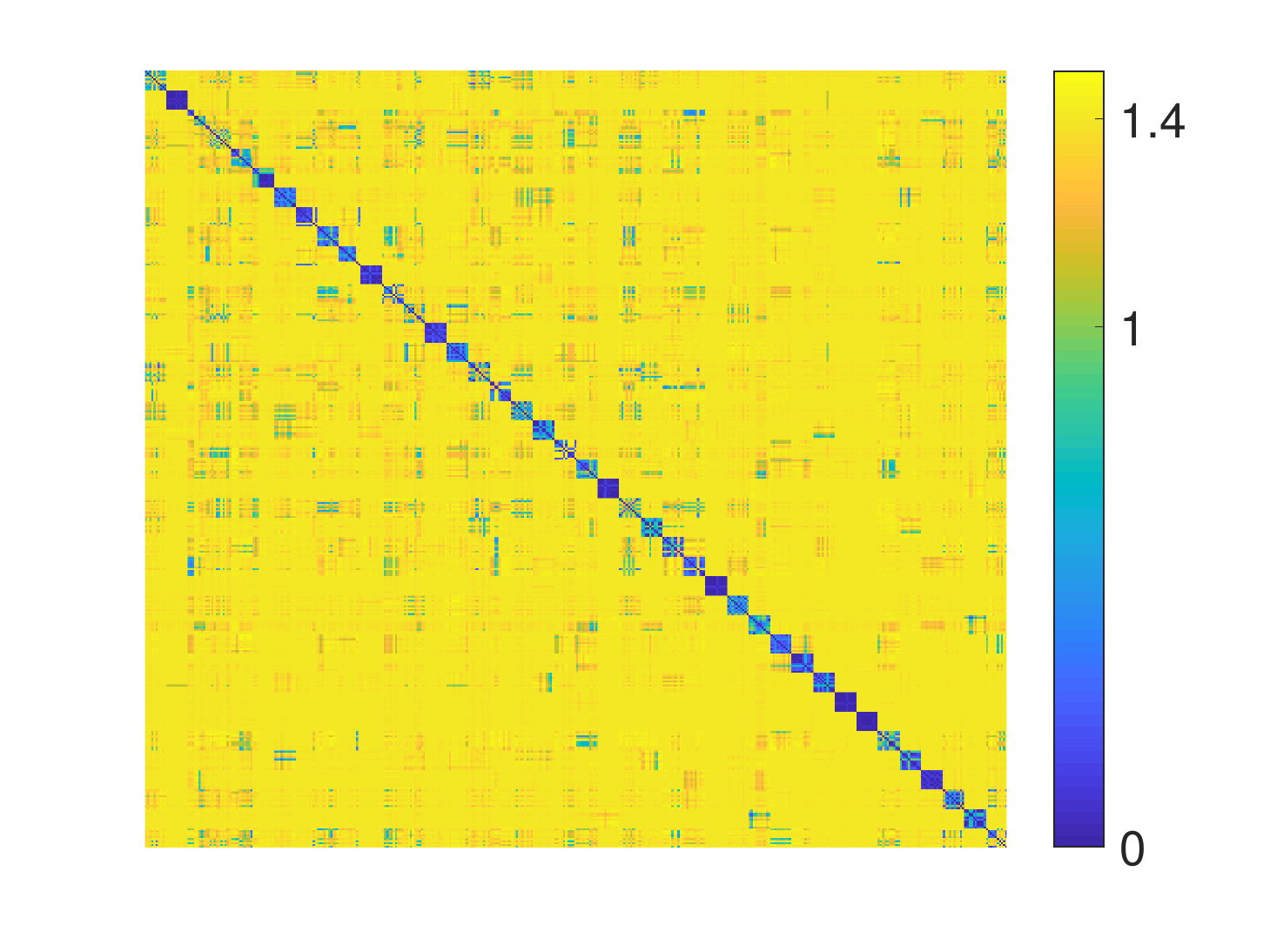}%
}
\hfil
\subfloat[]{\includegraphics[width=1.7in]{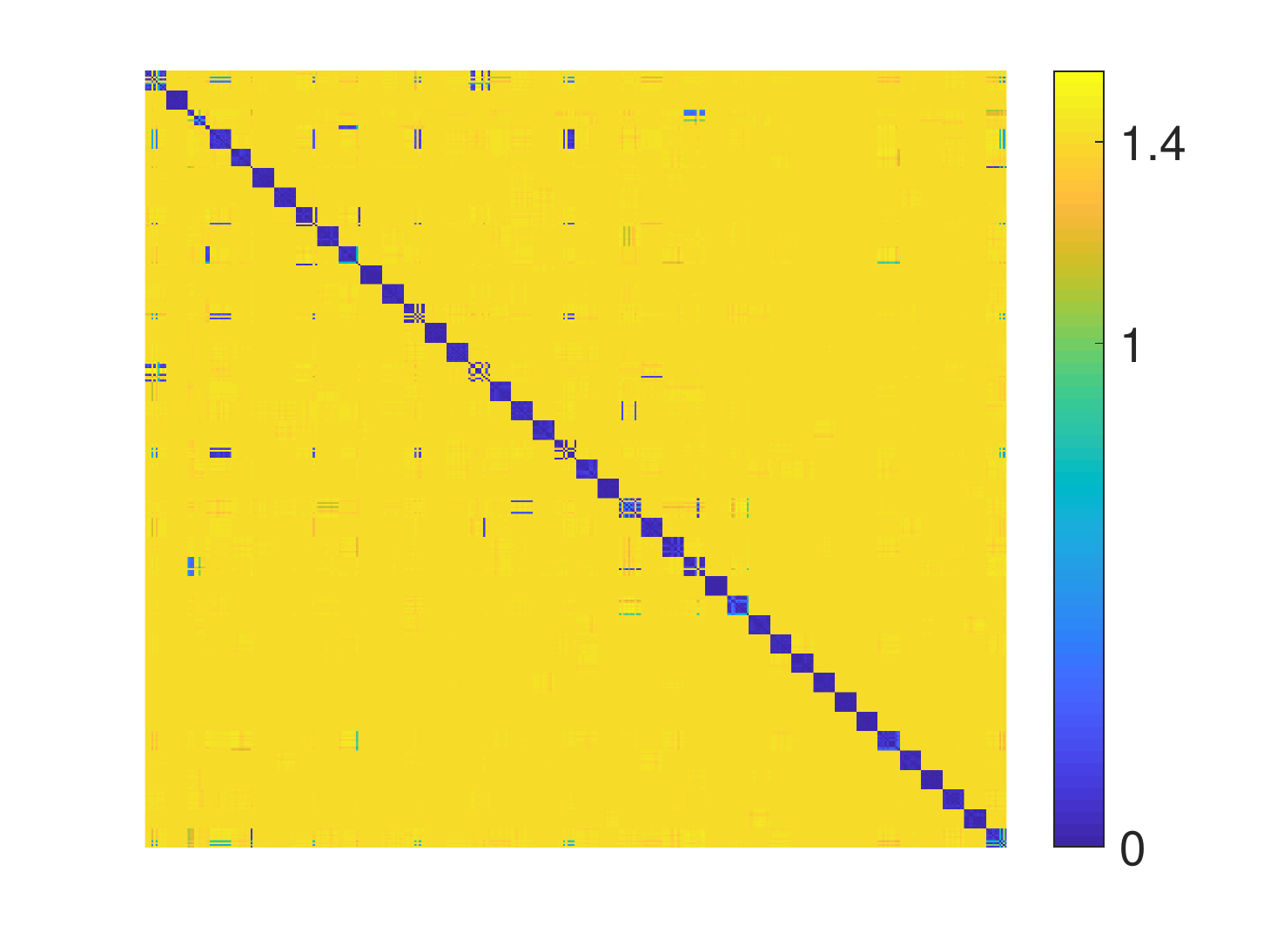}}%
\caption{Plots of matrices in Algorithm \ref{alg:DGSLv1 with normalization operation} for ORL dataset. (a) $|\mathbf{Z}^k|$, k=1. (b) $|\mathbf{Z}^k|$, k=30. (c) Distance matrix $\mathbf{P}^k$, with $\mathbf{P}^k_{ij}=\|\frac{\mathbf{h}_i}{\|\mathbf{h}_i\|}-\frac{\mathbf{h}_j}{\|\mathbf{h}_j\|}\|$, k=1. (d) Distance matrix $\mathbf{P}^k$, with $\mathbf{P}^k_{ij}=\|\frac{\mathbf{h}_i}{\|\mathbf{h}_i\|}-\frac{\mathbf{h}_j}{\|\mathbf{h}_j\|}\|$, k=30.}
\label{fig:visualization of representation}
\end{figure*}
\subsection{Experiments on Incomplete Labeled Classes}
To further verify that our proposed DGSL can utilize the supervisory information effectively, we conduct experiments on generating pairwise constraints with incomplete labeled classes. For each dataset, $k_0$ labeled classes are chosen. For any chosen class $i$, we randomly select $f$ data points, denoted as $S_i$, to generate the pairwise constraints. Then we define the pairwise constraints, $\mathcal{M}=\{(\mathbf{x}_i,\mathbf{x}_j)|\mathbf{x}_i\in S_k,\mathbf{x}_j\in S_t,k=t\}$ and $\mathcal{C}=\{(\mathbf{x}_i,\mathbf{x}_j)|\mathbf{x}_i\in S_k,\mathbf{x}_j\in S_t,k\neq t\}$. We set $f$ as 2, 2, 5, and 5 for the datasets ORL, Yale, MNIST, and isolet1, respectively. Then we choose $k_0$ as $20\%,40\%,60\%,80\%$ and $100\%$ of the total number of classes for each dataset. Each experiment is repeated 20 times with different supervisory information, and we report the average results in Fig. \ref{fig:incomplete}. The following observations can be made: (1) The performance of DGSL improves rapidly on datasets ORL, MNIST, and isolet1, with the increasing of the number of labeled classes $k_0$. It further demonstrates that DGSL can utilize the supervisory information effectively; (2) DGSL achieves higher ACC and NMI than the compared methods in most cases, which further verifies the effectiveness of our approach.




\subsection{Parameters Sensitivity Study}
Our proposed DGSL introduces five parameters, i.e., $\alpha_1,\alpha_2,\lambda$, $\lambda_M$ and $\lambda_Z$ to control the ratio of different components. Since we fix $\lambda=100$, we study the effect of parameters $\alpha_1,\alpha_2$, $\lambda_M$ and $\lambda_Z$ in this section. We generate the pairwise constraints by the first setting in Section \ref{section:Comparisons with Methods Using Pairwise Constraints} and set $f$ as 2, 2, 5 and 5 for ORL, Yale, MNIST and isolet1, respectively. We fix all other parameters except the tested one and show the results in Fig. \ref{fig:sensitivity of different parameters}. For parameter $\alpha_1$, we set $\alpha_1 = 2\tau\lambda\text{Tr}(\mathbf{H}^1\mathbf{L}_\mathbf{C}{\mathbf{H}^1}^\top)$ and tune $\tau$ from  $\{0.01,0.02,0.03,0.05,0.075,0.1,0.2,0.3\}$. From Fig. \ref{fig:sensitivity of different parameters}, the following observations can be made: (1) The parameter $\alpha_1$ is relatively vital to the clustering performance. Thus the regularization term $\text{Tr}(\mathbf{HL}_{\mathbf{Z}}\mathbf{H}^\top)$ plays an important role in the reciprocal learning of $\mathbf{H}$ and $\mathbf{Z}$; (2) The clustering performance of DGSL is robust to the change of the parameter $\alpha_2$. Moreover, $\frac{\alpha_2}{\alpha_1}=0.2$ is a good choice; (3) The clustering performance of DGSL on datasets ORL, Yale and MNIST is robust to the parameters $\lambda_M$ and $\lambda_Z$ where $\lambda_M=10$ is a good choice for these three datasets; (4) The metric NMI is more robust to the metric ACC for the four parameters.

\subsection{Visualization of Representation and Affinity}
We plot the affinity matrix $|\mathbf{Z}|$ and the distance matrix $\mathbf{P}$ of the low-dimensional representations $\mathbf{H}$ at the first iteration and the 30th iteration in Algorithm \ref{alg:DGSLv1 with normalization operation}, as shown in Fig. \ref{fig:visualization of representation}. The first setting of pairwise constraints in Section \ref{section:Comparisons with Methods Using Pairwise Constraints} is used. We set $f$ as 2 for the ORL dataset. From Fig. \ref{fig:visualization of representation}, we can observe that the affinity matrix $|\mathbf{Z}|$ and the distance matrix $\mathbf{P}$ have a clear block diagonal structure at the 30th iteration, which means that the intra-class distance is small and the inter-class distance is relatively large for $\mathbf{H}$ while the intra-class affinity is high and the inter-class affinity is low for $|\mathbf{Z}|$. Thus the low-dimensional representations $\mathbf{H}$ and the affinity matrix $|\mathbf{Z}|$ are mutually refined during the iteration process.

\subsection{Experiments on Datasets with Hypergraph Structure}
A hypergraph includes a vertex set $\mathcal{V}$ and a hyperedge set $\mathcal{E}$. Each hyperedge $e$ is a subset of $\mathcal{V}$  associated with a positive weight $w(e)$. The hypergraph structure can be represented by a $|\mathcal{V}|\times|\mathcal{E}|$ incidence matrix $\mathbf{U}$, with elements $
\mathbf{U}(v,e) = 1$ if $ v\in e$ and 0 otherwise. The degree of a vertex $v$  is defined as $d(v)=\sum_{e\in\mathcal{E}}~\omega(e)\mathbf{U}(v,e)$ and the degree of a hyperedge $e$ is defined as $\delta(e)=\sum_{v\in\mathcal{V}}\mathbf{U}(v,e)$. We denote $\mathbf{D}_e$,  $\mathbf{D}_v$ and $\mathbf{W}_e$  as the diagonal matrices containing the hyperedge degrees, the vertex degrees, and the hyperedge weights, respectively. Then the hypergraph Laplacian  is defined as $\Delta=\mathbf{I}-\mathbf{O}$, where $\mathbf{O}=\mathbf{D}_{v}^{-1 / 2} \mathbf{U} \mathbf{W}_e \mathbf{D}_{e}^{-1} \mathbf{U}^{\top} \mathbf{D}_{v}^{-1 / 2}$.

\begin{figure}[H]
\centering
\includegraphics[width=2.6in]{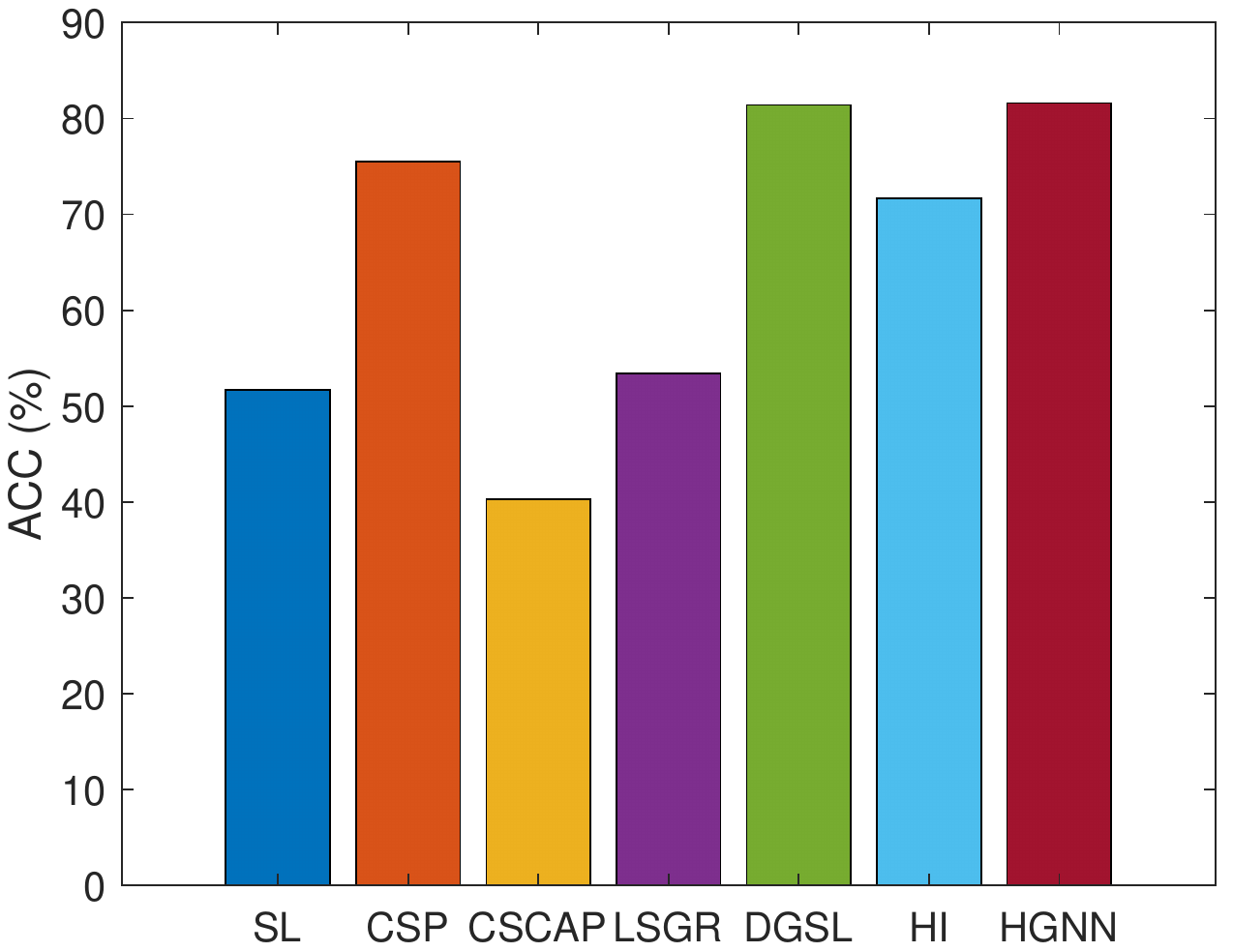}
\caption{Clustering performance (ACC$\%$) on the Cora dataset.}
\label{fig:experiments on hypergraph}
\end{figure}

We conduct experiments on the dataset Cora \cite{sen2008collective}  and construct the hypergraph following the setting in \cite{feng2019hypergraph}: each hyperedge is built by linking one vertex and their neighbors according to the adjacency relation on the graph. For the compared methods SL \cite{kamvar2003spectral}, CSP \cite{wang2014constrained}, CSCAP \cite{lu2008constrained} and LSGR \cite{yang2014unified}, we construct the affinity matrix as $\mathbf{W} + \gamma_2 \mathbf{O}$, where $\mathbf{W}$ is defined in (\ref{eq:construct W}) and we set $m=7,l=5$ for the construction of $\mathbf{W}$. For our proposed DGSL, we substitute $\mathbf{W}$ in (\ref{optim:SSC-TR-bridge-v1}) by $\mathbf{W} + \gamma_2 \mathbf{O}$. We also compare with two state-of-the-art hypergraph learning methods, i.e., HI \cite{zhou2006learning} and HGNN \cite{feng2019hypergraph}, and we follow the setting in \cite{feng2019hypergraph} to generate the labeled data and the test data. The clustering performance is shown in Fig. \ref{fig:experiments on hypergraph}. From Fig. \ref{fig:experiments on hypergraph}, it can be observed that our proposed DGSL can achieve competitive clustering accuracy compared with the state-of-the-art hypergraph learning method. Moreover, DGSL adopts weaker and more flexible supervisory information, i.e., pairwise constraints, than HGNN which adopts partial labels as supervision. 

\section{CONCLUSION} 
We propose a novel dynamic graph structure learning method for semi-supervised clustering. In this method, we simultaneously optimize the graph structure and the low-dimensional representations of data points in a unified optimization framework. Moreover, we construct the graph structure by integrating the local distance and global self-representation among data points. An alternating minimization scheme is proposed to solve the unified optimization framework with proven convergence. Extensive experiments are conducted on eight benchmark datasets, including face images, object images, spoken letters, and handwritten digits, to demonstrate the effectiveness of our approach. Furthermore, we extend our approach to a benchmark hypergraph dataset and achieve competitive performance compared with state-of-the-art hypergraph learning methods.


{\appendix
\begin{IEEEproof}[Proof of Proposition \ref{prop:laplacian equation}]
Note that $\mathbf{L}_{\mathbf{S}}=\mathbf{D}_{\mathbf{S}}-\frac{|\mathbf{S}|+|\mathbf{S}|^{\top}}{2}$, where $\mathbf{D}_{\mathrm{S}}$ is a diagonal matrix with the $i$-th diagonal element being $\sum_{j=1}^{n} \frac{\left|s_{i j}\right|+\left|s_{j i}\right|}{2}$. Then we can derive that
\begin{equation}
    \begin{aligned}
    &\operatorname{Tr}\left(\mathbf{H L}_{\mathbf{S}} \mathbf{H}^{\top}\right) \\
    =&\operatorname{Tr}\left(\mathbf{H D}_{\mathbf{S}} \mathbf{H}^{\top}\right) - \operatorname{Tr}\left(\mathbf{H}\left(\frac{|\mathbf{S}|+|\mathbf{S}|^{\top}}{2}\right) \mathbf{H}^{\top}\right)\\
    =&\sum_{i=1}^n\left(\sum_{j=1}^{n} \frac{\left|s_{i j}\right|+\left|s_{j i}\right|}{2}\right)\|\mathbf{h}_i\|^2-\sum_{i,j=1}^n\frac{\left|s_{i j}\right|+\left|s_{j i}\right|}{2}\mathbf{h}_i^\top\mathbf{h}_j \\
    =&\frac{1}{2}\sum_{i,j=1}^n\frac{\left|s_{i j}\right|+\left|s_{j i}\right|}{2} \left(\|\mathbf{h}_i\|^2+\|\mathbf{h}_j\|^2\right)-\sum_{i,j=1}^n\frac{\left|s_{i j}\right|+\left|s_{j i}\right|}{2}\mathbf{h}_i^\top\mathbf{h}_j\\
    =&\frac{1}{2}\sum_{i,j=1}^n\frac{\left|s_{i j}\right|+\left|s_{j i}\right|}{2} \left(\|\mathbf{h}_i-\mathbf{h}_j\|^2\right) \\
    =&\frac{1}{2}\sum_{i,j=1}^n\left|s_{i j}\right| \|\mathbf{h}_i-\mathbf{h}_j\|^2.
    \end{aligned}
\end{equation}
The proof is completed.
\end{IEEEproof}

\begin{IEEEproof}[Proof of Proposition \ref{main proposition}]
We follow the proof of Proposition 8 in \cite{lu2018subspace}. From the updating rule of $\mathbf{H}^{k+1}$ in (\ref{eq:update H}), we can obtain
\begin{equation}
\label{ineq:H}
f\left(\mathbf{A}^{k}, \mathbf{Z}^{k}, \mathbf{H}^{k+1}\right)+\iota_{S_{2}}\left(\mathbf{H}^{k+1}\right) \leq f\left(\mathbf{A}^{k}, \mathbf{Z}^{k}, \mathbf{H}^{k}\right)+\iota_{S_{2}}\left(\mathbf{H}^{k}\right) .
\end{equation}
From the updating rule of $\mathbf{A}^{k+1}$ in (\ref{eq:update A}), we can obtain
\begin{equation}
\mathbf{A}^{k+1}=\arg \min _{\mathbf{A}} f\left(\mathbf{A}, \mathbf{Z}^{k}, \mathbf{H}^{k+1}\right) .
\end{equation}
Note that $f\left(\mathbf{A}, \mathbf{Z}^{k}, \mathbf{H}^{k+1}\right)$ is $\lambda$-strongly convex w.r.t. $\mathbf{A}$. We can obtain
\begin{equation}
\label{ineq:A}
f\left(\mathbf{A}^{k+1}, \mathbf{Z}^{k}, \mathbf{H}^{k+1}\right) \leq f\left(\mathbf{A}^{k}, \mathbf{Z}^{k}, \mathbf{H}^{k+1}\right)-\frac{\lambda}{2}\left\|\mathbf{A}^{k+1}-\mathbf{A}^{k}\right\|^{2}
\end{equation}
where we use the Lemma B.5 in \cite{mairal2013optimization}. Similarly, note that $f\left(\mathbf{A}^{k+1}, \mathbf{Z}, \mathbf{H}^{k+1}\right)+\iota_{S_{1}}(\mathbf{Z})$ is $\lambda$-strongly convex w.r.t. $\mathbf{Z}$. We can obtain
\begin{equation}
\label{ineq:3}
\begin{aligned}
 &f\left(\mathbf{A}^{k+1},\mathbf{Z}^{k+1}, \mathbf{H}^{k+1}\right)+\iota_{S_{1}}\left(\mathbf{Z}^{k+1}\right)\\ \leq &f\left(\mathbf{A}^{k+1}, \mathbf{Z}^{k}, \mathbf{H}^{k+1}\right)+\iota_{S_{1}}\left(\mathbf{Z}^{k}\right)-\frac{\lambda}{2}\left\|\mathbf{Z}^{k+1}-\mathbf{Z}^{k}\right\|^{2}.
\end{aligned}
\end{equation}
Integrating (\ref{ineq:H}), (\ref{ineq:A}) and (\ref{ineq:3}), we can obtain
\begin{equation}
\label{eq:SSC-TR-conv}
\begin{aligned}
& f\left(\mathbf{A}^{k+1}, \mathbf{Z}^{k+1}, \mathbf{H}^{k+1}\right)+\iota_{S_{1}}\left(\mathbf{Z}^{k+1}\right)+\iota_{S_{2}}\left(\mathbf{H}^{k+1}\right) \\
\leq & f\left(\mathbf{A}^{k}, \mathbf{Z}^{k}, \mathbf{H}^{k}\right)+\iota_{S_{1}}\left(\mathbf{Z}^{k}\right)+\iota_{S_{2}}\left(\mathbf{H}^{k}\right) \\
&-\frac{\lambda}{2}\left\|\mathbf{Z}^{k+1}-\mathbf{Z}^{k}\right\|^{2}-\frac{\lambda}{2}\left\|\mathbf{A}^{k+1}-\mathbf{A}^{k}\right\|^{2} .
\end{aligned}
\end{equation}
Note that $f\left(\mathbf{A}^{k}, \mathbf{Z}^{k}, \mathbf{H}^{k}\right)+\iota_{S_{1}}\left(\mathbf{Z}^{k}\right)+\iota_{S_{2}}\left(\mathbf{H}^{k}\right) \geq 0 .$ Now, summing (\ref{eq:SSC-TR-conv}) over $k=0,1, \dots,$ we can obtain
\begin{equation}
\sum_{k=0}^{+\infty} \frac{\lambda}{2}\left(\left\|\mathbf{Z}^{k+1}-\mathbf{Z}^{k}\right\|^{2}+\left\|\mathbf{A}^{k+1}-\mathbf{A}^{k}\right\|^{2}\right) \leq f\left(\mathbf{A}^{0}, \mathbf{Z}^{0}, \mathbf{H}^{0}\right) .
\end{equation}

This implies
\begin{equation}
\label{BDR:convergence of Z}
\mathbf{Z}^{k+1}-\mathbf{Z}^{k} \rightarrow 0,
\end{equation}
\begin{equation}
\label{BDR:convergence of A}
\mathbf{A}^{k+1}-\mathbf{A}^{k} \rightarrow 0.
\end{equation}

From (\ref{eq:SSC-TR-conv}), $f(\mathbf{A}^k,\mathbf{Z}^k,\mathbf{H}^k)+\iota_{S_1}(\mathbf{Z}^k)+\iota_{S_2}(\mathbf{H}^k)$ is monotonically decreasing and thus it is upper bounded. From the expression of $f(\mathbf{A}^k,\mathbf{Z}^k,\mathbf{H}^k)$, it is easy to see that $\{\|\mathbf{Z}^k\|_1\}$ and $\{\|\mathbf{A}^k-\mathbf{Z}^k\|^2\}$ are bounded. Then $\{\mathbf{A}^k\}$ and $\{\mathbf{Z}^k\}$ are bounded. Also, $\mathbf{H}^k\in S_2$ implies that $\mathbf{H}^k\mathbf{H}^{k^\top}=\mathbf{I}$ and thus $\{\mathbf{H}^k\}$ is bounded. The proof is completed.
\end{IEEEproof}

\begin{IEEEproof}[Proof of Theorem \ref{main theorem}]
From the boundedness of $\{\mathbf{A}^k,\mathbf{Z}^k,\mathbf{H}^{k}\}$, there exists a point $(\mathbf{A}^*,\mathbf{Z}^*,\mathbf{H}^*)$ and a subsequence $\{\mathbf{A}^{k_j+1},\mathbf{Z}^{k_j+1},\mathbf{H}^{k_j+1}\}$ such that $\mathbf{A}^{k_j+1}\rightarrow \mathbf{A}^*$, $\mathbf{Z}^{k_j+1}\rightarrow \mathbf{Z}^*$ and $\mathbf{H}^{k_j+1}\rightarrow \mathbf{H}^*$. Then by (\ref{BDR:convergence of Z}) and (\ref{BDR:convergence of A}), we have $\mathbf{A}^{k_j}\rightarrow \mathbf{A}^*$, $\mathbf{Z}^{k_j}\rightarrow \mathbf{Z}^*$. On the other hand, from the optimality of $\mathbf{H}^{k_j+1}$ for (\ref{eq:update H}), $\mathbf{A}^{k_j+1}$ for (\ref{eq:update A}), $\mathbf{Z}^{k_j+1}$ for ($\ref{eq:update Z}$), we have
    \begin{align}
    & f(\mathbf{A}^{k_j},\mathbf{Z}^{k_j},\mathbf{H}^{k_j+1}) + \iota_{S_2}(\mathbf{H}^{k_j+1}) \label{limit1} \\
    &\leq f(\mathbf{A}^{k_j},\mathbf{Z}^{k_j},\mathbf{H}) + \iota_{S_2}(\mathbf{H}), \forall ~ \mathbf{H} \label{limit2}\\
    & f(\mathbf{A}^{k_j+1},\mathbf{Z}^{k_j},\mathbf{H}^{k_j+1}) \leq f(\mathbf{A},\mathbf{Z}^{k_j},\mathbf{H}^{k_j+1}), \forall ~\mathbf{A} \label{limit3}\\
    & f(\mathbf{A}^{k_j+1},\mathbf{Z}^{k_j+1},\mathbf{H}^{k_j+1}) + \iota_{S_1}(\mathbf{Z}^{k_j+1}) \label{limit4}\\
    &\leq f(\mathbf{A}^{k_j+1},\mathbf{Z},\mathbf{H}^{k_j+1}) + \iota_{S_1}(\mathbf{Z}), \forall ~ \mathbf{Z} \label{limit5}
    \end{align}
Let $k_j \rightarrow +\infty$ in (\ref{limit1})-(\ref{limit5}). We can obtain
  \begin{align*}
     f(\mathbf{A}^*,\mathbf{Z}^*,\mathbf{H}^{*}) + \iota_{S_2}(\mathbf{H}^{*})  &\leq f(\mathbf{A}^{*},\mathbf{Z}^{*},\mathbf{H}) + \iota_{S_2}(\mathbf{H}), \forall ~ \mathbf{H} \\
    f(\mathbf{A}^{*},\mathbf{Z}^{*},\mathbf{H}^{*}) &\leq f(\mathbf{A},\mathbf{Z}^{*},\mathbf{H}^{*}), \forall ~\mathbf{A} \\
    f(\mathbf{A}^{*},\mathbf{Z}^{*},\mathbf{H}^{*}) + \iota_{S_1}(\mathbf{Z}^{*}) &\leq f(\mathbf{A}^{*},\mathbf{Z},\mathbf{H}^{*}) + \iota_{S_1}(\mathbf{Z}), \forall ~ \mathbf{Z} 
    \end{align*}
which implies
\begin{align*}
    0 &\in \partial_{\mathbf{H}}\left(f(\mathbf{A}^*,\mathbf{Z}^*,\mathbf{H}^*) + \iota_{S_1}(\mathbf{Z}^*) + \iota_{S_2}(\mathbf{H}^*)\right) \\
    0 &\in \partial_{\mathbf{A}}\left(f(\mathbf{A}^*,\mathbf{Z}^*,\mathbf{H}^*) + \iota_{S_1}(\mathbf{Z}^*) + \iota_{S_2}(\mathbf{H}^*)\right) \\
    0 &\in \partial_{\mathbf{Z}}\left(f(\mathbf{A}^*,\mathbf{Z}^*,\mathbf{H}^*) + \iota_{S_1}(\mathbf{Z}^*) + \iota_{S_2}(\mathbf{H}^*)\right) \\
\end{align*}
Thus $(\mathbf{A}^*,\mathbf{Z}^*,\mathbf{H}^*)$ is a stationary point of (\ref{optim:SSC-TR-bridge-v1}).
\end{IEEEproof}
}
\bibliographystyle{IEEEtran}
\bibliography{Bibliography}

\begin{IEEEbiography}[{\includegraphics[width=1in,height=1.2in,clip,keepaspectratio]{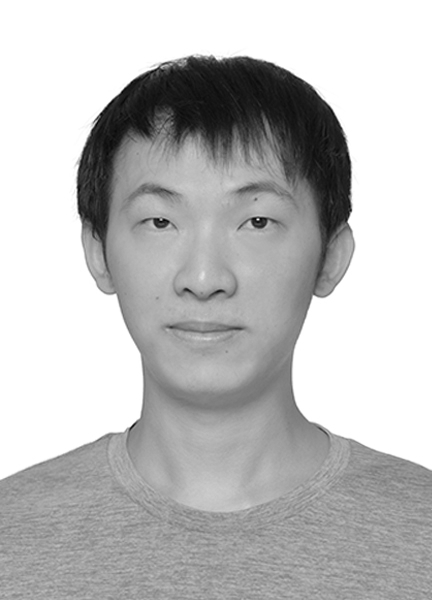}}]{Huaming Ling} received the B.S. degree from Sun Yat-Sen University. He is currently pursuing the Ph.D. degree with the Department of Mathematics, Tsinghua University. His main research interests include pattern recognition, computer vision and machine learning.
\end{IEEEbiography}

\begin{IEEEbiography}[{\includegraphics[width=1in,height=1.2in,clip,keepaspectratio]{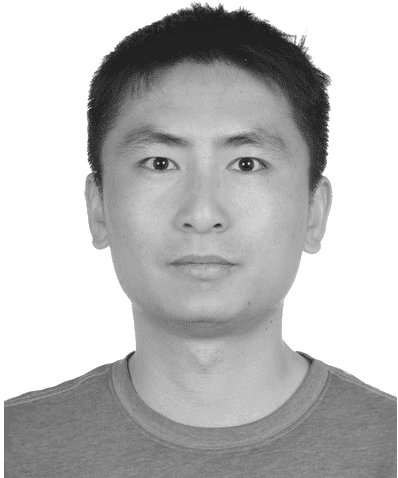}}]{Chenglong Bao} is an assistant professor in Yau Mathematical Sciences Center, Tsinghua University and Yanqi Lake Beijing Institute of Mathematical Sciences and Applications. He received his Ph.D. from the department of mathematics, National University of Singapore in 2014. His main research interests include mathematical image processing, large scale optimization and its applications.
\end{IEEEbiography}

\begin{IEEEbiography}[{\includegraphics[width=1in,height=1.2in,clip,keepaspectratio]{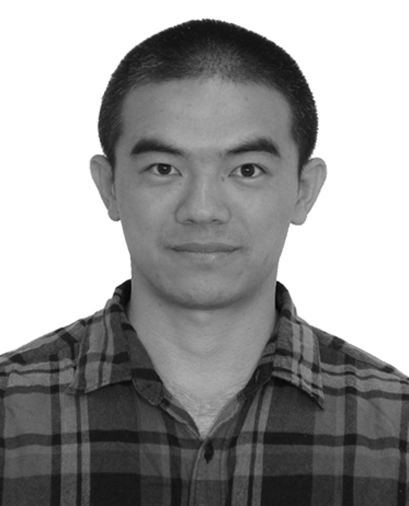}}]{Xin Liang} is an assistant professor in Yau Mathematical Sciences Center, Tsinghua University and Yanqi Lake Beijing Institute of Mathematical Sciences and Applications. He received his Ph.D. from the School of Mathematical Sciences, Peking University in 2014. His main research interests include numerical linear algebra, matrix analysis, and their applications.
\end{IEEEbiography}

\begin{IEEEbiography}[{\includegraphics[width=1in,height=1.2in,clip,keepaspectratio]{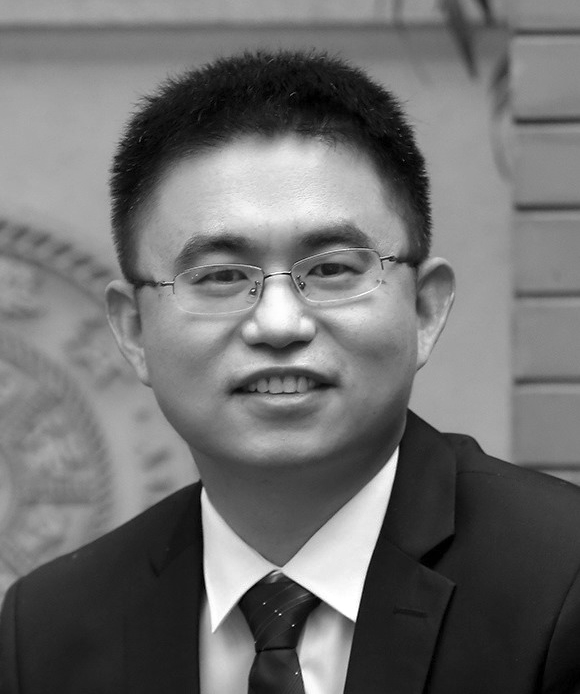}}]{Zuoqiang Shi} is a professor in Yau Mathematical Sciences Center, Tsinghua University and Yanqi Lake Beijing Institute of Mathematical Sciences and Applications. He received his Ph.D. from Zhou Pei-Yuan Center for Applied Mathematics, Tsinghua University in 2008. His main research interests include numerical methods of partial differential equations and their applications, mathematical image processing.
\end{IEEEbiography}

\end{document}